\documentclass{article}
\usepackage{amsmath}

\usepackage[accepted]{icml2018t}

\usepackage{enumitem}
\usepackage{times}
\usepackage{graphicx} 
\usepackage{subfigure} 

\graphicspath{{./}{./fig/}{./mnst/}}

\usepackage{natbib}

\usepackage{algorithm}
\usepackage{algorithmic}

\usepackage{hyperref}

\icmltitlerunning{Active Metric Learning for Supervised Classification}

\newcommand{\mc}{\mathcal}

\usepackage{graphicx} 
\usepackage{hyperref}       
\usepackage{url}            
\usepackage{booktabs}       
\usepackage{amsfonts}       
\usepackage{nicefrac}       
\usepackage{microtype}      

\usepackage{natbib}

\usepackage{subfigure}

\usepackage{algorithm}
\usepackage{algorithmic}
\usepackage{xargs}                      
\usepackage[pdftex,dvipsnames]{xcolor}  

\def\st{{\rm s.t.}}
\usepackage[colorinlistoftodos,prependcaption,textsize=tiny]{todonotes}
\newcommandx{\unsure}[2][1=]{\todo[linecolor=red,backgroundcolor=red!25,bordercolor=red,#1]{#2}}
\newcommandx{\change}[2][1=]{\todo[linecolor=blue,backgroundcolor=blue!25,bordercolor=blue,#1]{#2}}
\newcommandx{\info}[2][1=]{\todo[linecolor=OliveGreen,backgroundcolor=OliveGreen!25,bordercolor=OliveGreen,#1]{#2}}
\newcommandx{\improvement}[2][1=]{\todo[linecolor=Plum,backgroundcolor=Plum!25,bordercolor=Plum,#1]{#2}}
\newcommandx{\thiswillnotshow}[2][1=]{\todo[disable,#1]{#2}}

\usepackage{amssymb}
\renewcommand{\v}[1]{\ensuremath{\mathbf{#1}}}

\newcommand{\D}{\mathbb{D}} 

\newcommand{\vdelta}{\boldsymbol{\delta}}
\newcommand{\vDelta}{\boldsymbol{\Delta}}

\begin{document}

\twocolumn[
\icmltitle{Active Metric Learning for Supervised Classification}
 






\begin{icmlauthorlist}
\icmlauthor{Krishnan Kumaran}{EM}
\icmlauthor{Dimitri Papageorgiou}{EM}
\icmlauthor{Yutong Chang}{LU}
\icmlauthor{Minhan Li}{LU}
\icmlauthor{Martin Taka{c}}{LU}
\end{icmlauthorlist}

\icmlaffiliation{EM}{ExxonMobil Corporate Strategic Research, Annandale, NJ}
\icmlaffiliation{LU}{Lehigh University,
 Bethlehem, PA} 

\icmlcorrespondingauthor{Krishnan Kumaran}{krishnan.kumaran@exxonmobil.com}
\icmlcorrespondingauthor{Dimitri Papageorgiou}{dimitri.j.papageorgiou@exxonmobil.com}
\icmlkeywords{metric learning, mixed-integer optimization}

\vskip 0.1in
]



\printAffiliationsAndNotice{}  

\setlength{\abovedisplayskip}{3pt}
\setlength{\belowdisplayskip}{3pt}

\begin{abstract} 

Clustering and classification critically rely on distance metrics that provide meaningful comparisons between data points.  We present mixed-integer optimization approaches to find optimal distance metrics that generalize the  Mahalanobis metric extensively studied in the literature.  
Additionally, we generalize and improve upon leading methods by removing reliance on pre-designated ``target neighbors,'' ``triplets,'' and ``similarity pairs.''
Another salient feature of our method is its ability to enable active learning by recommending precise regions to sample after an optimal metric is computed to improve classification performance. This targeted acquisition can significantly reduce computational burden by ensuring training data completeness, representativeness, and economy.
We demonstrate classification and computational performance of the algorithms through several simple and intuitive examples, followed by results on real image and medical datasets.
\end{abstract} 

\section{Introduction and Motivation}  


Selecting an appropriate distance metric is fundamental to many learning algorithms such as k-means, nearest neighbor searches, and others, as observed by \cite{davis2007information} and other researchers in this field. Further, they observe that choosing such a measure is highly problem-specific and ultimately dictates the success - or failure - of the learning algorithm. Nevertheless, the algorithm used to select the metric based on data can be designed to be more general, and designing such algorithm(s) is indeed the objective of this work, as well as past research on this problem.

In this work, we formulate a general framework for choosing such metrics that improves and extends previous formulations in some important ways.In particular, we attempt to couple the metric learning problem with that of recommending targeted data acquisition, which has not been sufficiently addressed in past work which has mostly assumed that the data is a given, static collection of N-dimensional vectors.
However, in many real-world settings, one does not have the luxury of learning a once-and-for-all distance metric.
Rather, an iterative approach is required whereby an initial distance metric is learned,
new data is acquired, the distance metric is refined, and so on.   
A main goal of this work is to present a systematic framework for optimizing this iterative procedure so that an optimal and interpretable metric is learned and is, in turn, used to recommend precise regions to sample in order to acquire new data to be used to further refine the metric and improve classification performance.

\subsection{Problem Setting}

As described in \cite{xiang:2008}, there are two prominent batch distance metric learning settings, both of which assume that we are given a set of $N$ points $\v{x}_i \in \Re^D$.  In the first setting, a class (or label) $C_i$ is explicitly given for each point $i \in \mc{N} = \{1,\dots,N\}$.  In the second setting, classes are implicitly furnished through pairwise constraints in the form of must-links and cannot-links. Must-links are given as $\{(i,j) : i \textrm{ and } j \textrm{ are in the same class} \}$, whereas cannot-links are specified as $\{(i,k) : i \textrm{ and } k \textrm{ are not in the same class} \}$.
For both settings, we let $\mc{C}_i$ and $\bar{\mc{C}}_i$ denote the co-class and non-class neighbors of $i$, respectively.  

We ask the question: 
Is there a metric $\D(\mathbf{x},\mathbf{y})$ that enforces the condition that the nearest neighbor of each point is a co-class point, i.e., $\forall i \in \mc{N}$ 
\begin{equation} \label{eq:metric_feasibility}
\textstyle{\min}_{j \in \mc{C}_i} \,\, \D(\mathbf{x}_i,\mathbf{x}_j) < \textstyle{\min}_{k \in \bar{\mc{C}}_i}\,\, \D(\mathbf{x}_i,\mathbf{x}_k)?
\end{equation}
More generally, let $\mc{N}_i^K(\D)=\{j_1,\dots,j_K\}$ be the $K$ nearest neighbors to $i$ with respect to a distance metric $\D$ (assume no ties).  Then, we are interested in finding a distance metric $\D$ satisfying the condition: Given $K$, the majority of the $K$ nearest neighbors are co-class points, i.e., $\forall i \in \mc{N}$   
\begin{equation} \label{eq:metric_feasibility_knn}
\exists \bar{K} =  \lfloor \tfrac{K}{2} \rfloor + 1 \textrm{ points } j_1,\dots,j_{\bar{K}} \in \mc{N}_i^K(\D) \cap \mc{C}_i.
\end{equation}
Note that the above condition does not {\em a priori} define target neighbors, as required by most previous work.  This is an important distinction because the closest neighbors of a point are not determined unless the metric is specified.  Our formulation incorporates variables that compare distances between true neighbors contingent on the distance metric.  This property avoids the pitfalls of pre-specified target neighbors as shown below, while preserving the desirable characteristics of agglomerative and $K$-nearest neighbor clustering methods such as permitting multiple disjoint islands of the same class and non-convex class regions while maintaining simplicity and interpretability of the metric.


\paragraph{Form of Distance Metric.}
In general, we allow the metric to be a power series of the form
\begin{eqnarray}
\D(\mathbf{x},\mathbf{y}) &\equiv& \mathbf{a \cdot |x-y| + (x-y) \cdot B\cdot (x-y)} + \nonumber \\ & &\textstyle{\sum}_{p,q,r=1}^D C_{pqr} |x_p - y_p| \cdot |x_q - y_q| \cdot |x_r - y_r| \nonumber\\ & &+ \textrm{higher order terms}.
\label{eq:metric-definition}
\end{eqnarray}
Strictly speaking, the mapping $\D$ proposed in \eqref{eq:metric-definition} may not satisfy the four properties - non-negativity, symmetry, triangle inequality, distinguishability - that are required to be a ``metric.'' Nevertheless, we use this terminology throughout.  

Restricting to the first term loosely corresponds to SVM/discriminant analysis, while the second term is commonly known in the literature as the \textit{Mahalanobis metric} if the matrix \textbf{B} is symmetric and positive definite. Higher order terms introduce additional parameters at a power law rate, e.g., the fully symmetric tensor $C_{pqr}$ has $O(D^3)$ parameters. In principle, almost any dataset can eventually be fitted with a metric with a potentially infinite number of these parameters. However, in practice, a mis-classification trade-off curve and knee-point can be used to prevent over-fitting. Further, the mis-classified points could point to possible outliers, errors in input data, or class boundaries. In all of these cases, the algorithm points to regions in the space where further data acquisition/quality testing would be of most value. This aspect of our method is unique, and provides significant value in selecting the most effective training for supervised classification in general, even applied to SVM/Deep Neural Networks or other algorithms. Further, we will show in our experimental results that the ratio of closest co-class to closest non-class point, defined as
\begin{equation} \label{eq:R-ratio}
R_i = \tfrac{\min_{j \in \mc{C}_i} \,\, \D(\mathbf{x}_i,\mathbf{x}_j)}{\min_{k \in \bar{\mc{C}}_i}\,\, \D(\mathbf{x}_i,\mathbf{x}_k)}
\end{equation}
is a useful metric to separate ``interior'' points from ``boundary points'' of classes.

Even in that case where we consider only the second-order term, the Mahalanobis distance, our approach differs from past approaches due to condition~\eqref{eq:metric_feasibility} which we show empirically results in better solutions. 


\subsection{Comparison with Prior Work} \label{sec:lit_review}
\cite{wang2015} survey distance metric learning in unsupervised and supervised settings.
Several recent publications have proposed metric learning methods similar to ours, see~\cite{weinberger2009distance,davis2007information,ying2012distance,rosales2006learning,xing2003distance} and references therein.

A fundamental distinction that makes our approach more general than the distance metric learning approaches of these authors
is that ours does not rely on auxiliary input information in the form of so-called target co-class neighbors~\cite{weinberger2009distance,ying2012distance,xing2003distance}, similar and dissimilar point pairs \cite{davis2007information}, nor triplets with one co-class and non-class neighbor~\cite{rosales2006learning}.
The target neighbors of a point are co-class points that the user desires to be closest to it.
Target neighbor-based methods fix a priori a set of points and attempt to learn a linear transformation of the input space such that the resulting nearest neighbors of the point are indeed its target neighbors.
Unfortunately, in many applications target neighbors or triples are not available.
In the absence of prior target neighbor knowledge, \cite{weinberger2009distance} suggest using the $K$ nearest neighbors with the same class label, as determined by Euclidean distance. While these requirements appear reasonable, they can be misleading and highly data-dependent as shown in the example below.
Further, target neighbors and/or triplets, even if available with an initial data set, may become burdensome and/or error-prone to update when additional data becomes available.

\begin{figure}
\centering
\includegraphics[scale=0.6]{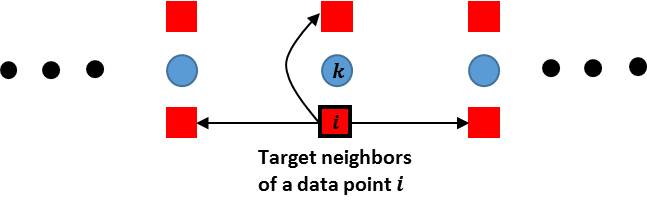}
\caption{Poor choice of target neighbors~\cite{weinberger2009distance} can result in infeasible/distorted metrics. No Mahalanobis metric can bring the target neighbors closer while simultaneously pushing the intermediate non-class neighbor $k$ further. Observe that a simple metric with high vertical weighting and low horizontal weighting will classify correctly in our approach, which does not use target neighbors.}
\label{fig:target-neighbors}
\includegraphics[scale=0.6]{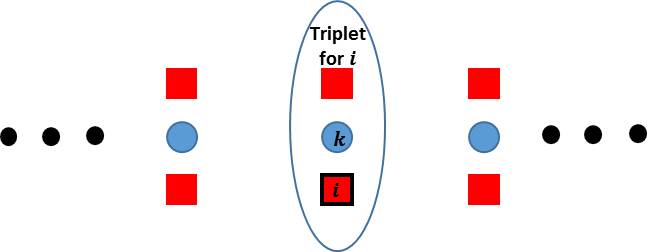}
\caption{Poorly chosen triplets~\cite{rosales2006learning} can lead to similar distortions as target neighbors. The non-class point $k$ can distort the metric, or result in infeasibility of desired metric condition~\ref{eq:metric_feasibility}. As with target neighbors, we can produce a simple metric here without using triplets.}
\label{fig:triplet}
\end{figure}
Figures~\ref{fig:target-neighbors} and~\ref{fig:triplet} illustrate the potential for distorted distance metrics when target neighbors or triplets are pre-defined. It is worth emphasizing that ~\cite{weinberger2009distance}   and ~\cite{ying2012distance} rely on Euclidean distance in their computational experiments. 
 
\subsection{Contributions}
The contributions of this work are:

\begin{enumerate}[noitemsep,nolistsep]

\item \vskip-5pt  We present a distance metric learning algorithm that is competitive with other state-of-the-art metric learning algorithms including \cite{weinberger2009distance} and \cite{davis2007information}. Moreover, our approach is more general than the aforementioned algorithms since we do not require pre-specification of target neighbors or triplets, which involves a high degree of user and data choice dependency, and hence possible errors. 

\item  Our method \textbf{provides recommendations} for new data acquisition and data quality control to improve classification performance. This is a key ``value-of-information'' criterion that can significantly improve both classification performance and computational burden by ensuring training data completeness, representativeness and economy, which are not adequately addressed in current applications of DNN and other methods which often depend on very large quantities of training data (e.g. internet cat images).

\item  We show that our underlying metric learning problem can be formulated and solved as a \textbf{mixed-integer linear optimization (MIO)} problem. To the best of our knowledge, this is the first time such a claim has been made in the metric learning arena.
Indeed, this work is also timely as it builds on what \cite{hastie2017extended} call ``exciting new work'' applying MIO to prominent machine learning problems with great success \cite{bertsimas2016best,bertsimas2017logistic,bertsimas2017sparse,bertsimas2015or,bertsimas2014least,friesen2017deep,wilson2017alamo}

\end{enumerate}


%
%
%
%
%



\section{Mixed-Integer Linear Optimization Formulations for Metric Learning}

In this section, we present mixed-integer linear formulations to determine an ``optimal'' distance metric that satisfies condition~\eqref{eq:metric_feasibility} or \eqref{eq:metric_feasibility_knn}, where optimality is governed by an appropriately chosen loss function. 
For ease of exposition, we describe the formulations for a distance metric~\eqref{eq:metric-definition} with only first- and second-order terms. Specifically, the feasible region $\mc{F}$ for the distance metric is
\begin{align} \label{set:feasible_distance_metrics}
   \mc{F} =  
    \{ &(\v{a},\v{B},\v{d}) \in \Re^D \times \Re^{D \times D} \times [0,1]^{N \times N} : \v{B} = \v{B}^{\top},\notag \\ 
   &  d_{ij} = \v{a}^{\top} |\vdelta_{ij}| + \vdelta_{ij}^{\top} \v{B}\vdelta_{ij} ~\forall i,j \in \mc{N} \\
   &  d_{ik} \geq d^{\min} ~\forall i,k \in \mc{N}: i \textrm{ and } k \textrm{ are not in same class} \} \notag  
\end{align}
Here $\vdelta_{ij} = \v{x}_i - \v{x}_j$ for all $i,j \in \mc{N}$. $d^{\min} > 0$ is a given positive parameter that prevents the degenerate metric $(\v{a},\v{B},\v{d}) = \v{0}$.
Note that $\mc{F}$ is a polyhedron, since we do not enforce $\v{B}$ to be positive semidefinite; this extension is addressed in Section~\ref{sec:extensions}.
Distances are bounded above by 1 (although any positive upper bound suffices) since our MIO methods require an upper bound.

\subsection{Metric Learning for Single Nearest Neighbor}

Consider first the task of determining an ``optimal'' distance metric satisfying condition~\eqref{eq:metric_feasibility} assuming that such a metric exists.  We relax this assumption below.
Let $\lambda_i = \min\{ \D(\v{x}_i,\v{x}_k) : k \in \bar{\mc{C}}_i \}  - \min\{ \D(\v{x}_i,\v{x}_j) : j \in \mc{C}_i \}$ be the separation between the distance to point $i$'s nearest non-class neighbor and the distance to its nearest co-class neighbor.
Since condition~\eqref{eq:metric_feasibility} seeks a distance metric such that $\lambda_i > 0$ for all $i \in \mc{N}$, we first consider the loss function
\begin{equation} \label{eq:loss_function_0}
L^0(\lambda_1,\dots,\lambda_N) = - \min\{ \lambda_i : i \in \mc{N} \}~,
\end{equation}
which rewards the minimum separation over all points. 
The following mixed-integer \textit{nonlinear} formulation attempts to minimize the loss function $L^0$, or equivalently, to maximize $\lambda$, the minimum separation over all points:  
\begin{subequations} \label{model:minlp_no_outliers}
\begin{alignat}{3}
\max_{\lambda,\v{a},\v{B},\v{d},\v{y}}~~& \lambda  \\
\st \ \  & \textstyle{\sum}_{j \in \mc{C}_i} d_{ij}y_{ij} + \lambda \leq d_{ik}, \ & \forall i \in \mc{N}, k \in \bar{\mc{C}}_i, \label{eq:separation_constraint_no_outliers} \\
	& \textstyle{\sum}_{j \in \mc{C}_i} y_{ij} = 1, & \forall i \in \mc{N}, \label{eq:bigM_select_exactly_one_neighbor_no_outliers} \\
    & y_{ij} \in \{0,1\}, & \forall i \in \mc{N}, j \in \mc{C}_i, \\
    & \lambda \in \Re, (\v{a},\v{B},\v{d}) \in \mc{F}.
\end{alignat}
\end{subequations}
Binary decision variables $y_{ij}$, which take value 1 if point $i$ is assigned to co-class point $j$ (0 otherwise), are required to select a single co-class neighbor as nearest neighbor. 
Together, constraints~\eqref{eq:separation_constraint_no_outliers} and \eqref{eq:bigM_select_exactly_one_neighbor_no_outliers} attempt to ensure that at least one co-class neighbor is closer than all other non-class neighbors to point $i$. 
Although it is helpful to think that $y_{ij}$ will take value 1 if point $j$ is chosen as the nearest co-class point to $i$ (under the resulting optimal distance metric), it is possible in an optimal solution $(\lambda^*,\v{a}^*,\v{B}^*,\v{d}^*,\v{y}^*)$ that $y_{i\hat{j}}^* = 0$ when $d_{i\hat{j}}^* < d_{ij}^*$ for all $j \in \mc{C}_i, j \neq \hat{j}$ for some point $i$.
This simply means that, for that point $i$, constraint~\eqref{eq:separation_constraint_no_outliers} is not tight in that optimal solution.  
Nevertheless, an optimal solution to formulation~\eqref{model:minlp_no_outliers} is guaranteed to find the largest minimum separation $\lambda^*$ over all points.  If $\lambda^* > 0$, then there exists a distance metric satisfying condition~\eqref{eq:metric_feasibility}. 

There are at least two deficiencies with formulation~\eqref{model:minlp_no_outliers}.
First and most important, although it is guaranteed to be feasible, it is not guaranteed to return a distance metric satifying condition~\eqref{eq:metric_feasibility}. Second, it contains bilinear terms $d_{ij}y_{ij}$ (the multiplication of two decision variables), which are non-convex and are undesirable when solving \eqref{model:minlp_no_outliers} with an off-the-shelf optimization engine. We now discuss how to overcome these two deficiencies.

Enforcing condition~\eqref{eq:metric_feasibility} to hold for all points $i \in \mc{N}$ could be too stringent.
For example, in a sparse data set, it is quite possible that certain classes are both ``islanded'' and under-represented leading to nearest co-class neighbors that are far away under most optimized metrics.
Furthermore, it is possible that a small number of points severely limit classes from being separated by a large margin.  In such cases, we may wish to identify this subset of bottleneck points as ``outliers'' and only enforce condition~\eqref{eq:metric_feasibility} for non-outliers.

To this end, let $\mc{O} \subseteq \mc{N}$ be a set of outliers and
re-define $\lambda_i$ as  
$\lambda_i = \min\{ d_{ik} : k \in \bar{\mc{C}}_i \backslash \mc{O}\}  - \min\{ d_{ij} : j \in \mc{C}_i \backslash \mc{O} \}$
for all non-outliers $i \in \mc{N} \backslash \mc{O}$.  To obtain a distance metric such that $\lambda_i > 0$ for all $i \in \mc{N} \backslash \mc{O}$, we adopt the loss function 
\begin{equation}
L(\lambda_1,\dots,\lambda_N) = \rho |\mc{O}| - \min_{i \in \mc{N} \backslash \mc{O}} \lambda_i~,
\end{equation}
which penalizes all outliers (with respect to the given distance metric) and rewards the minimum separation over all non-outliers. 
It is important to emphasize that $\rho$ is the only user-defined parameter in our approach
as a large value of $\rho$ signals that outliers are highly undesirable, whereas a small value indicates greater tolerance of outliers and more preference for larger margin.   

In order to handle outliers within an optimization framework,
we introduce binary decision variables $z_i$ that take value 1 if point $i$ is deemed an outlier; 0 otherwise.
The main interactions between the binary variables $y_{ij}$ and $z_i$ are captured through the following set: 
\begin{subequations} \label{set:feasible_region_YZ}
\begin{alignat}{3}
\mc{YZ} =  \{  (\v{y},\v{z}) : 
& \textstyle{\sum}_{j \in \mc{C}_i} y_{ij} = 1 - z_i, 
& \forall i \in \mc{N}, \label{eq:bigM_select_exactly_one_neighbor} \\
    & y_{ij} \leq 1 - z_j, & \forall i \in \mc{N},
      j \in \mc{C}_i, \label{eq:bigM_co_class_neighbor_iff_non_outlier} \\
    & y_{ij} \in \{0,1\}, 
    &\forall i \in \mc{N}, j \in \mc{C}_i, \\
    & z_{i} \in \{0,1\}, & \forall i \in \mc{N}  \}.
\end{alignat}
\end{subequations}
Constraints \eqref{eq:bigM_select_exactly_one_neighbor} ensure that each point is assigned to exactly one co-class neighbor or is deemed an outlier.
Constraints \eqref{eq:bigM_co_class_neighbor_iff_non_outlier} only allow point $i$ to be assigned to point $j$ if $j$ is not an outlier.
With these additions for outliers, we can re-formulate \eqref{model:minlp_no_outliers} as follows:
\begin{subequations} \label{model:minlp}
\begin{alignat}{3}
\max_{\lambda,\v{a},\v{B},\v{d},\v{y},\v{z}}~~& \lambda - \rho \textstyle{\sum}_{i \in \mc{N}} z_i  \\
\st \ \ & \textstyle{\sum}_{j \in \mc{C}_i} d_{ij}y_{ij} + \lambda \leq d_{ik} + M_{ik} z_k,   \forall i \in \mc{N}, k \in \bar{\mc{C}}_i \label{eq:separation_constraint_1} \\
    & \lambda \geq 0, (\v{a},\v{B},\v{d}) \in \mc{F}, (\v{y},\v{z}) \in \mc{YZ}.
\end{alignat}
\end{subequations}
Here, $M_{ik}$ is known as a ``Big M'' parameter and can be set to 1 for all $(i,k)$ pairs since all distance variables $d_{ik}$ are bounded above by 1. 
Besides the additions noted above, constraints~\eqref{eq:separation_constraint_1} are now active only if point $i$ and point $k$ are non-outliers. 

Finally, to overcome the computational issues with the nonlinear terms $d_{ij}y_{ij}$, 
we apply a standard ``trick'' in mixed-integer optimization to arrive at the following MILP formulation: 
\begin{subequations} \label{model:sub_milp_with_outlier_detection}
\begin{alignat}{2}
\max_{\substack{\lambda,\v{w},\v{y},\v{z}\\\v{a},\v{B},\v{d}}}~~& \lambda - \rho \sum_{i \in \mc{N}} z_i \\
\st ~\ \  & \sum_{j \in \mc{C}_i} w_{ij} + \lambda \leq d_{ik} + M_{ik} z_k, ~\forall i \in \mc{N}, k \in \bar{\mc{C}}_i, \label{eq:separation_constraint} \\
    & w_{ij} \leq d_{ij}, \qquad \forall i \in \mc{N}, j \in \mc{C}_i, \label{eq:w_mccormick_1} \\
    & w_{ij} \leq y_{ij}, \qquad \forall i \in \mc{N}, j \in \mc{C}_i, \label{eq:w_mccormick_2} \\
    & w_{ij} \geq d_{ij}+y_{ij}\mbox{--}1, \qquad \forall i \in \mc{N}, j \in \mc{C}_i, \label{eq:w_mccormick_3} \\
    & w_{ij} \geq 0 , \qquad \forall i \in \mc{N}, j \in \mc{C}_i, \\
    & \lambda \geq 0, (\v{a},\v{B},\v{d}) \in \mc{F}, (\v{y},\v{z}) \in \mc{YZ}. 
\end{alignat}
\end{subequations}
Constraints \eqref{eq:w_mccormick_1}-\eqref{eq:w_mccormick_3} are known as McCormick envelopes of the bilinear expression $w_{ij}=d_{ij}y_{ij}$.  
Note that, in actual implementation, the decision variables $d_{ij}$ are replaced by their definition given in \eqref{set:feasible_distance_metrics}.

\subsection{Metric Learning for $K$ Nearest Neighbors}

We next turn to the task of determining an ``optimal'' distance metric satisfying condition~\eqref{eq:metric_feasibility_knn}, i.e., the majority of the $K$ nearest neighbors are co-class points. For the remainder of this subsection, we assume $K$ is given and that each class has at least $K$ points. As above, we first present a formulation that assumes such a metric exists and then relax this assumption to allow for outliers.

Let $\mc{P} = \{ (i,j) : i \in \mc{N}, j \in \mc{N}, i \neq j \}$.
Let $\kappa_{i} = |\mc{N}_i^K(D) \cap \mc{C}_i|$ be the number of co-class points that are among point $i$'s $K$ nearest neighbors (with respect to the distance metric $\D$).
Since condition~\eqref{eq:metric_feasibility_knn} seeks a distance metric such that $\kappa_i \geq \frac{K}{2} + 1$ for all $i \in \mc{N}$, analogous to loss function~\eqref{eq:loss_function_0}, we first consider the loss function
\begin{equation} \label{eq:loss_function_0_knn}
L_K^0(\kappa_1,\dots,\kappa_N) = - \min \{ \kappa_i : i \in \mc{N} \} ~.
\end{equation} 
The following MIO formulation attempts to minimize the loss function $L^0$.  Binary decision variables $u_{ij}$, which take value 1 if point $i \in \mc{N}$ is assigned to point $j \in \mc{N}$ (0 otherwise), are required to keep track of which points are selected as the $K$ nearest nearest neighbors of each point $i \in \mc{N}$. 
\begin{subequations} \label{model:knn_milp_no_outliers}
\begin{alignat}{2}
\max_{\substack{\kappa,\vDelta,\v{u}\\\v{a},\v{B},\v{d}}}~~& \kappa \\
\st \ \ & 
\textstyle{\sum}_{ j \in \mc{C}_i } u_{ij} \geq \kappa, \qquad \forall i \in \mc{N}, \label{eq:knn_kappa_constr} \\
    & \textstyle{\sum}_{ j \in \mc{N} } u_{ij} \leq K, \qquad \forall i \in \mc{N}, \label{eq:knn_at_most_K} \\
    & d_{ij} \leq \Delta_i + M_{ij}(1-u_{ij}), ~ \forall (i,j) \in \mc{P}, \label{eq:knn_within_Delta_i} \\
   & d_{ik} \geq \Delta_i + \epsilon - M_{ik} u_{ik}, ~ \forall (i,k) \in \mc{P}:k \in \bar{C}_i, \label{eq:knn_greater_than_Delta_i} \\
    & u_{ij} \in \{0,1\}, \qquad \forall (i,j) \in \mc{P}, \\
    & \kappa \geq 0, \vDelta \geq \v{0}, (\v{a},\v{B},\v{d}) \in \mc{F}.
\end{alignat}
\end{subequations}
Constraints~\eqref{eq:knn_kappa_constr} allow us to maximize the minimum $\kappa_i$.  
Constraints~\eqref{eq:knn_at_most_K} ensure that no more than $K$ points are chosen as point $i$'s nearest neighbors.
Note that, since we are maximizing $\kappa$, constraints~\eqref{eq:knn_at_most_K} can be written with an inequality ``$\leq K$'' rather than an equality ``$= K$''. 
Constraints~\eqref{eq:knn_within_Delta_i} require point $i$'s (at most) $K$ nearest neighbors to be within a distance of $\Delta_i$, while constraints~\eqref{eq:knn_greater_than_Delta_i} enforce the opposite condition that all of point $i$'s non-nearest non-class neighbors be at least $\Delta_i + \epsilon$ units from $i$, where $\epsilon > 0$ is a user-defined parameter.
Assuming there are no co-located points (i.e., $\v{x}_i \neq \v{x}_j~\forall (i,j) \in \mc{P}$),
setting $\epsilon = \alpha \min\{||\v{x}_i -\v{x}_j|| : (i,j) \in \mc{P}\} $ for $\alpha \in (0,1]$ will guarantee that a feasible solution always exists.

An optimal solution $(\kappa^*,\vDelta^*,\v{u}^*,\v{a}^*,\v{B}^*,\v{d}^*)$ satisfies condition~\eqref{eq:metric_feasibility_knn} if $\kappa^* \geq \frac{K}{2} + 1$ since, together, constraints~\eqref{eq:knn_within_Delta_i} and \eqref{eq:knn_greater_than_Delta_i} ensure that there are at least $\kappa^*$ co-class points among the $K$ nearest neighbors of every point.
Note also that formulation~\eqref{model:knn_milp_no_outliers} allows for ``ties'' amongst co-class points, thus it could return a distance metric for which a given point's $K$ nearest neighbors are not unique.


There are at least two deficiencies with formulation~\eqref{model:knn_milp_no_outliers}. 
First, there may not exist a distance metric satisfying condition~\eqref{eq:metric_feasibility_knn}, i.e., an optimal solution to \eqref{model:knn_milp_no_outliers} could result in $\kappa^* < \frac{K}{2} + 1$.
Second, when using the loss function~\eqref{eq:loss_function_0_knn}, there may be a large number of optimal solutions to \eqref{model:knn_milp_no_outliers} with objective function value $\kappa^*$, even though we would prefer one in which the remaining $\kappa_i$'s are maximized.  
To this end, we modify the loss function to account for outliers and give weight for having more co-class points among the $K$ nearest neighbors:   
\begin{equation}
L_K(\kappa_1,\dots,\kappa_N) = \rho |\mc{O}| - \textstyle{\min}_{i \in \mc{N}} \kappa_i - W \textstyle{\sum}_{i \in \mc{N} \backslash \mc{O} } \kappa_i .
\end{equation}
Here, $W$ is a non-negative user-defined scalar that governs the preference between maximizing the minimum $\kappa_i$ and encouraging more co-class points among the $K$ nearest neighbors. 
Setting $W = (NK/2)^{-1}$ suffices to ensure that maximizing the minimum $\kappa_i$ is the dominant objective. 

This leads to the following MIO formulation:
\begin{subequations} \label{model:knn_milp_with_outlier_detection}
\begin{alignat}{2}
\max_{\substack{\kappa,\vDelta,\v{u},\v{z}\\\v{a},\v{B},\v{d}}}~~& \kappa + W \textstyle{\sum}_{i \in \mc{N}} \sum_{ j \in \mc{C}_i } u_{ij} - \rho \textstyle{\sum}_{i \in \mc{N}} z_i \\
\st \ \  
& \textstyle{\sum}_{ j \in \mc{C}_i } u_{ij} \geq \kappa, ~~\forall i \in \mc{N}, \\
    & \textstyle{\sum}_{ j \in \mc{N} } u_{ij} \leq K,
    ~~\forall i \in \mc{N}, \\
    & u_{ij} \leq 1 - z_i, \qquad \forall (i,j) \in \mc{P}, \label{eq:knn_uij_outlier_i} \\
    & u_{ij} \leq 1 - z_j, \qquad \forall (i,j) \in \mc{P}, \label{eq:knn_uij_outlier_j} \\
    & d_{ij} \leq \Delta_i + M_{ij}(1-u_{ij}), \  \forall (i,j) \in \mc{P}, \\
    & d_{ik} \geq \Delta_i + \epsilon - M_{ik} (u_{ik} + z_i + z_k), \notag\\
    & \qquad \qquad \qquad \forall (i,k) \in \mc{P}:k \in \bar{C}_i, \label{eq:knn_greater_than_Delta_i_outliers} \\
    & u_{ij} \in \{0,1\}, \qquad \forall (i,j) \in \mc{P}, \\
    & z_{i} \in \{0,1\}, ~ \qquad \forall i \in \mc{N}, \\
    & \kappa \geq 0, \vDelta \geq \v{0}, (\v{a},\v{B},\v{d}) \in \mc{F}.
\end{alignat}
\end{subequations}
The objective function is the negative loss function $-L_K(\kappa_1,\dots,\kappa_N)$, where $\sum_{i \in \mc{N}} \sum_{ j \in \mc{C}_i } u_{ij}$ plays the role of $\kappa_i$.
Constraints~\eqref{eq:knn_uij_outlier_i} and \eqref{eq:knn_uij_outlier_j} ensure that point $j$ is not chosen as one of point $i$'s $K$ nearest neighbors if either $i$ or $j$ is deemed an outlier.
All other constraints resemble those of formulation~\eqref{model:knn_milp_no_outliers} except perhaps with modifications to account for outliers.

Note that the $K$ nearest neighbor approach requires more binary variables than the single nearest neighbor formulation. 
In particular, the former requires $(N-1)^2$ binary variables $u_{ij}$ for all$(i,j) \in \mc{P}$, whereas the latter requires far fewer since $y_{ij}$ is only defined for all $(i,j): i \in \mc{N}, j \in \mc{C}_i$.

\subsection{Extensions} \label{sec:extensions}

Thus far, positive semidefiniteness of the $\v{B}$ matrix is not enforced.  Although there is evidence that positive semidefiniteness is a desirable attribute and may improve interpretability of the resulting metric, it may not be as essential as others have described.  Indeed, several highly touted state-of-the-art approaches do not enforce positive semidefiniteness, e.g. NCA \cite{goldberger2004nca} and deep neural nets, yet are still garnering considerable attention.  Nevertheless, if positive semidefiniteness is strongly desired, we could trivially extend formulation~\eqref{model:sub_milp_with_outlier_detection} to include diagonal dominance constraints as done in \cite{rosales2006learning}.  Although such constraints would not allow for fully general psd matrices to be generated, they would keep the formulation mixed-integer linear.  On the other hand, if all psd matrices are desired, we would need to adopt a more sophisticated approach as in \cite{weinberger2009distance}. 

While we have thus far extolled the fact that our approach does not rely on a priori information, our MILP formulations can easily accommodate user-provided target neighbors and similarity/dissimilarity pairs.  Indeed, linear constraints, like those used in \cite{davis2007information} and Shavel-Shwartz \cite{shalev2004online}, that require user-specified similar points to be closer than dissimilar points are simple to incorporate.    

\section{Active Learning for Targeted Data Acquisition Using Boundary and Outlier Identification}
Our algorithm is particularly suited for continuous, online data acquisition aimed at converging to an optimal metric with the smallest amount of data. This type of {\em active learning} approach is critical to maintaining economy, completeness and representativeness in data selection. To the best of our knowledge, this work is the first to address the connection between metric learning and active data selection.  A similar approach is applicable to alternative metric and other learning paradigms including LMNN, ITML, DNN, and SVM. For example, for DNN, the current approach often involves using very large quantities of training data, causing high computational burdens and convergence issues.  Prioritization of data based on empirical boundary point and outlier determination could significantly improve performance.

\begin{figure}
\centering
\includegraphics[scale=0.85]{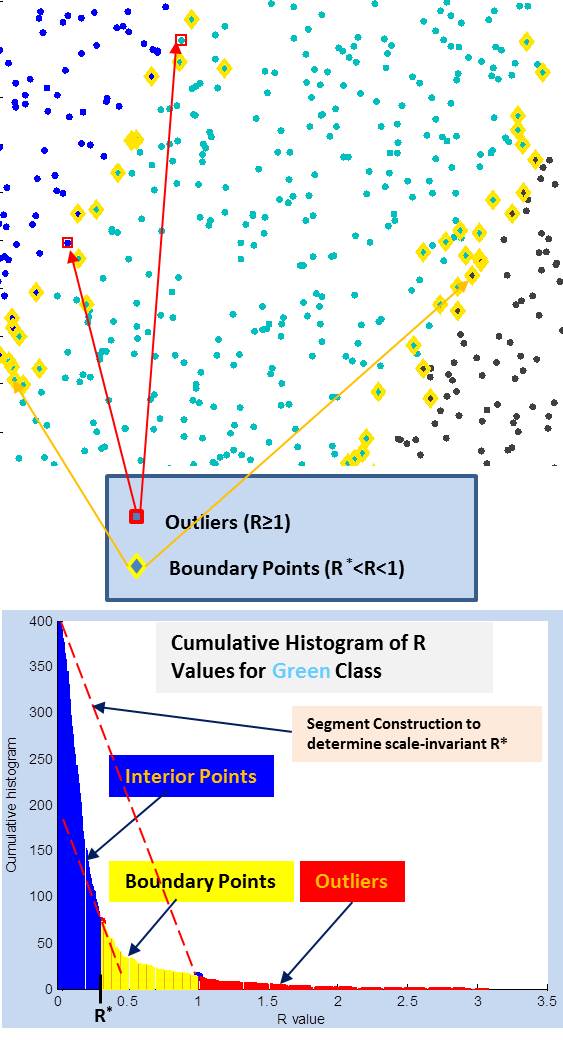}
\caption{Histogramming the data by $R$-ratio~(\ref{eq:R-ratio}) allows empirical identification of outlier and boundary points for each class separately using the histogram as shown on the bottom figure. The scatter plot shows the resulting boundary and outlier points. Active learning involves selectively acquiring more data at these outliers and boundaries (as opposed to the interior), in sorted order of $R$-value for better data economy and faster convergence to true metric.}
\label{fig:Tradeoff_curve_figure}
\end{figure}

\begin{figure}
 
 \centering
\includegraphics[width=3.2in]{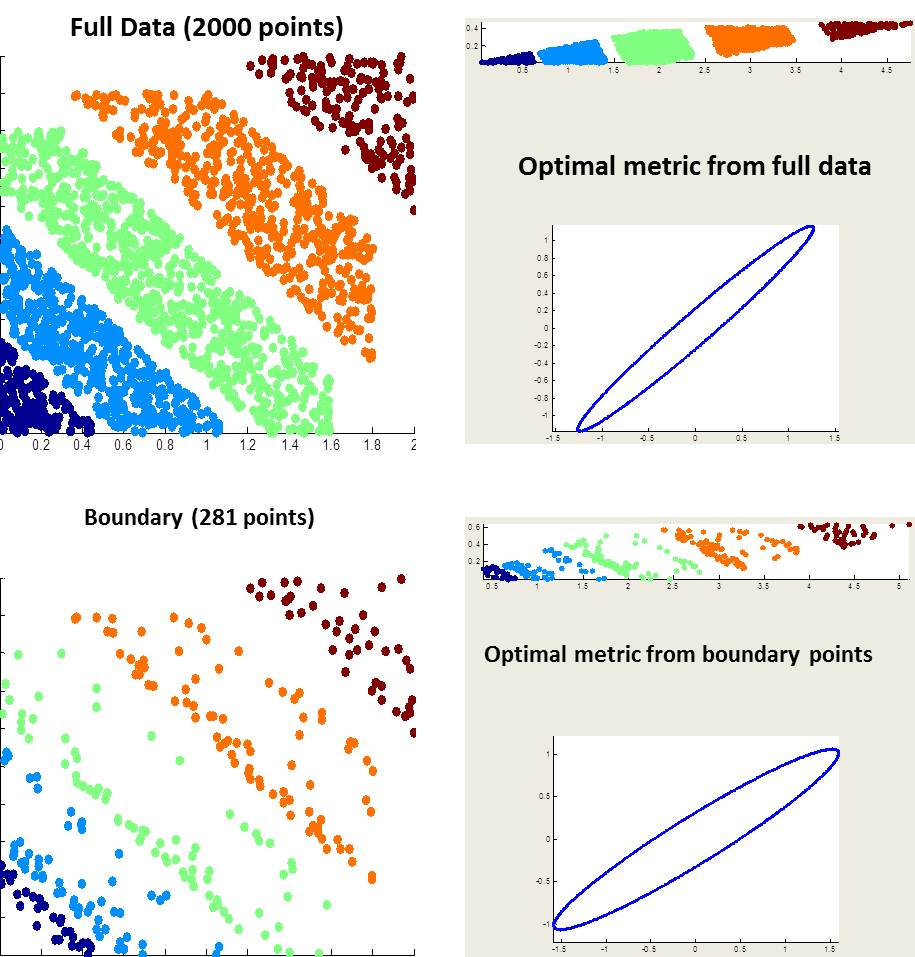}
\caption{The closest co-class to non-class neighbor ratios $R_i$ can reveal the class boundaries in combination with a trade-off curve as shown in the top right Figure.
 Note that the metric inferred from all class points (top ellipse) is very close to the metric from the much smaller set of boundary points only (bottom ellipse). The vertically squished images (top right and center right) are scatter plots of the points in the transformed coordinates from the optimal metric in each case. This example demonstrates our algorithm's ability to implement data economy. As mentioned in the text, the boundary points also suggest the most desirable regions for further data acquisition to improve classification results.}
\label{fig:boundary-points}
 
\includegraphics[width=3.2in]{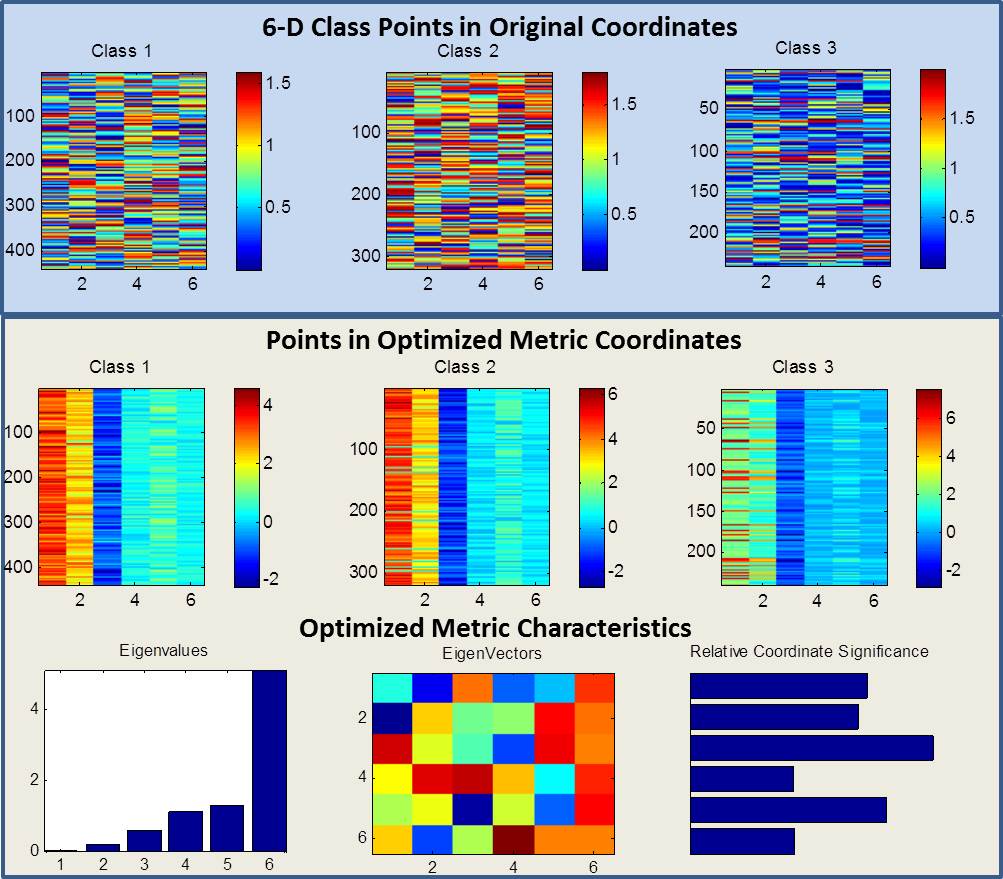}
\caption{Synthetic example of a 6-D dataset with 3 classes. The pattern of the data within classes, as well as the differences between classes, can be characterized and visualized much more easily on the scaled coordinates from our optimized metric, as shown in the figures titled 'Points in Optimized Metric Coordinates'. The occurrence of vertical stripes in each class visualization implies that each class has relatively uniform coordinate values in the transformed metric space.}
\label{fig:6-d-example}

\end{figure}

Figure~\ref{fig:Tradeoff_curve_figure} shows a summary of our approach for the $K=1$ nearest neighbor case\footnote{Both the formula for $R$ and the data selection procedure can be extended for $K>1$ with small modifications.}.  It involves the following steps:
\begin{itemize}[noitemsep,topsep=0pt]
\item \vskip-5pt {Compute the $R$-ratio~(\ref{eq:R-ratio}) for all current points.}
\item{Compute the cumulative histogram of each class separately (bottom of Figure~\ref{fig:Tradeoff_curve_figure}).}
\item{Points with $R \geq 1$ are potential outliers, while points between the "knee-point" $R^*$ and 1 are designated as {\em boundary points}. Points with $R \leq R^*$ are the {\em interior points}. \footnote{Note that the parallel line construction shown in the figure typically results in $R^*<1$ in practice as the cumulative histogram is typically convex and monotonically decreasing with high enough sampling density, which results in many more interior points than boundary points. However, when this is not the case, we suggest retaining all data points for the class, i.e. $R^*=0$ as this condition is an indication of insufficient sampling and more samples will be needed before discriminating data selection is possible.}}
\item{Our recommendation for new data acquisition is to selectively acquire additional data at the outliers and boundary points, prioritized in decreased order of $R$.}
\end{itemize}
The above described procedure is motivated by the self-evident observation that interior points typically have many co-class neighbors closer than their nearest non-class neighbor, and are hence less likely to be misclassified, while the reverse is true of the boundary points. Empirical experimental validation of this observation is shown in Figure~\ref{fig:boundary-points}, where the metric learned from all the data is very close to the metric learned from the boundary points only.
\section{Numerical Experiments }

In this section, we demonstrate the performance of our algorithm on real and synthetic datasets. 
In all experiments presented here, we restrict our attention to the search of optimal  Mahalanobis distances for which the leading method in the literature is LMNN \cite{weinberger2009distance}\footnote{We have used LMNN 3.0.0 available at \url{http://www.cs.cornell.edu/~kilian/code/code.html}.} against which we compare our approach.



\subsection{Synthetic Data}
\begin{figure} 
\centering
\includegraphics[width=3.2in]{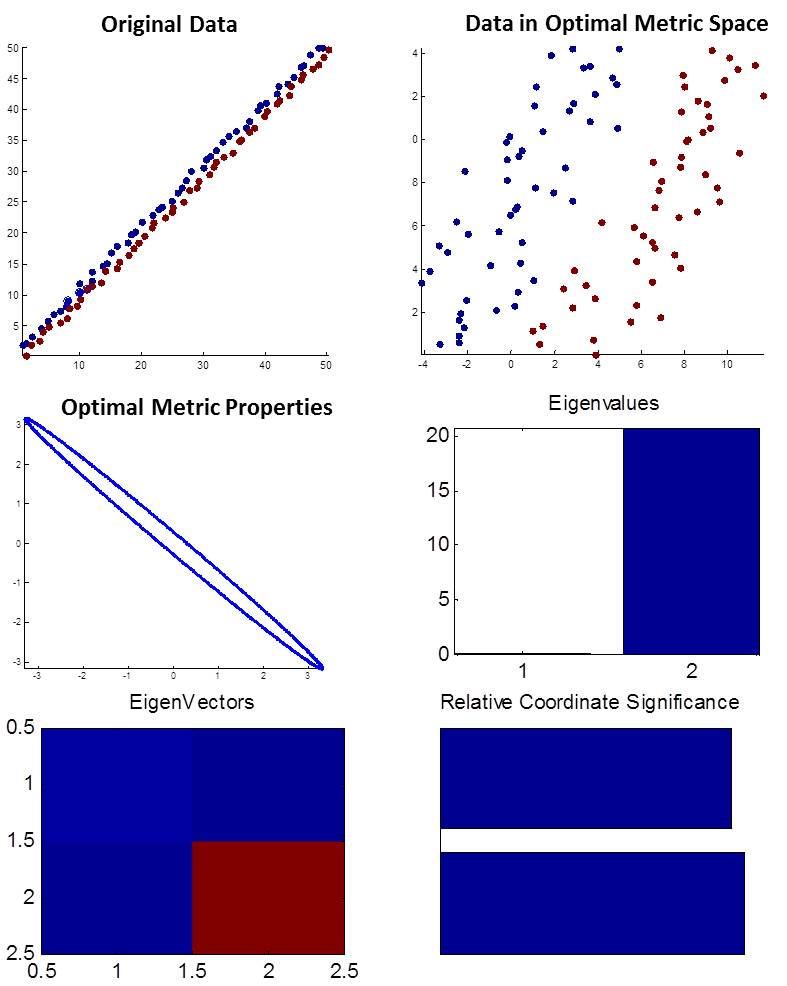}
\caption{Results on a closely spaced pair of noisy linear class examples show a metric highly weighted orthogonal to the line as expected. This example could be challenging for existing methods like LMNN if target neighbors are chosen using Euclidean metric at the start as normally suggested.}
\label{fig:lines}


\includegraphics[scale=.15]{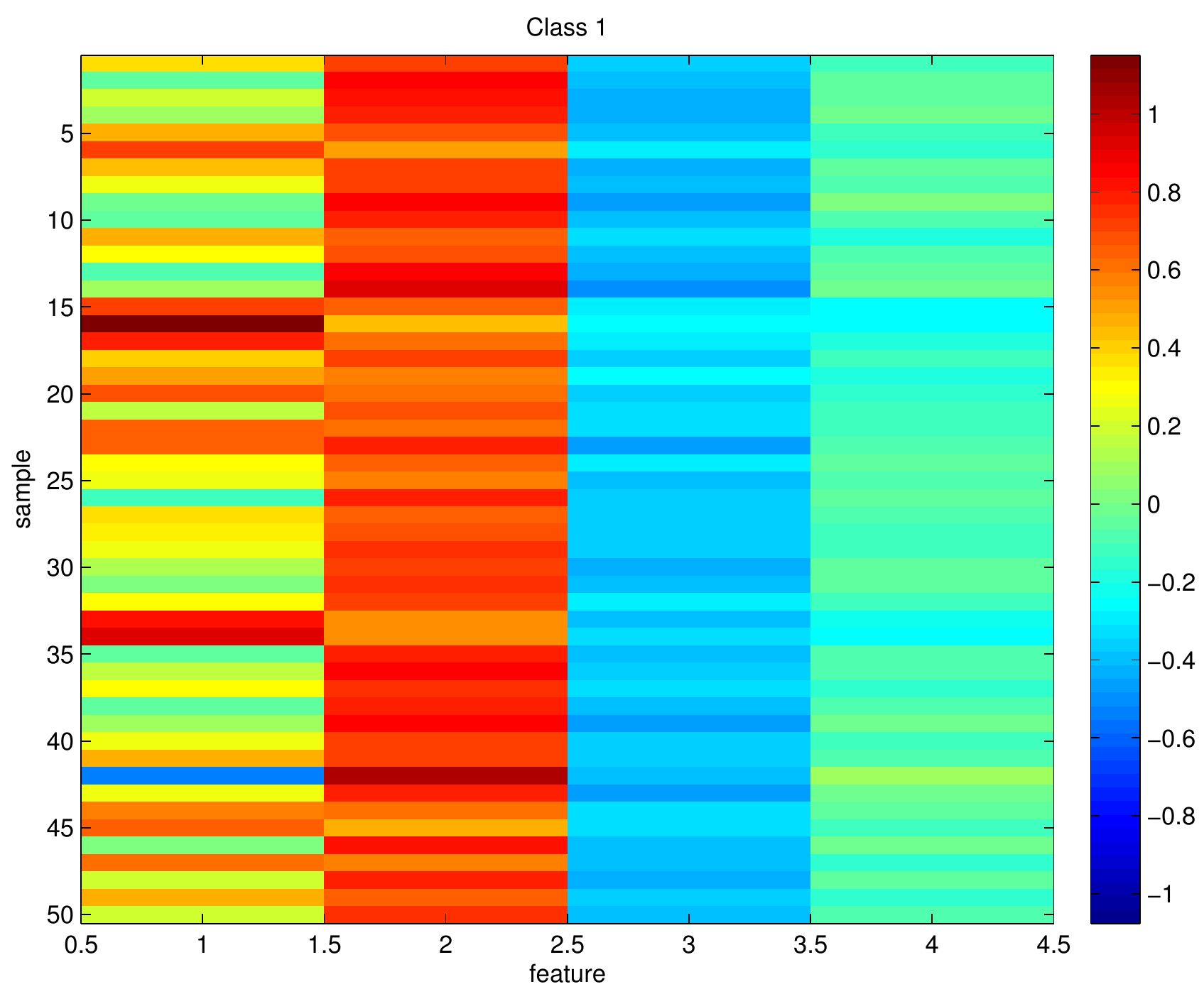}
\includegraphics[scale=.15]{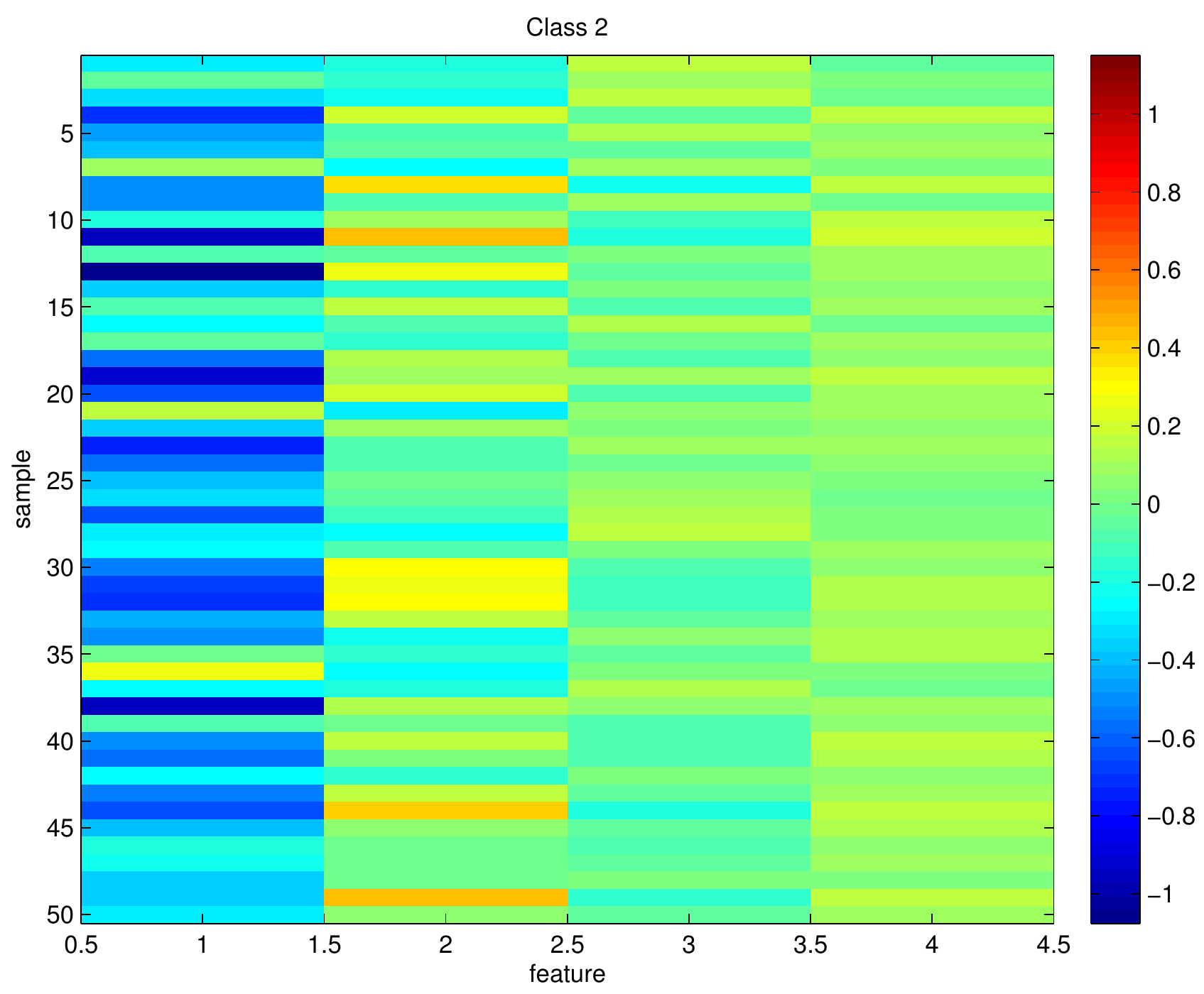}
\includegraphics[scale=.15]{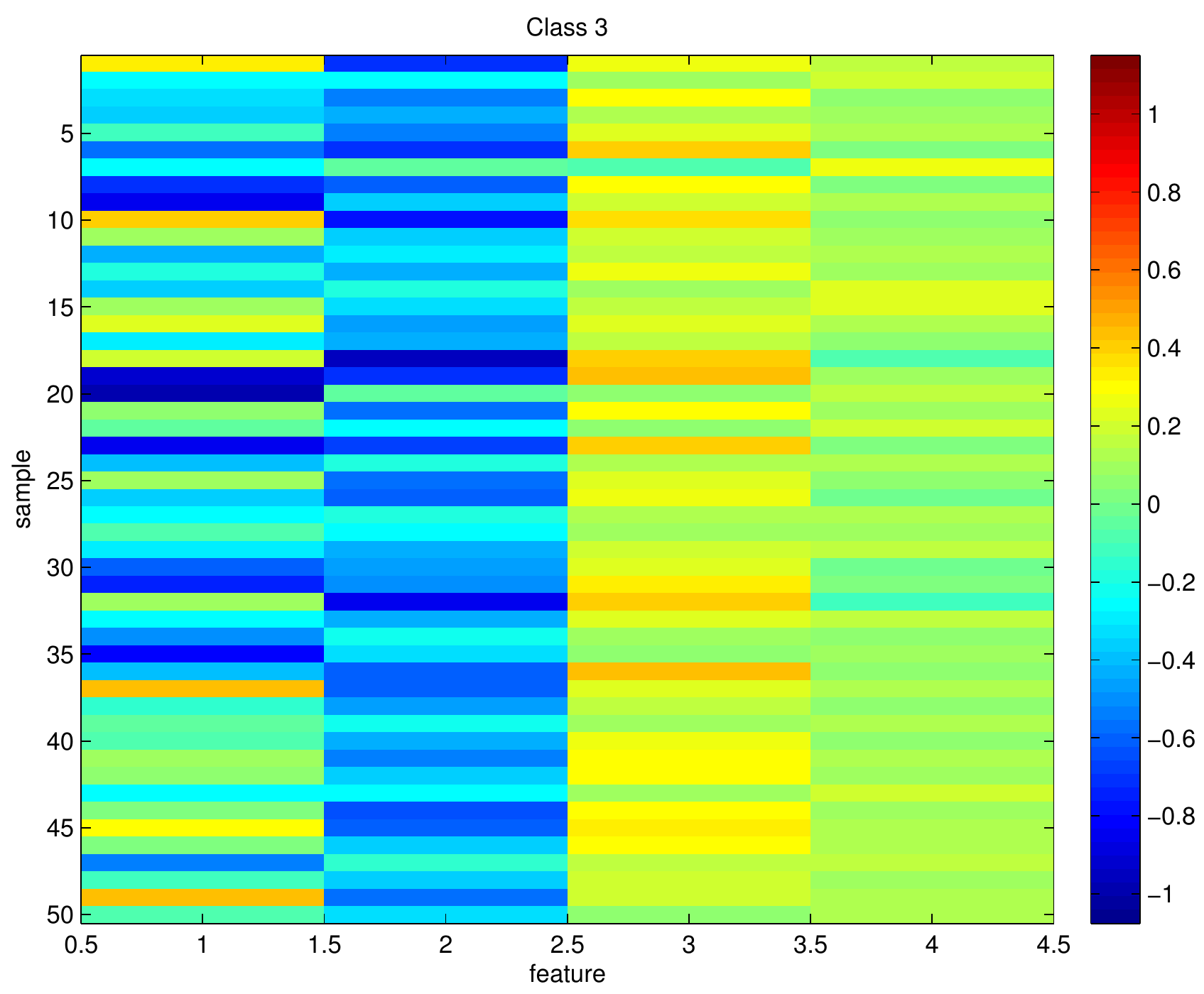}

\includegraphics[scale=.24]{distrib_Iris_0_0-eps-converted-to.pdf}
\includegraphics[scale=.24]{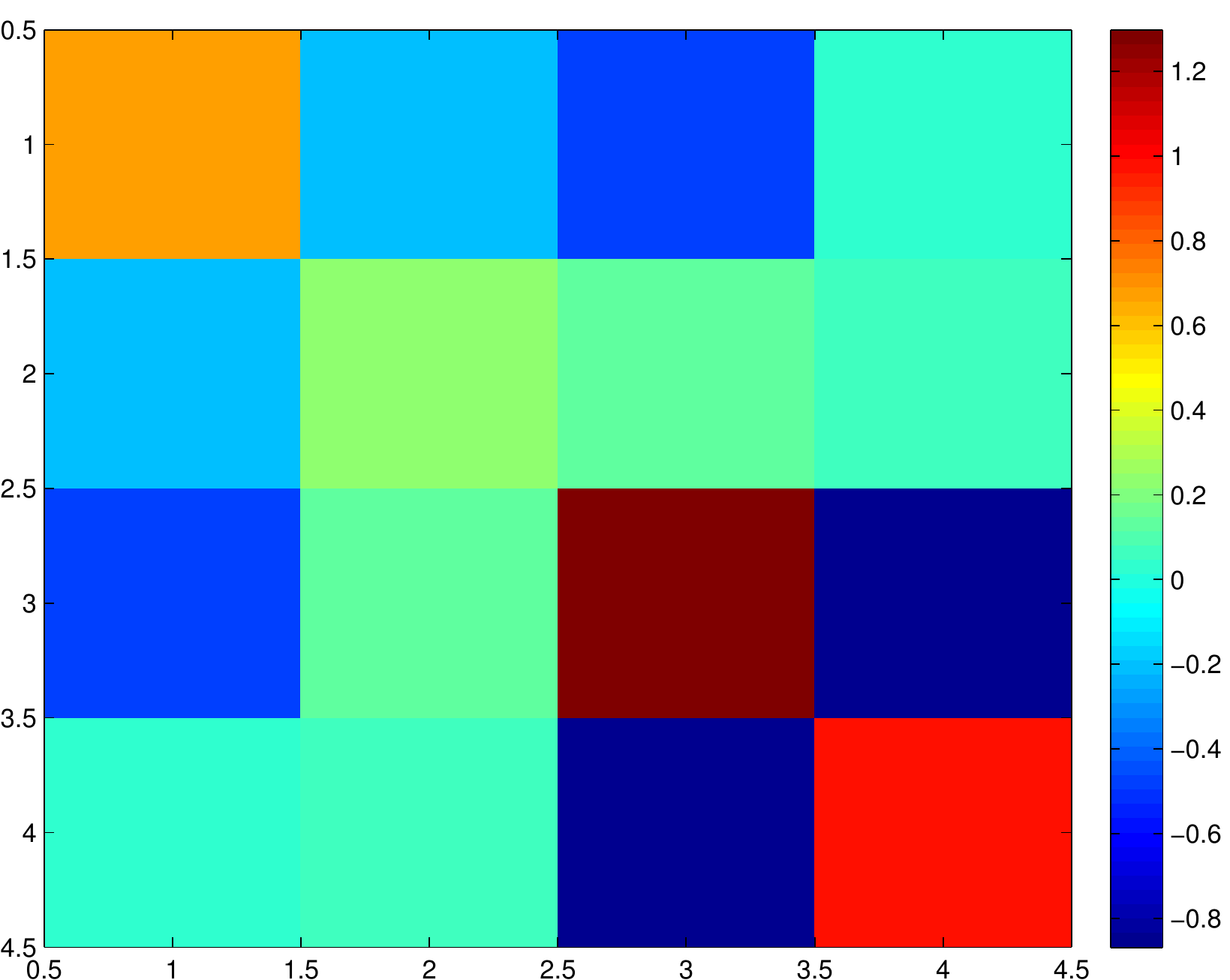}

\caption{Results on 4D IRIS dataset.
This example illustrates visualization of high-dimensional data using our optimized metric.
As shown in the synthetic example earlier, 
note that representation of class features (top row) on the scaled coordinates reveals unique signatures of each class as vertical streaks indicating relatively uniform feature values within each class.}
\label{fig:IRIS}

\end{figure}

The purpose of this section is to demonstrate the characteristics of the proposed algorithm on synthetic data sets designed to illustrate complex structure, including classes with multiple ``domains/islands'' and non-convex class shapes.  Further, we show how the boundary points of classes can be inferred from the metric for data economy and recommendations of further data acquisition.  Also, in the Appendix, we demonstrate that our method produces smoother class boundaries that standard metrics, thus reducing errors due to over-fitting. 
Finally, we provide a higher dimensional example to illustrate how classes are better represented and visualized using the optimal metric.
\\
{\bf Synthetic Data: No Outliers}
We start with 2D  examples that permit easy visualization and intuitive geometric explanation. In these examples, we assume that the data has no outliers, and the sampling is sufficiently large for robust classification. 
The Mahalanobis metric allows for simple and direct interpretation of the metric using eigenvalues and eigenvectors. 
We  use the co-class to non-class distance ratio of \eqref{eq:R-ratio}
to define
\textit{relative coordinate significance} for each coordinate as the eigenvalue-weighted absolute sum of the corresponding components of the eigenvectors, i.e., if the Mahalanobis matrix has (eigenvalue, eigenvector) pairs $(\lambda_i , \mathbf{v}_i), i=1,\ldots,D$, with each eigenvector $\mathbf{v}_i \equiv \{v_{ij}: j=1,\ldots,D\}$, then the relative significance of each coordinate dimension $j$ is defined as
\begin{equation}
W_j \equiv \textstyle{\sum}_{i=1}^D \lambda_i |v_{ij}|.
\label{eq:coordinate-significance}
\end{equation}
Figures \ref{fig:boundary-points}, \ref{fig:lines}, and \ref{fig:6-d-example} show how these concepts can be used, in 2D as well as higher dimensions, to produce more intuitive and interpretable metrics even when the sample points are restricted primarily to class boundaries. We deal with identification and exclusion of outliers in the Appendix. Now we show some applications of these methods to real data along with comparisons to competing alternatives.

\begin{figure}[h!]
 \centering

A \includegraphics[scale=.08,trim=2cm 2cm 2cm 2cm, clip=true]{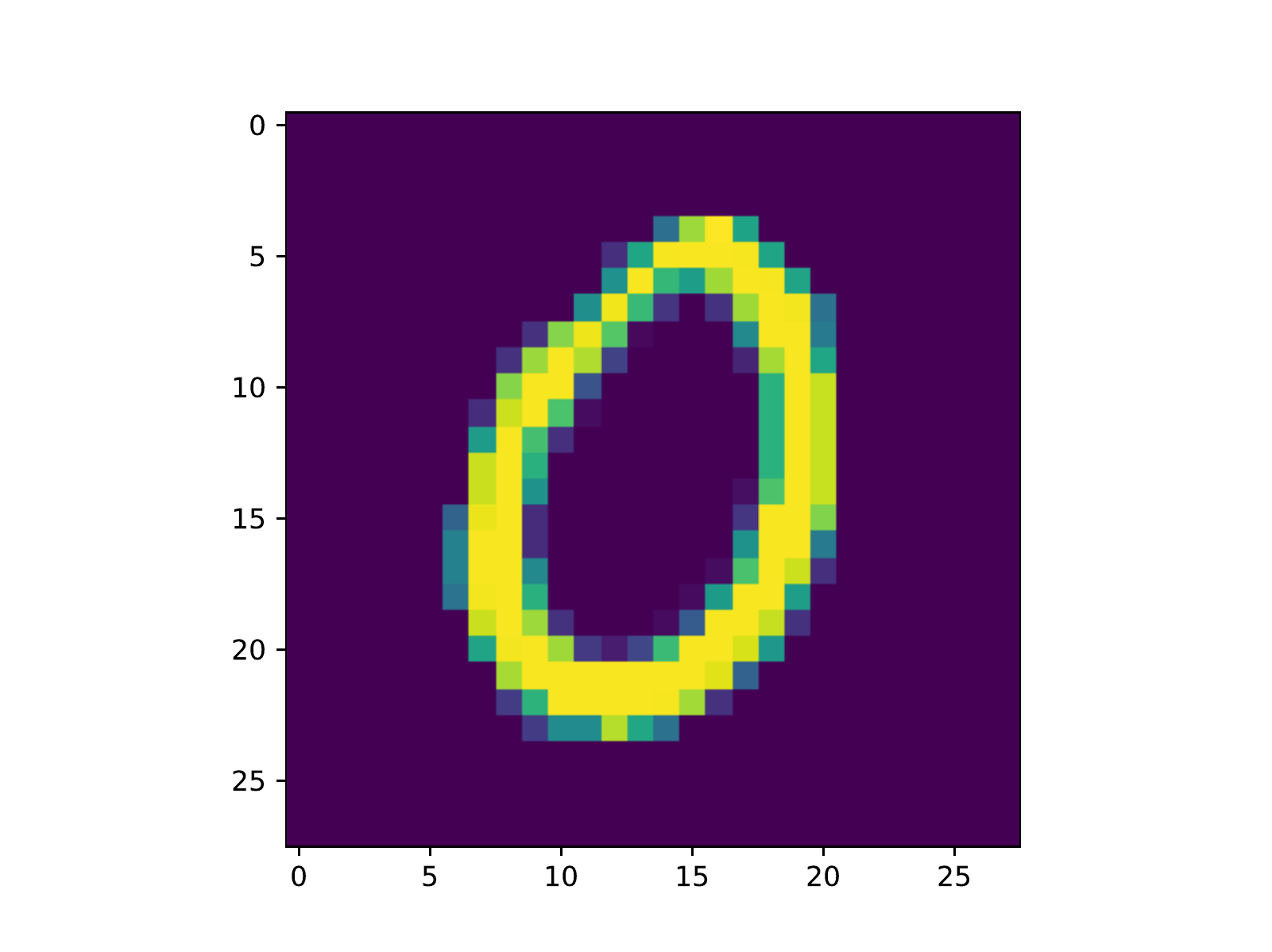}
\includegraphics[scale=.08,trim=2cm 2cm 2cm 2cm, clip=true]{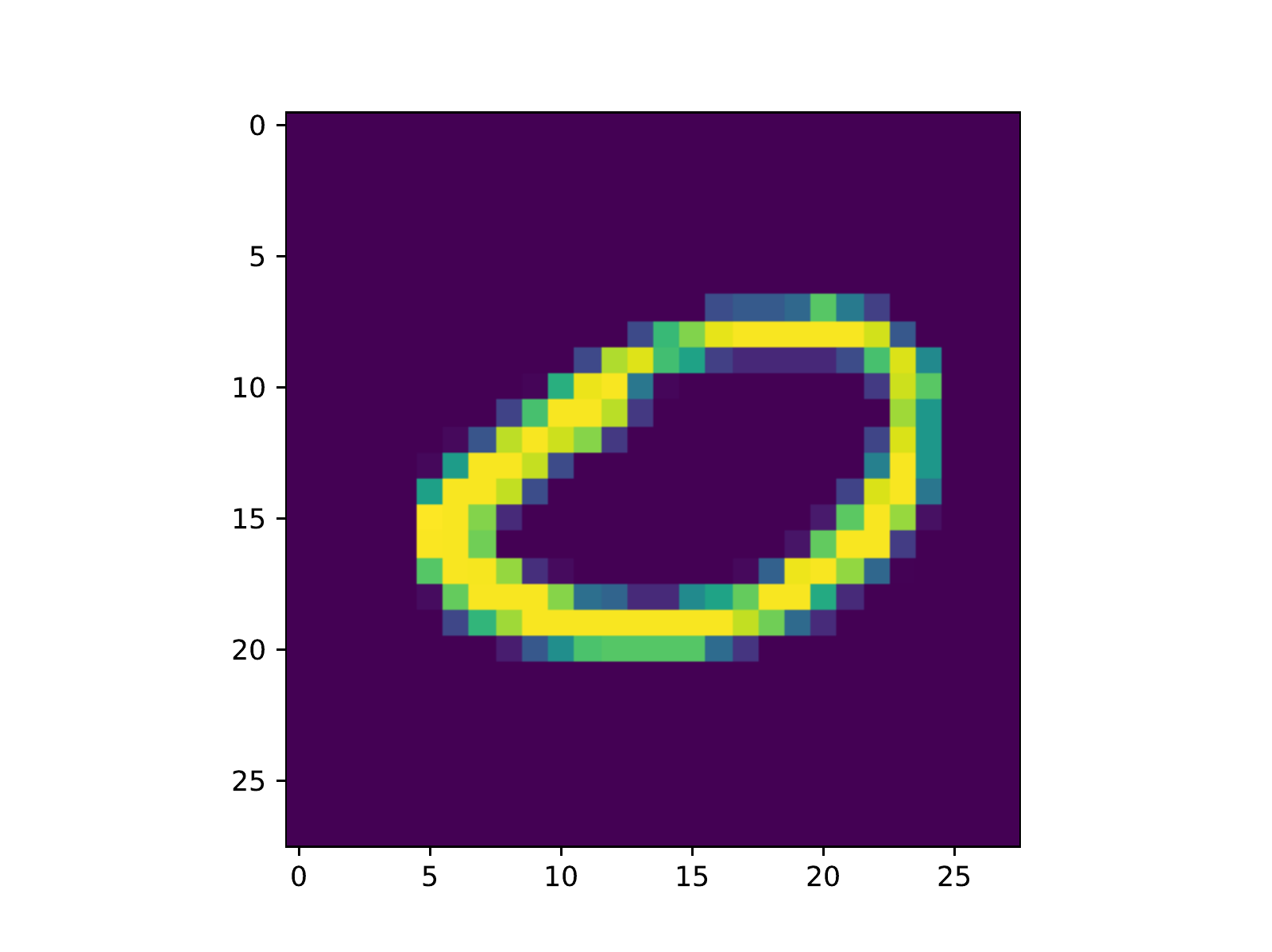}
\includegraphics[scale=.08,trim=2cm 2cm 2cm 2cm, clip=true]{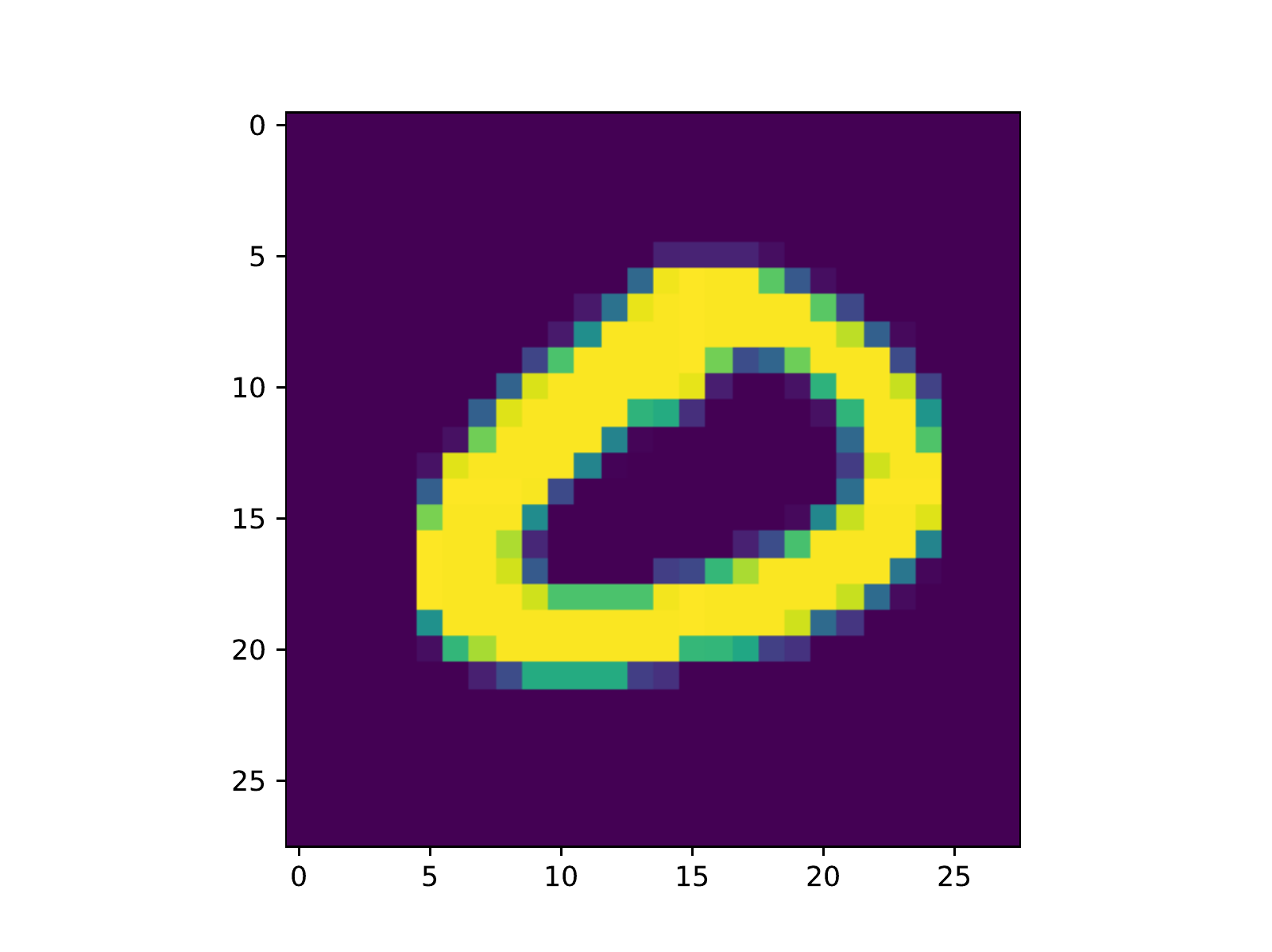}
\includegraphics[scale=.08,trim=2cm 2cm 2cm 2cm, clip=true]{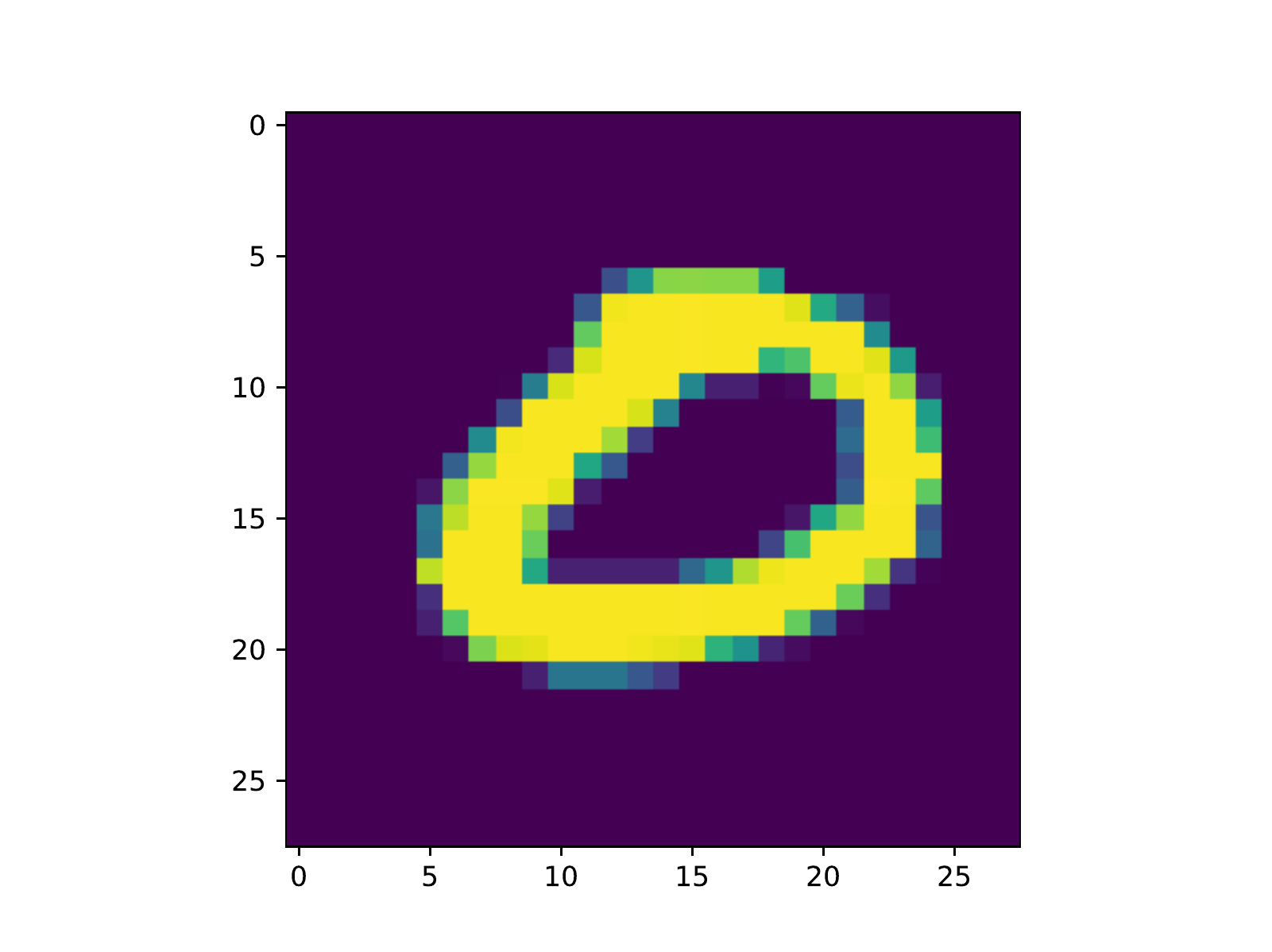}
\includegraphics[scale=.08,trim=2cm 2cm 2cm 2cm, clip=true]{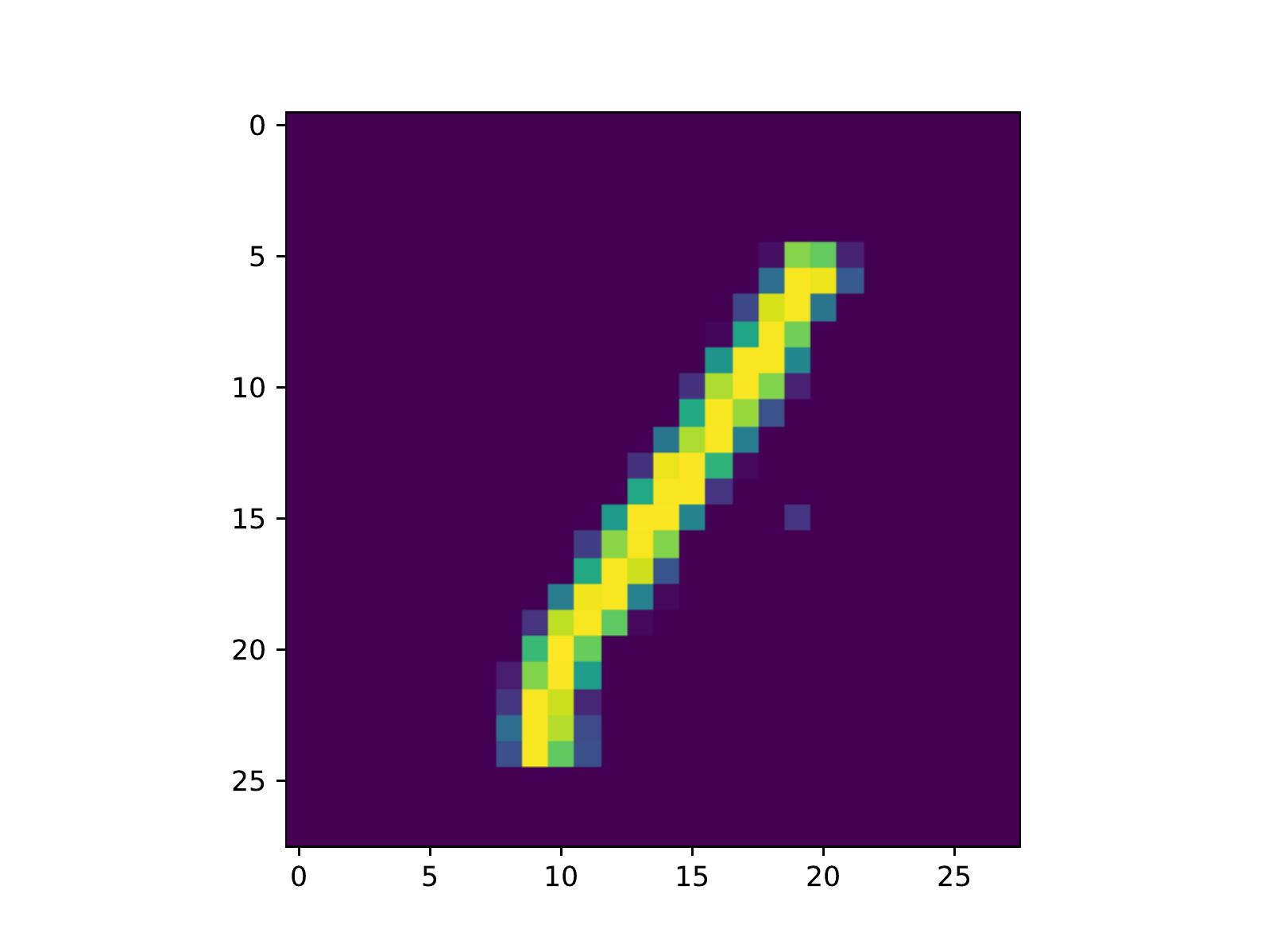}
\includegraphics[scale=.08,trim=2cm 2cm 2cm 2cm, clip=true]{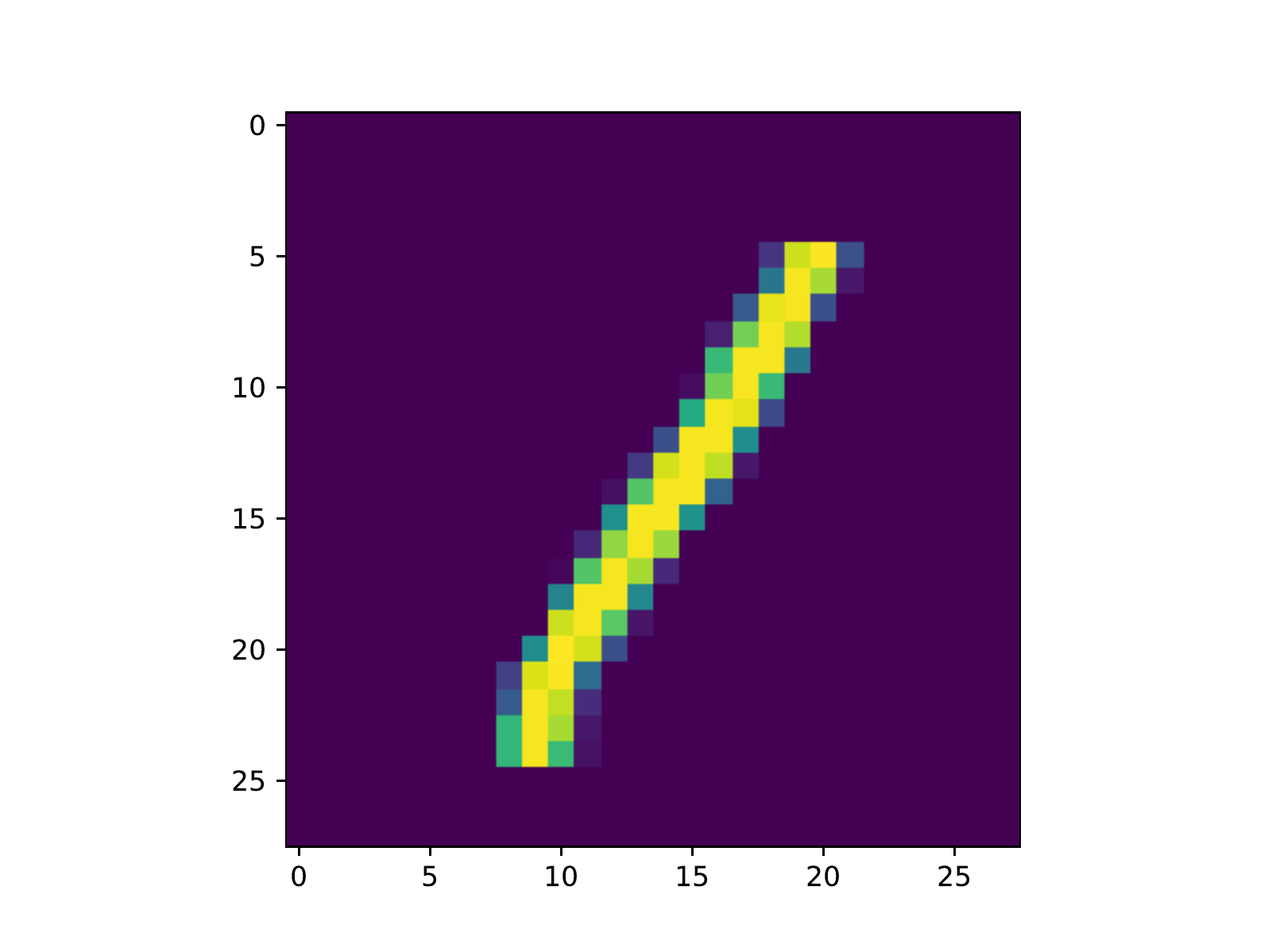}
\includegraphics[scale=.08,trim=2cm 2cm 2cm 2cm, clip=true]{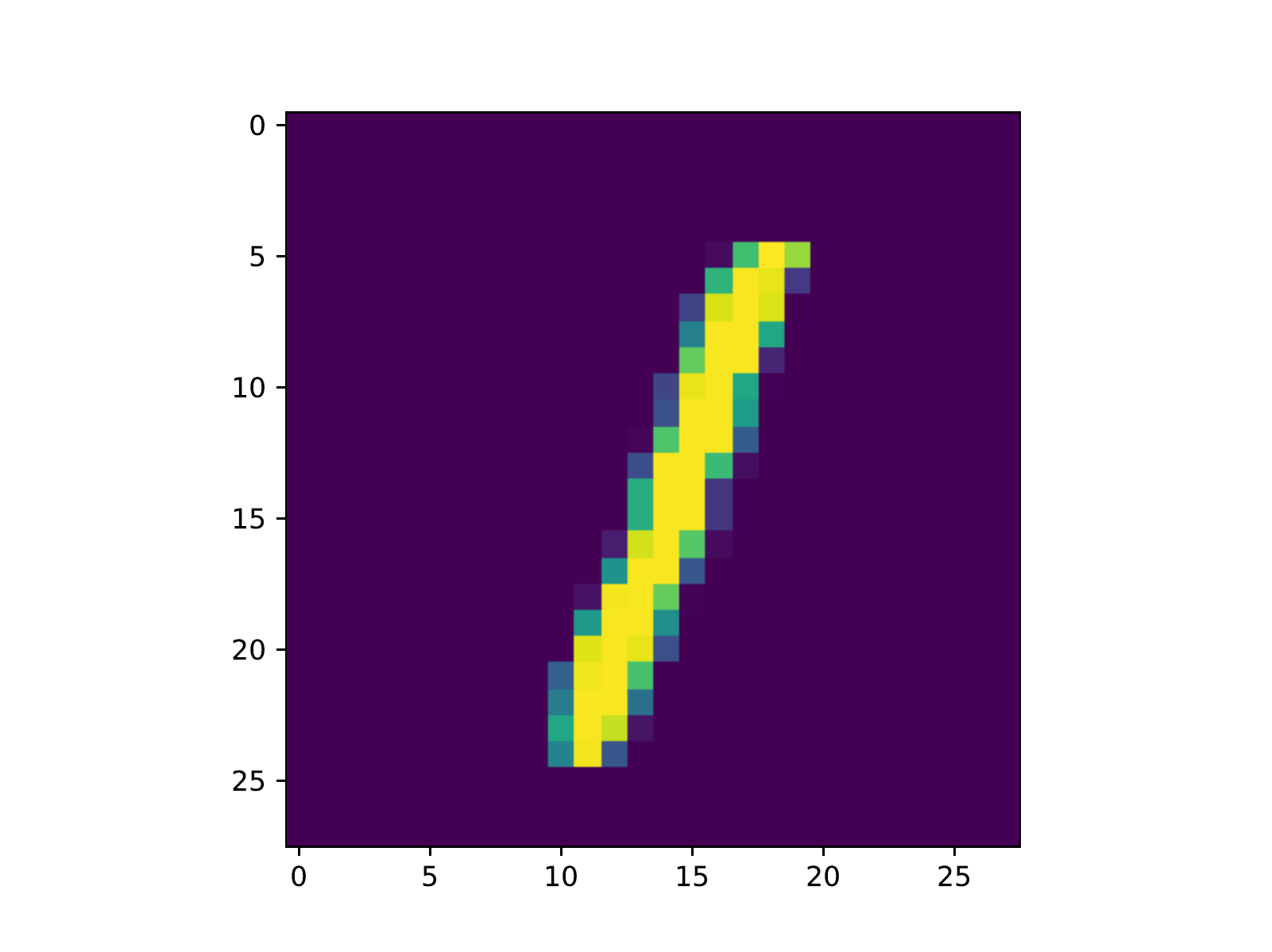}

B \includegraphics[scale=.08,trim=2cm 2cm 2cm 2cm, clip=true]{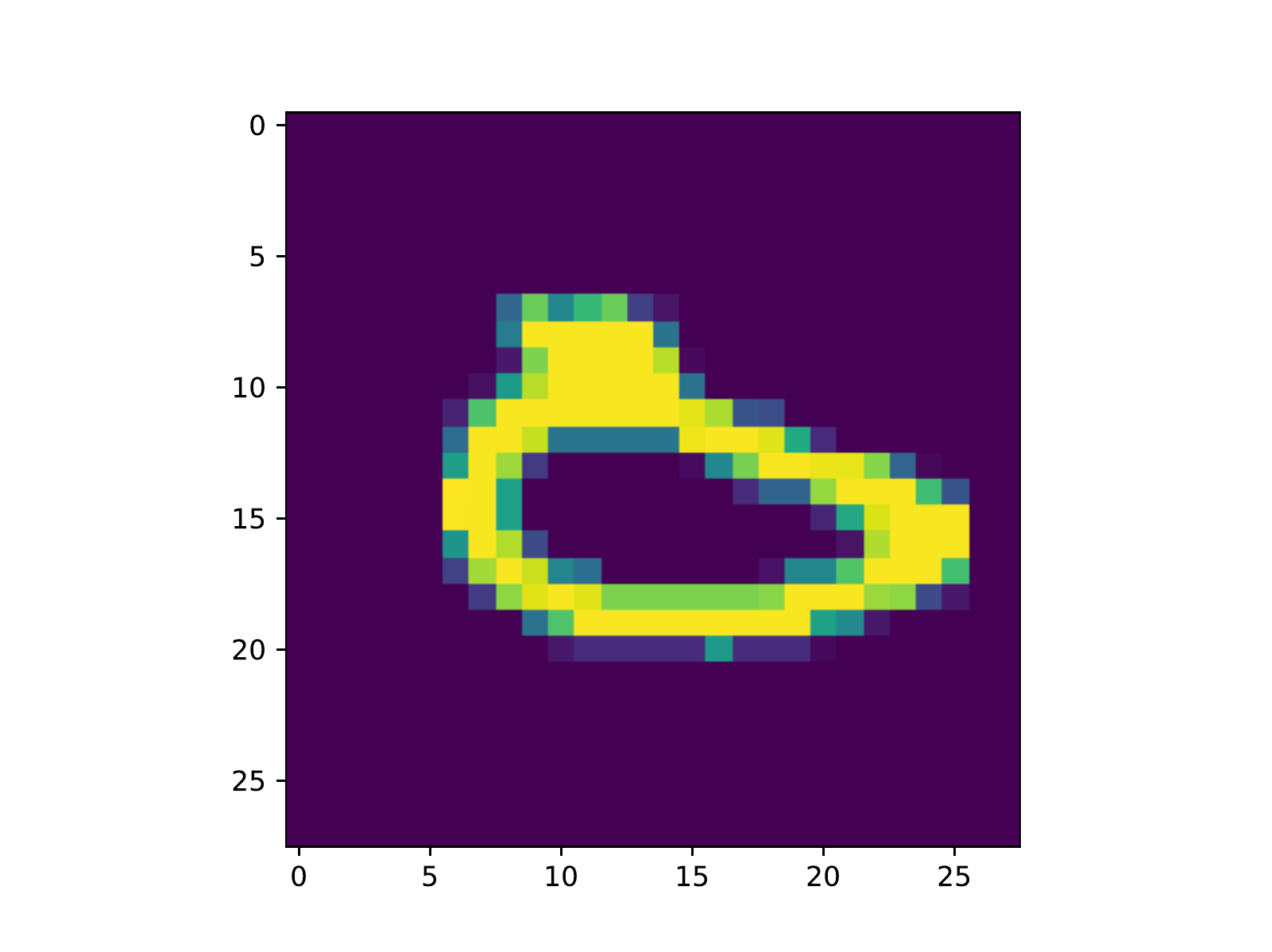}
\includegraphics[scale=.08,trim=2cm 2cm 2cm 2cm, clip=true]{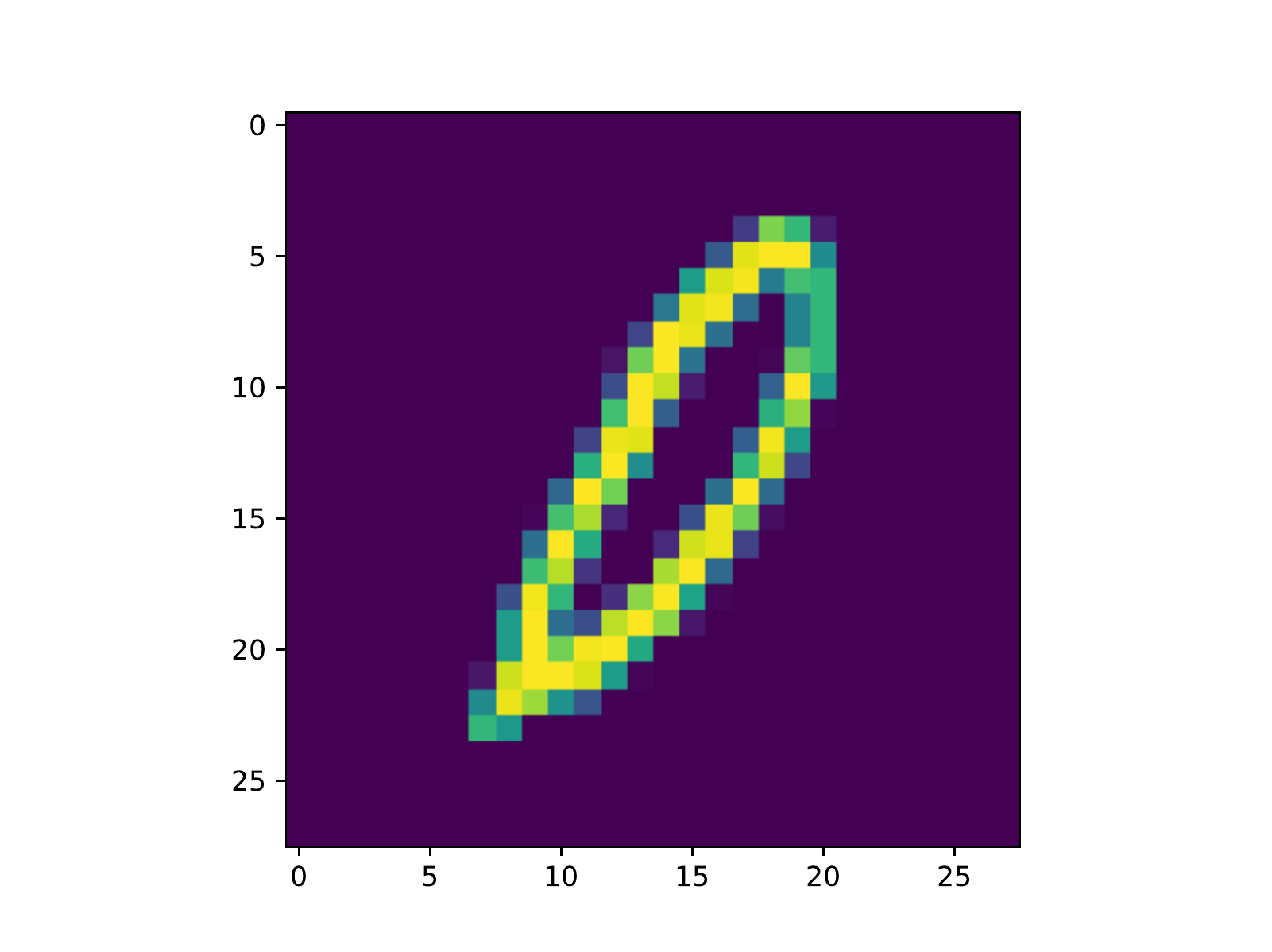}
\includegraphics[scale=.08,trim=2cm 2cm 2cm 2cm, clip=true]{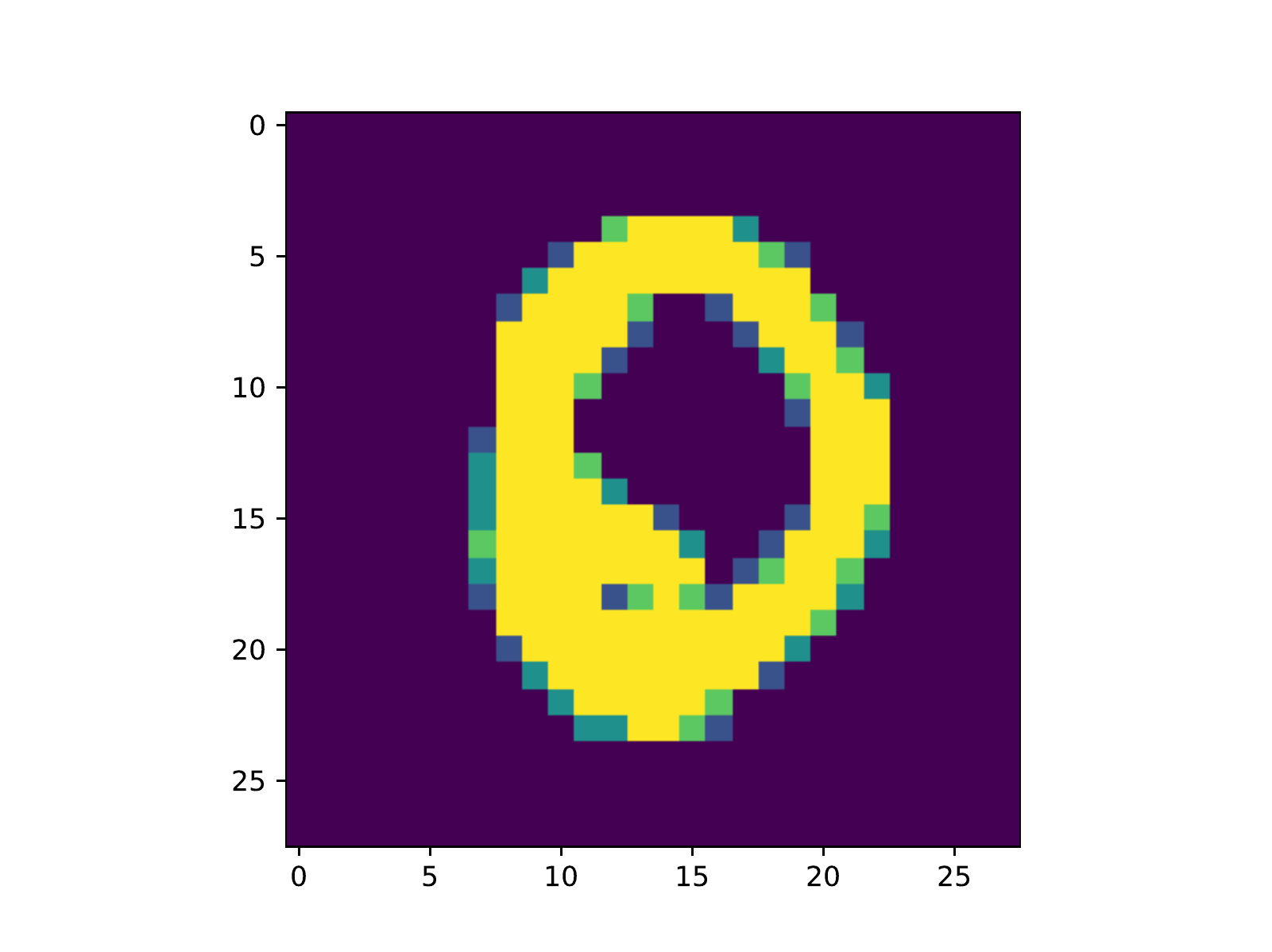}
\includegraphics[scale=.08,trim=2cm 2cm 2cm 2cm, clip=true]{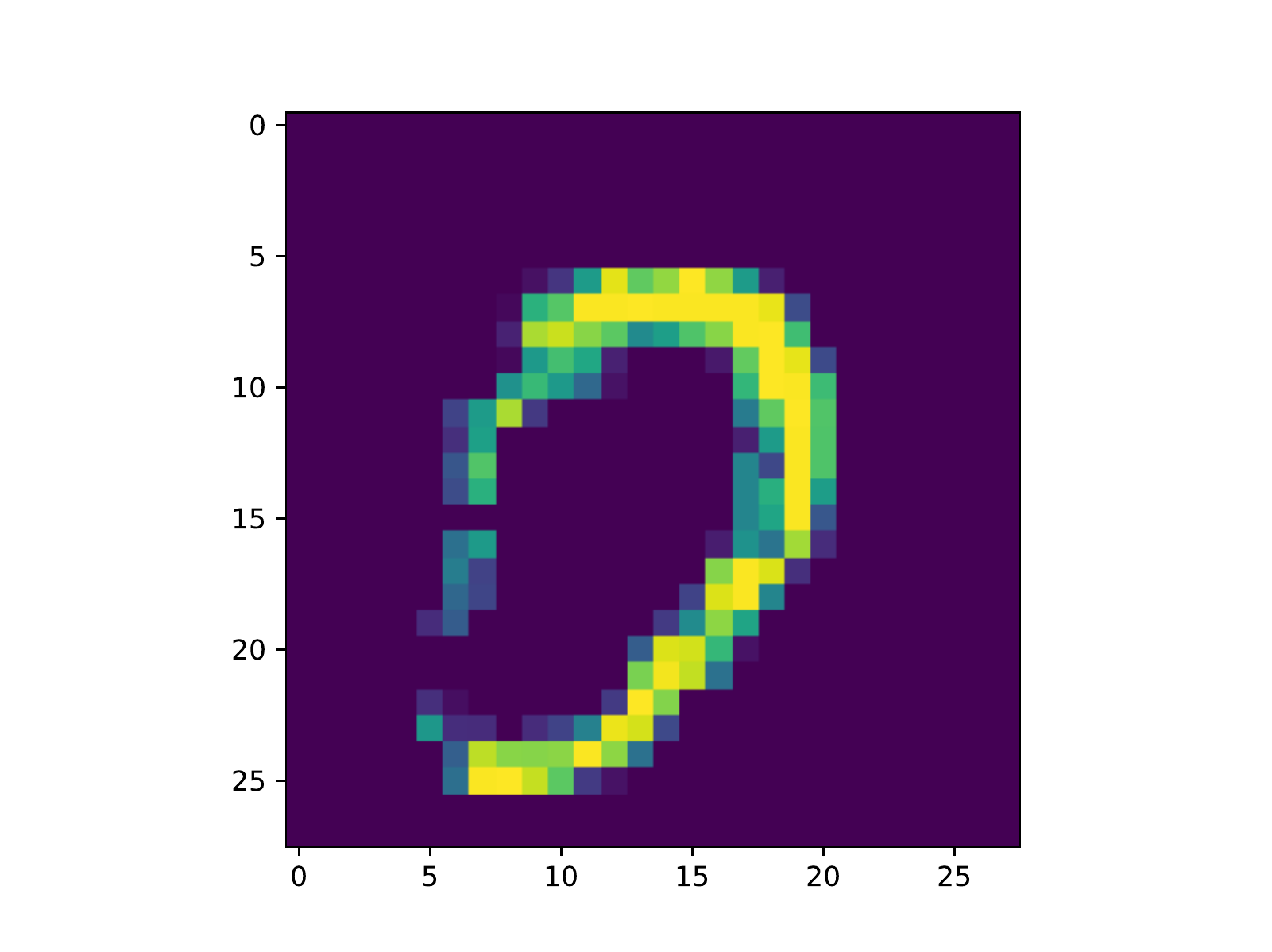}
\includegraphics[scale=.08,trim=2cm 2cm 2cm 2cm, clip=true]{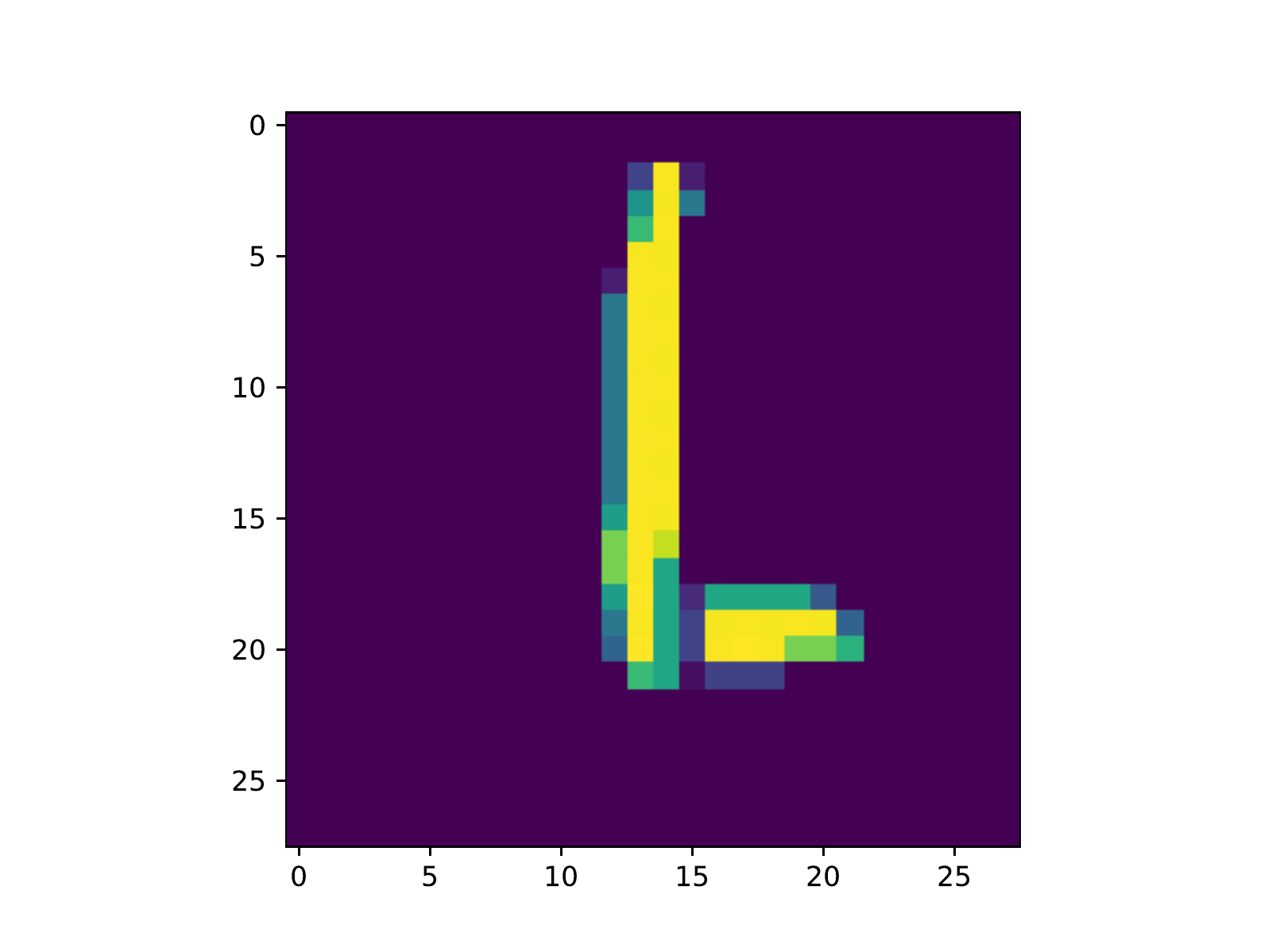}
\includegraphics[scale=.08,trim=2cm 2cm 2cm 2cm, clip=true]{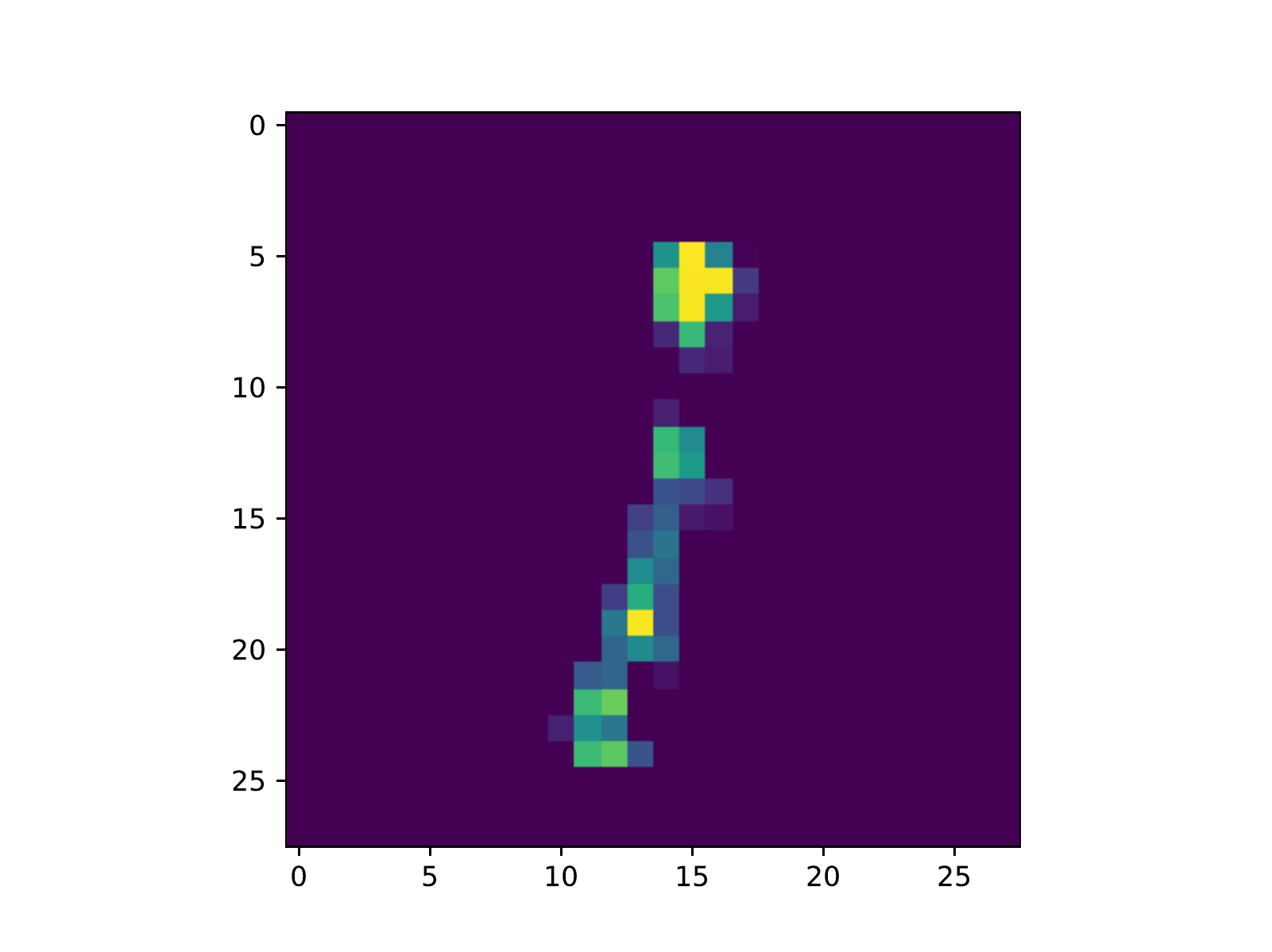}
\includegraphics[scale=.08,trim=2cm 2cm 2cm 2cm, clip=true]{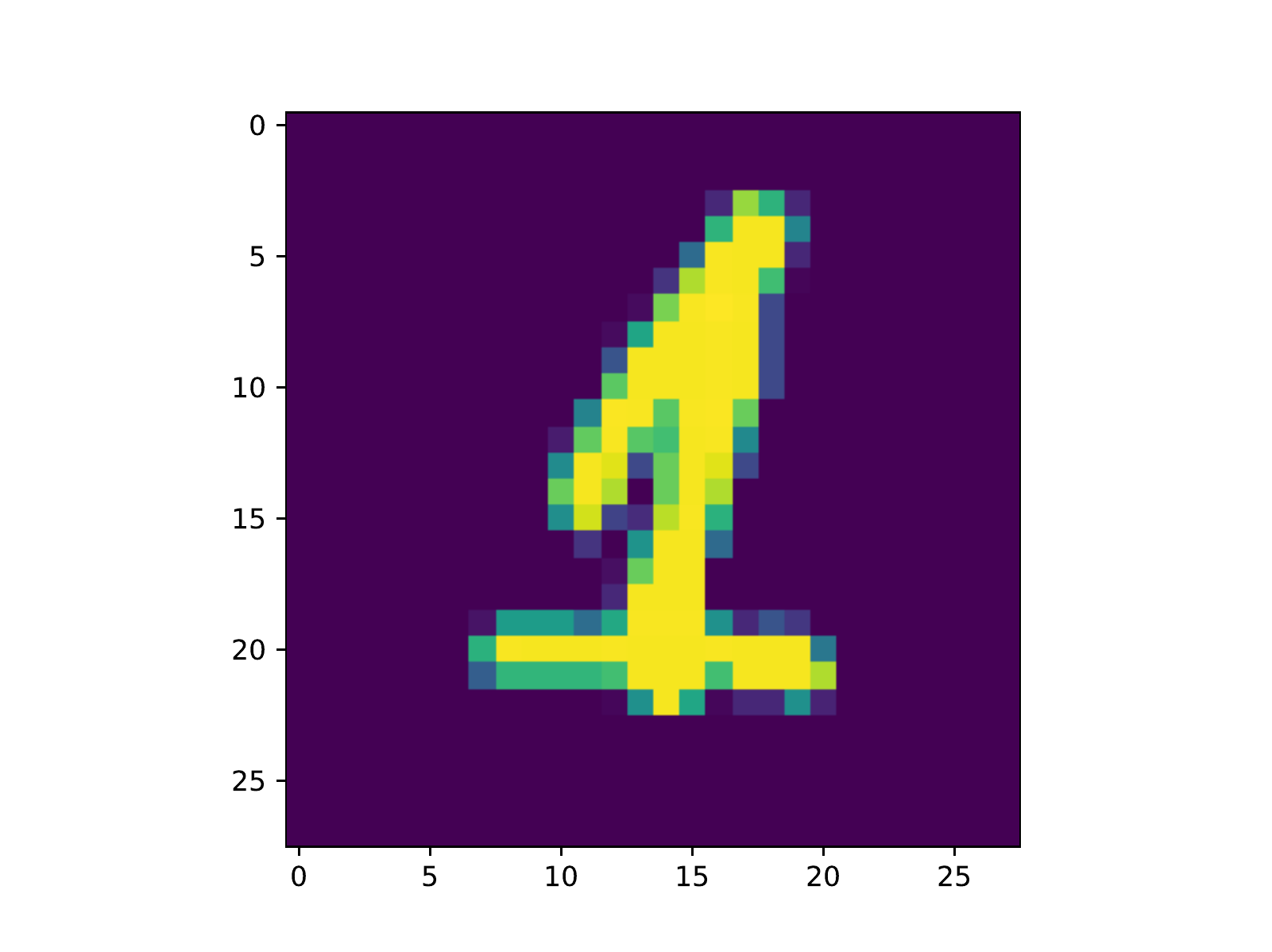}

C \includegraphics[scale=.08,trim=2cm 2cm 2cm 2cm, clip=true]{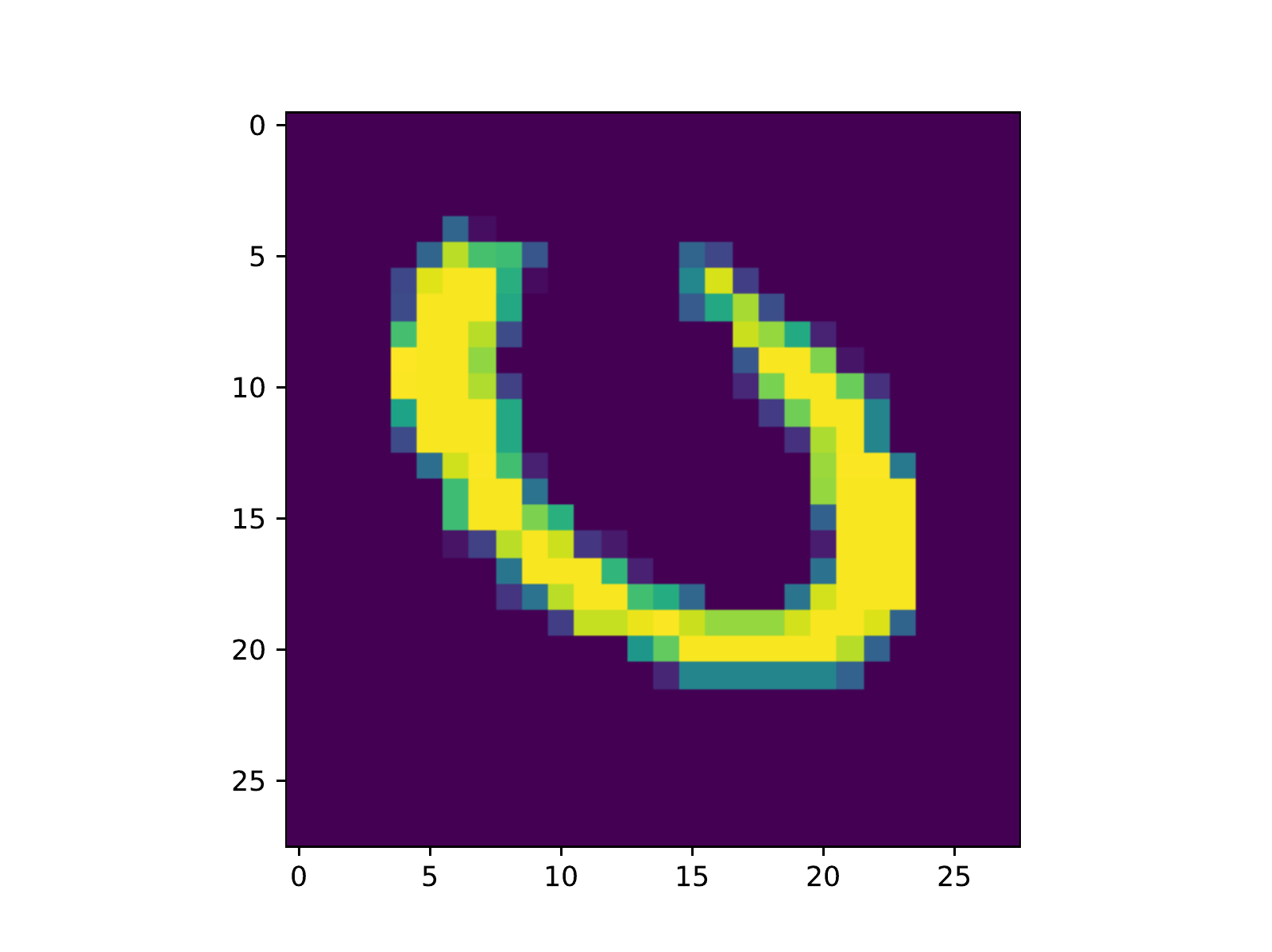}
\includegraphics[scale=.08,trim=2cm 2cm 2cm 2cm, clip=true]{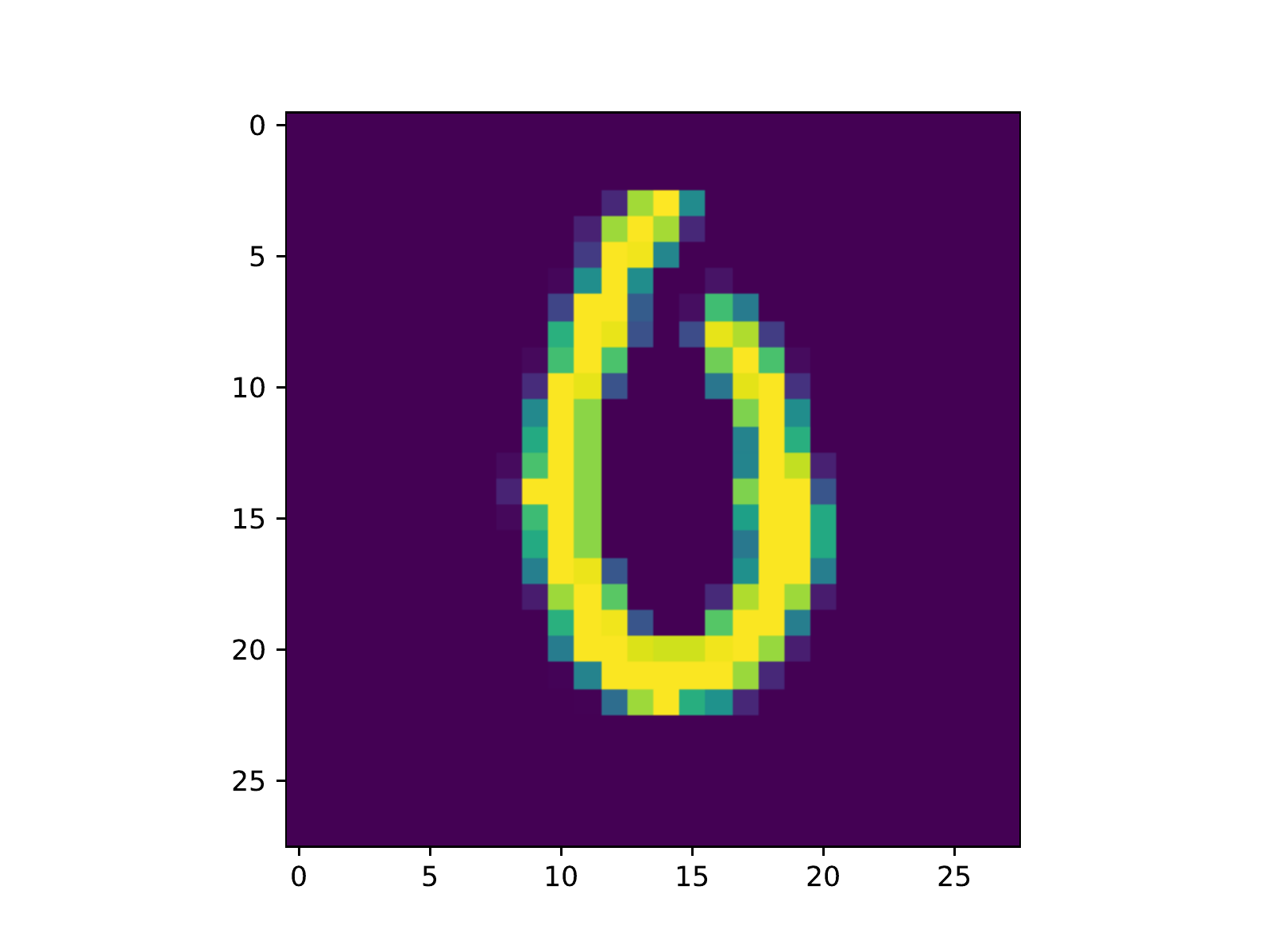}
\includegraphics[scale=.08,trim=2cm 2cm 2cm 2cm, clip=true]{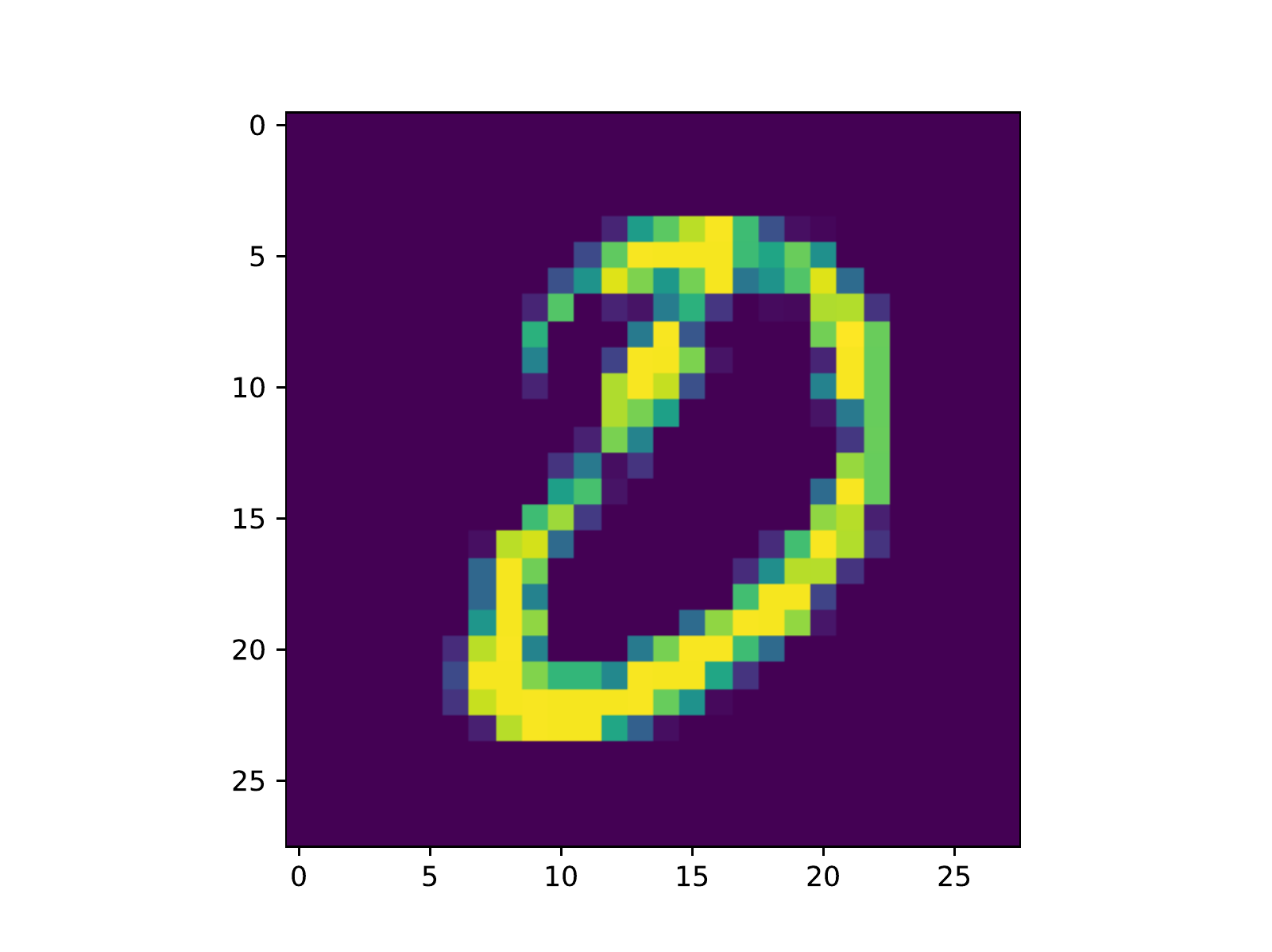}
\includegraphics[scale=.08,trim=2cm 2cm 2cm 2cm, clip=true]{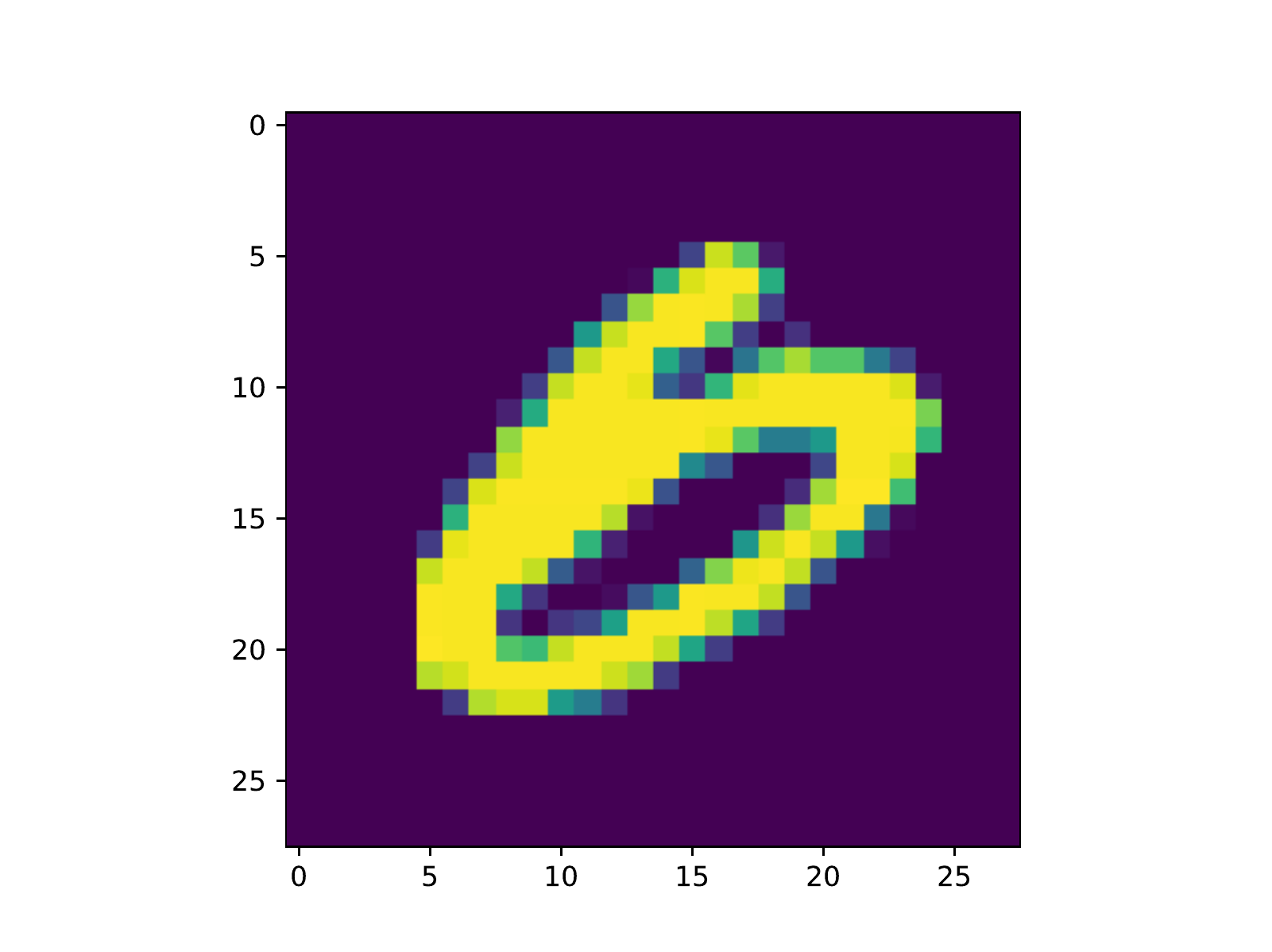}
 \includegraphics[scale=.08,trim=2cm 2cm 2cm 2cm, clip=true]{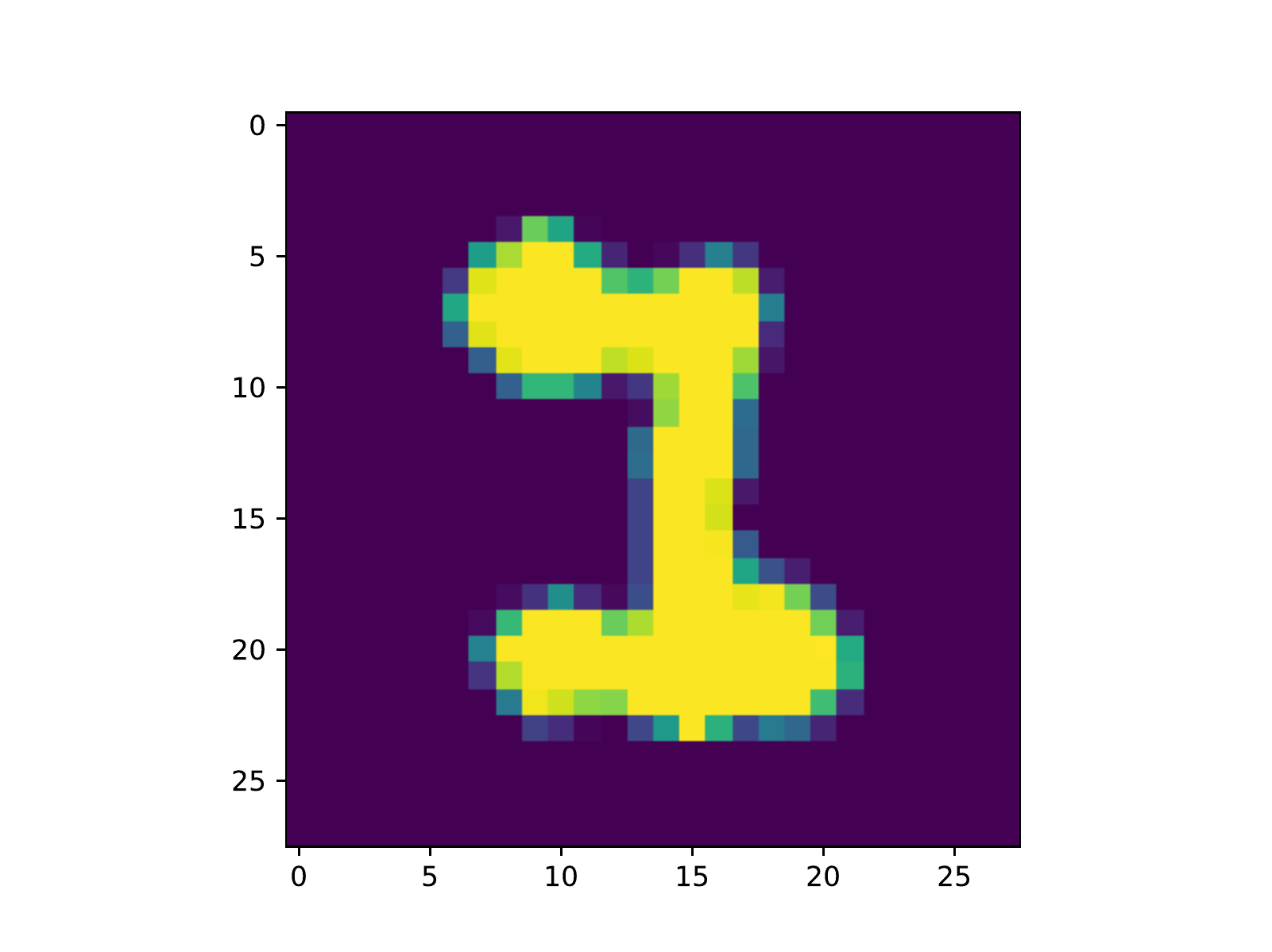}
\includegraphics[scale=.08,trim=2cm 2cm 2cm 2cm, clip=true]{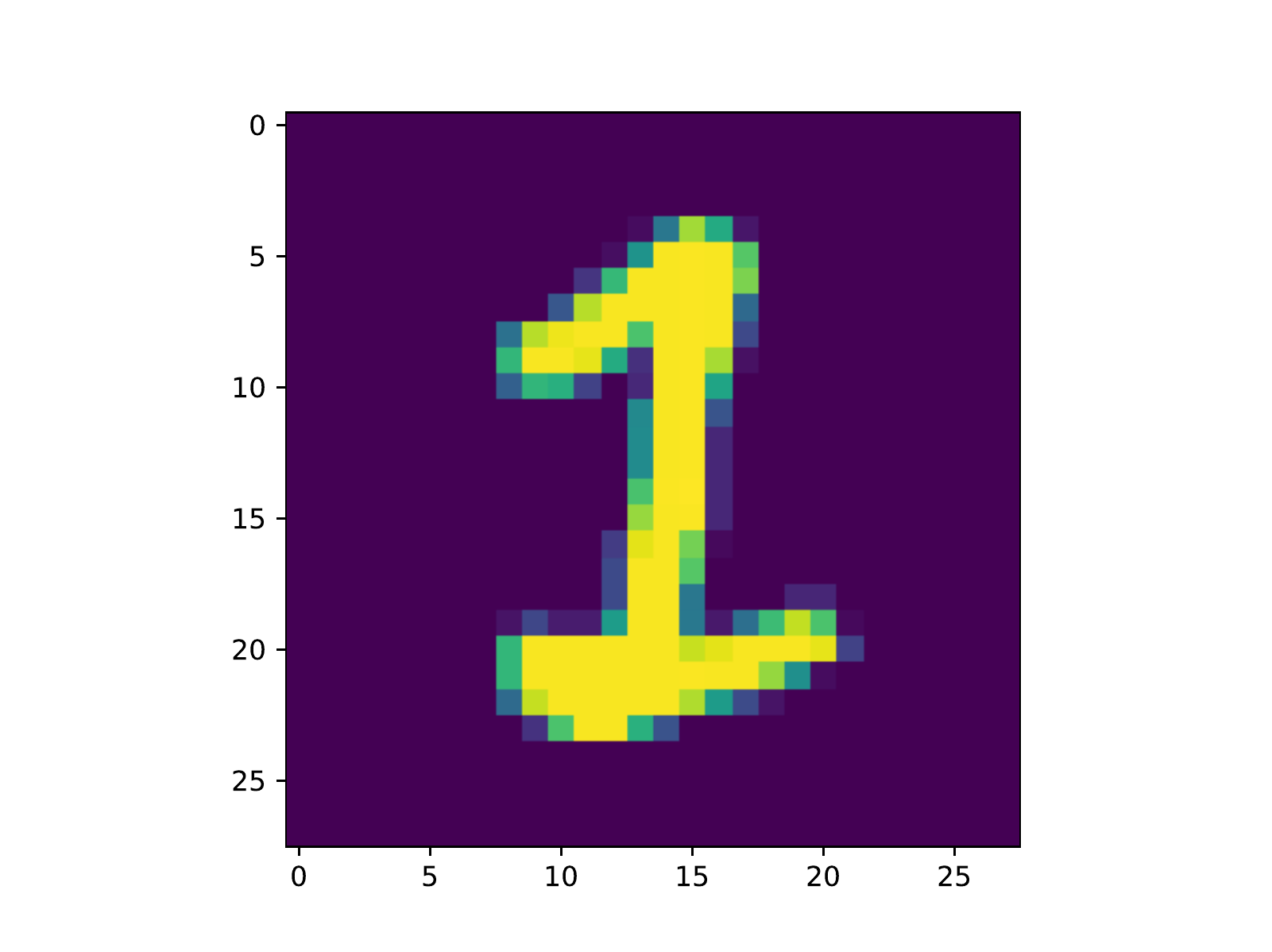}
\includegraphics[scale=.08,trim=2cm 2cm 2cm 2cm, clip=true]{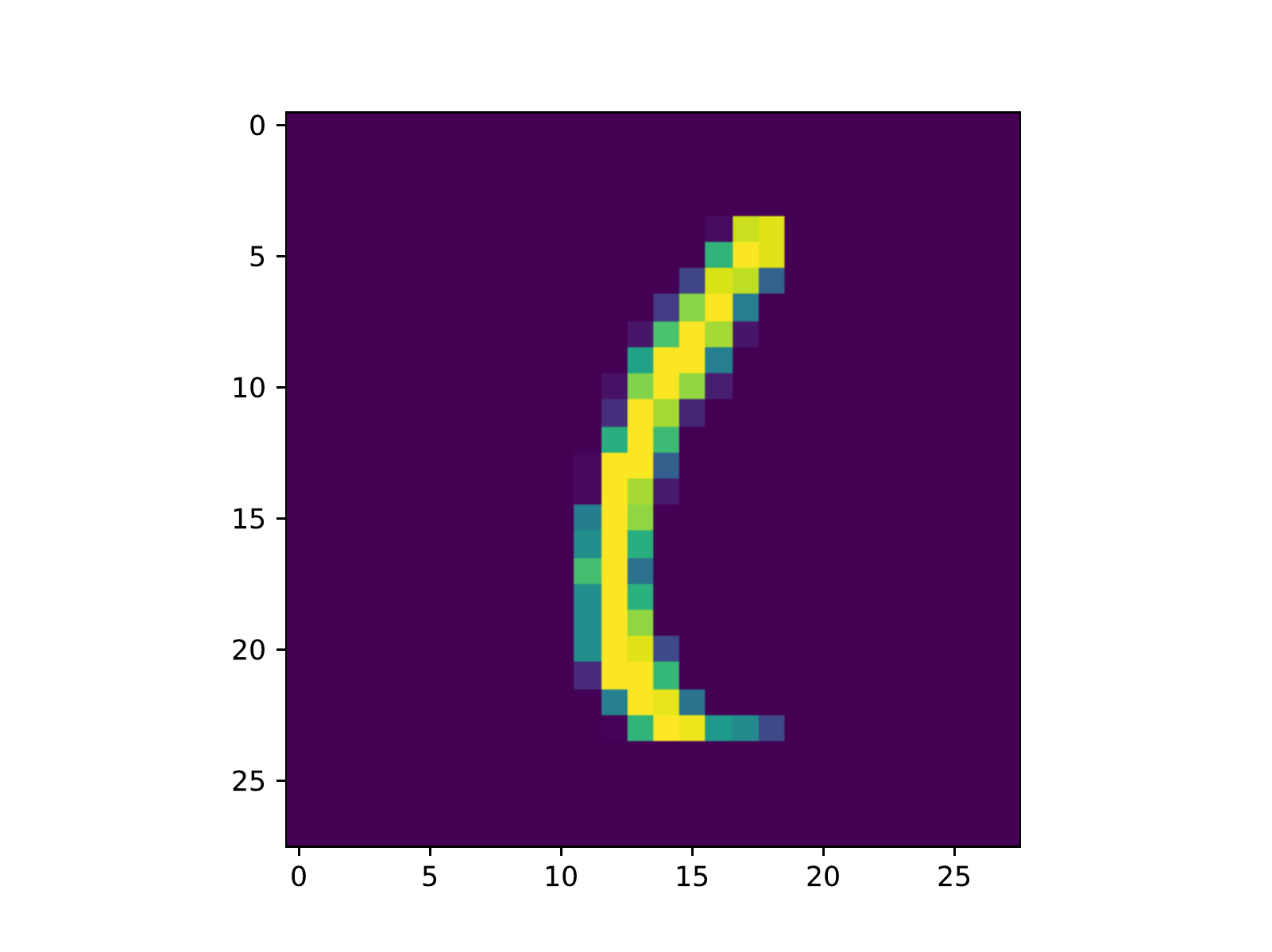}

D \includegraphics[scale=.08,trim=2cm 2cm 2cm 2cm, clip=true]{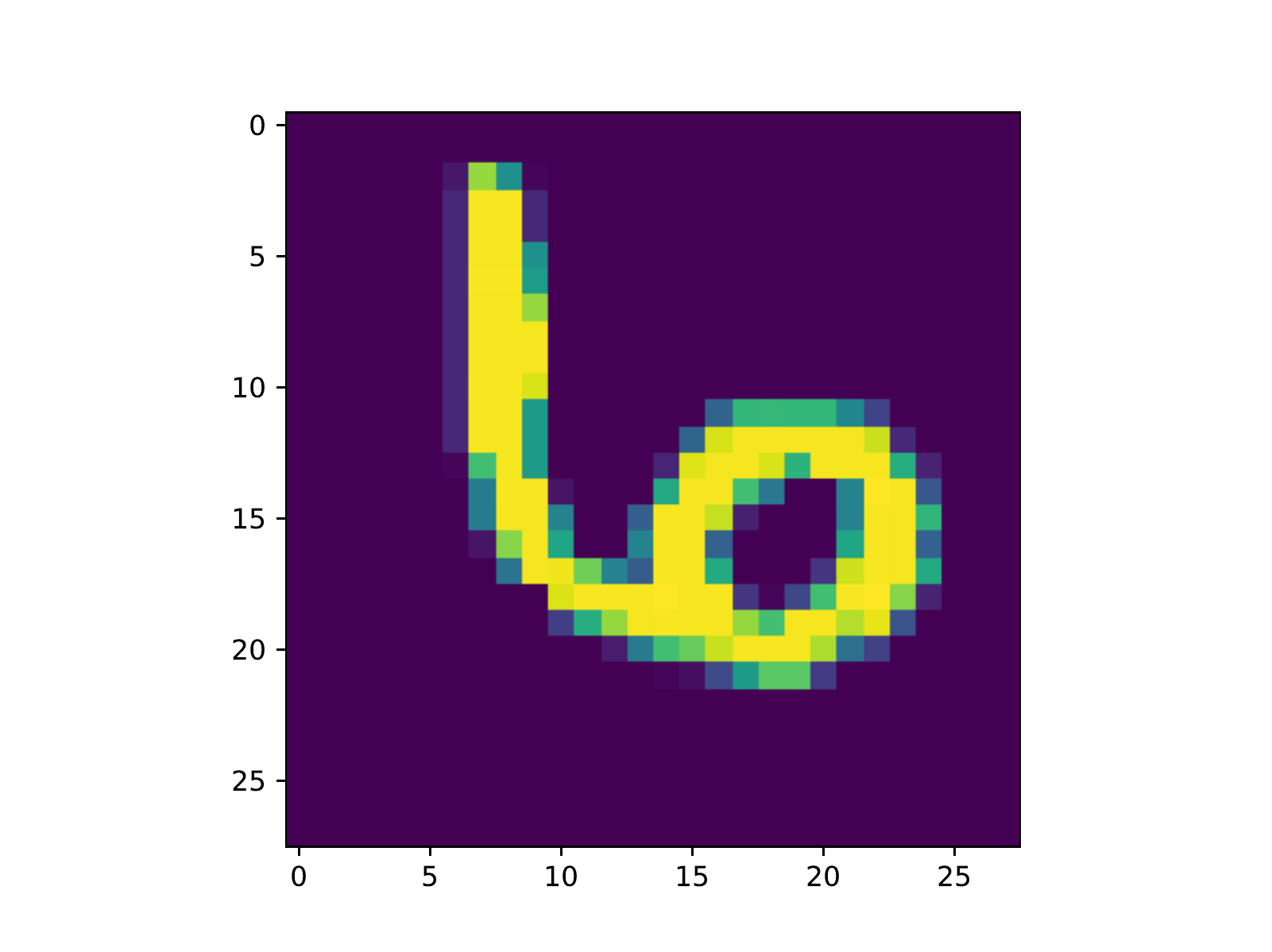}
\includegraphics[scale=.08,trim=2cm 2cm 2cm 2cm, clip=true]{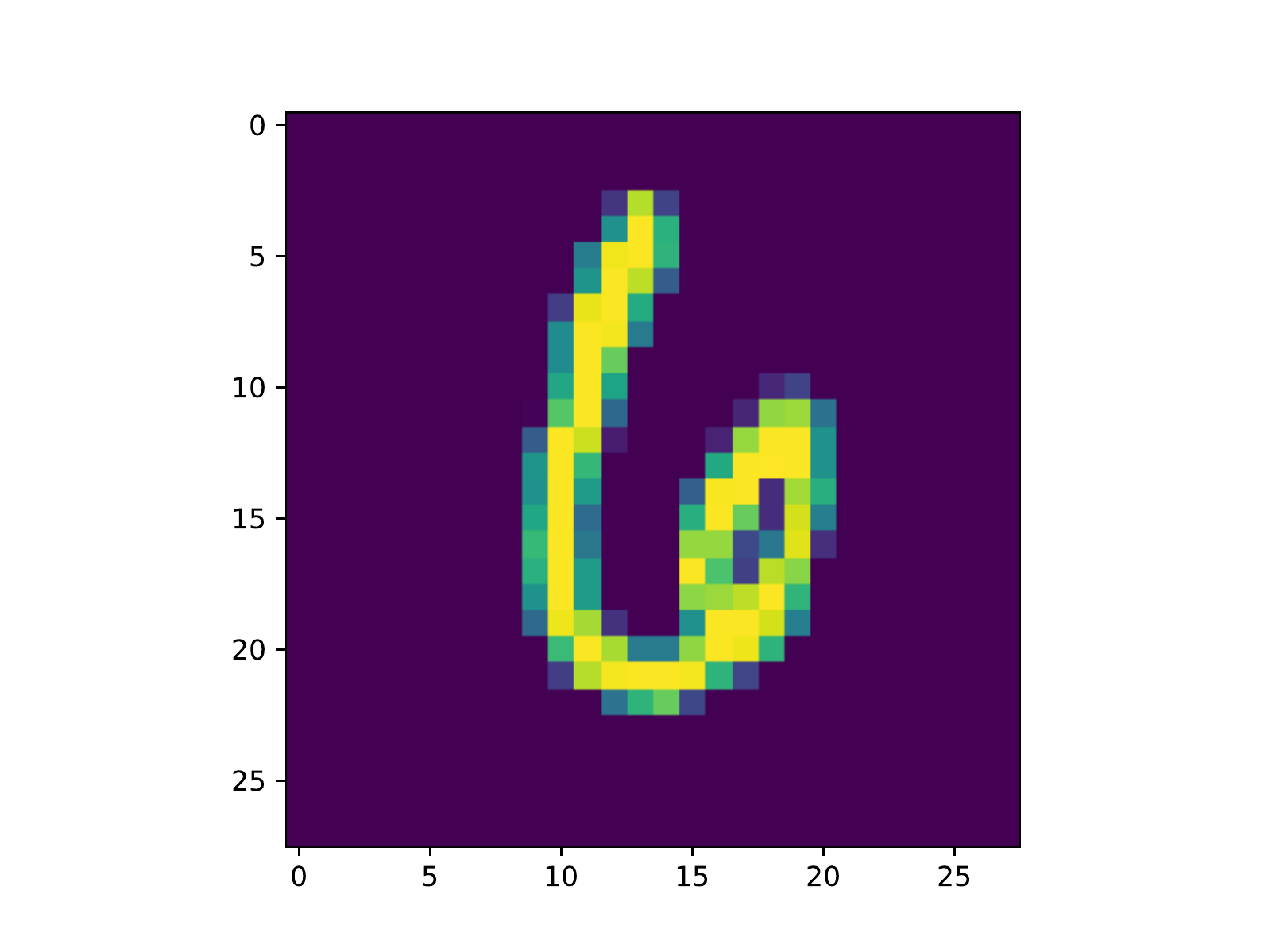}
\includegraphics[scale=.08,trim=2cm 2cm 2cm 2cm, clip=true]{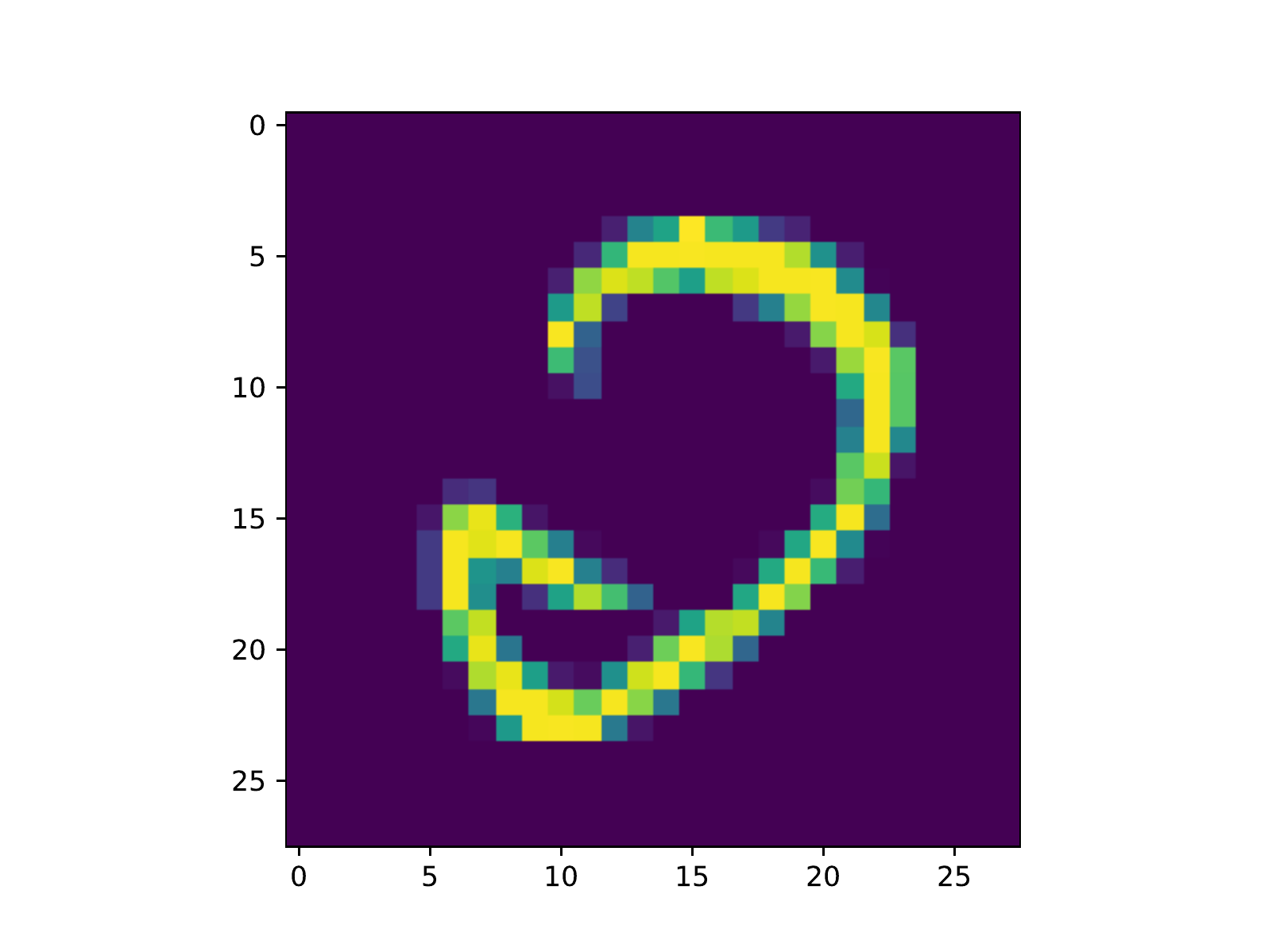}
\includegraphics[scale=.08,trim=2cm 2cm 2cm 2cm, clip=true]{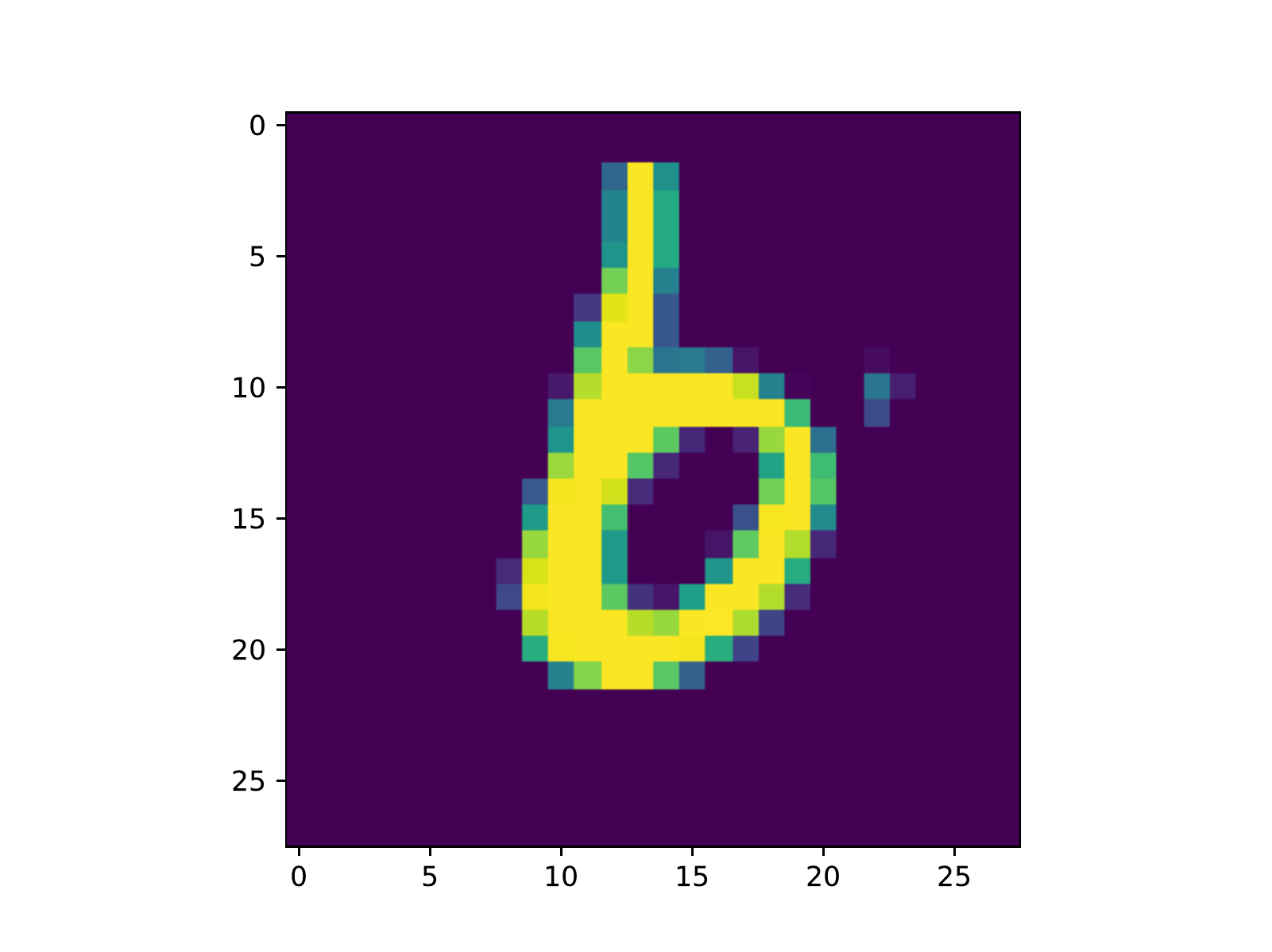}
\includegraphics[scale=.08,trim=2cm 2cm 2cm 2cm, clip=true]{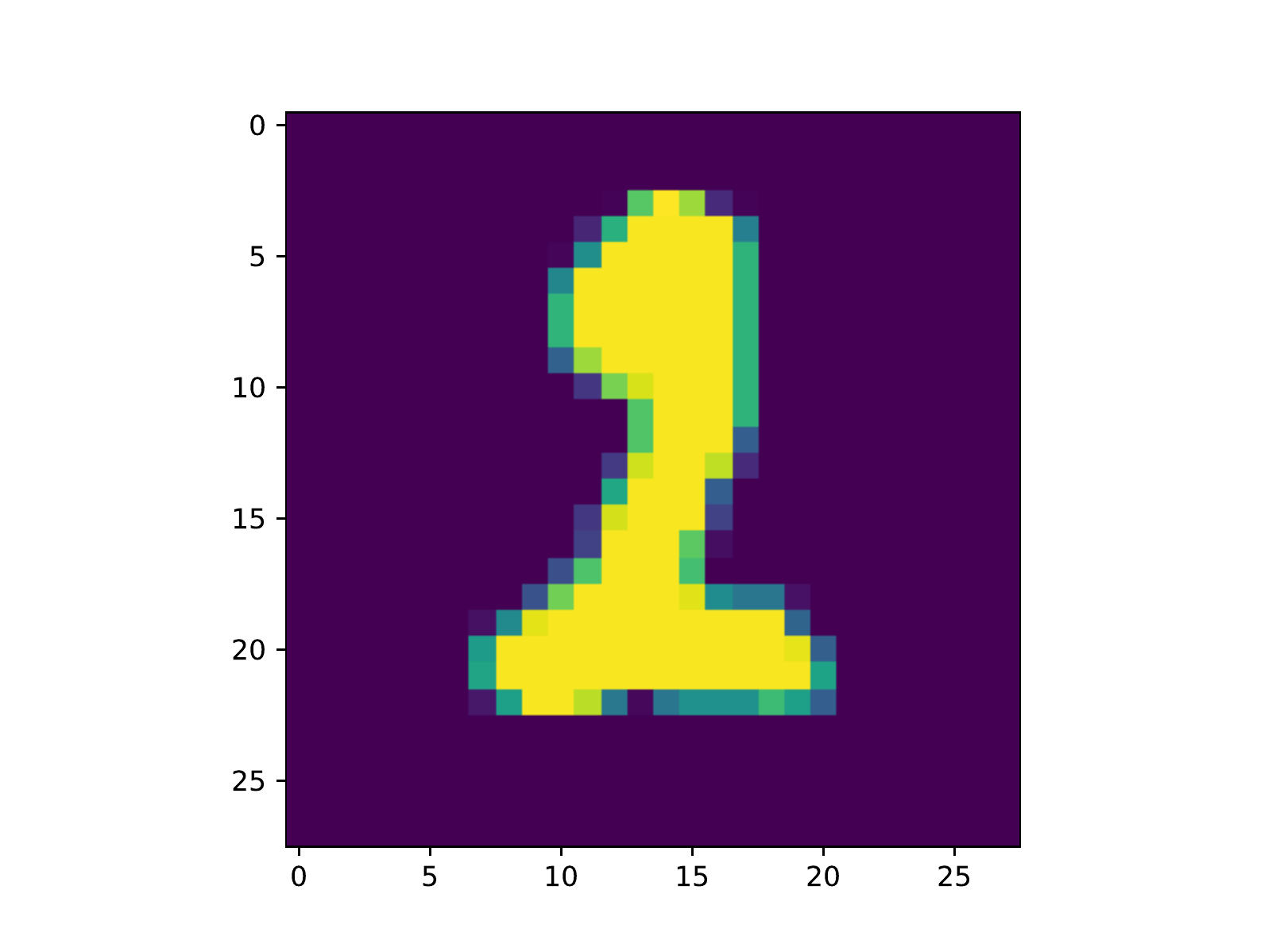}
\includegraphics[scale=.08,trim=2cm 2cm 2cm 2cm, clip=true]{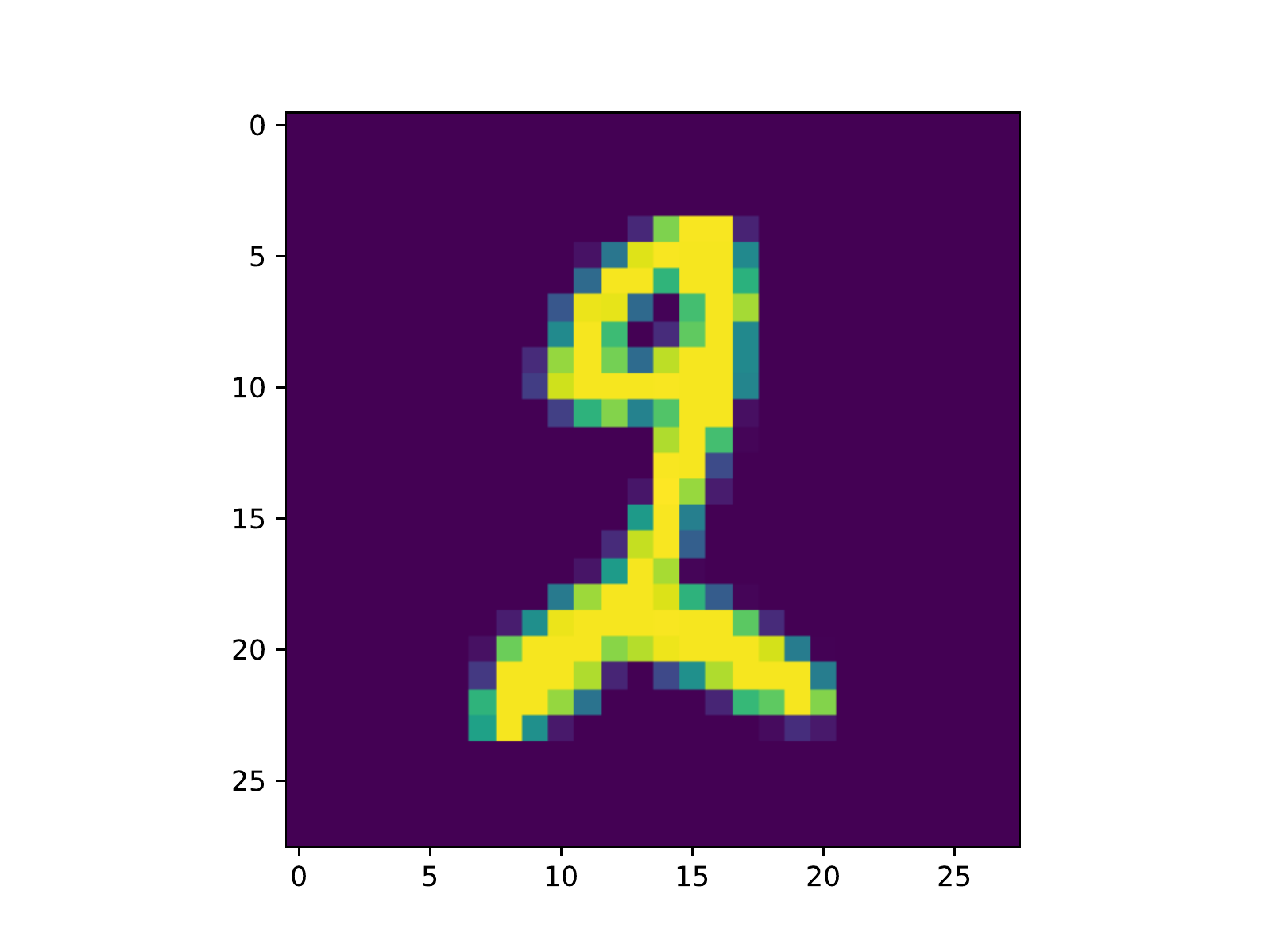}
\includegraphics[scale=.08,trim=2cm 2cm 2cm 2cm, clip=true]{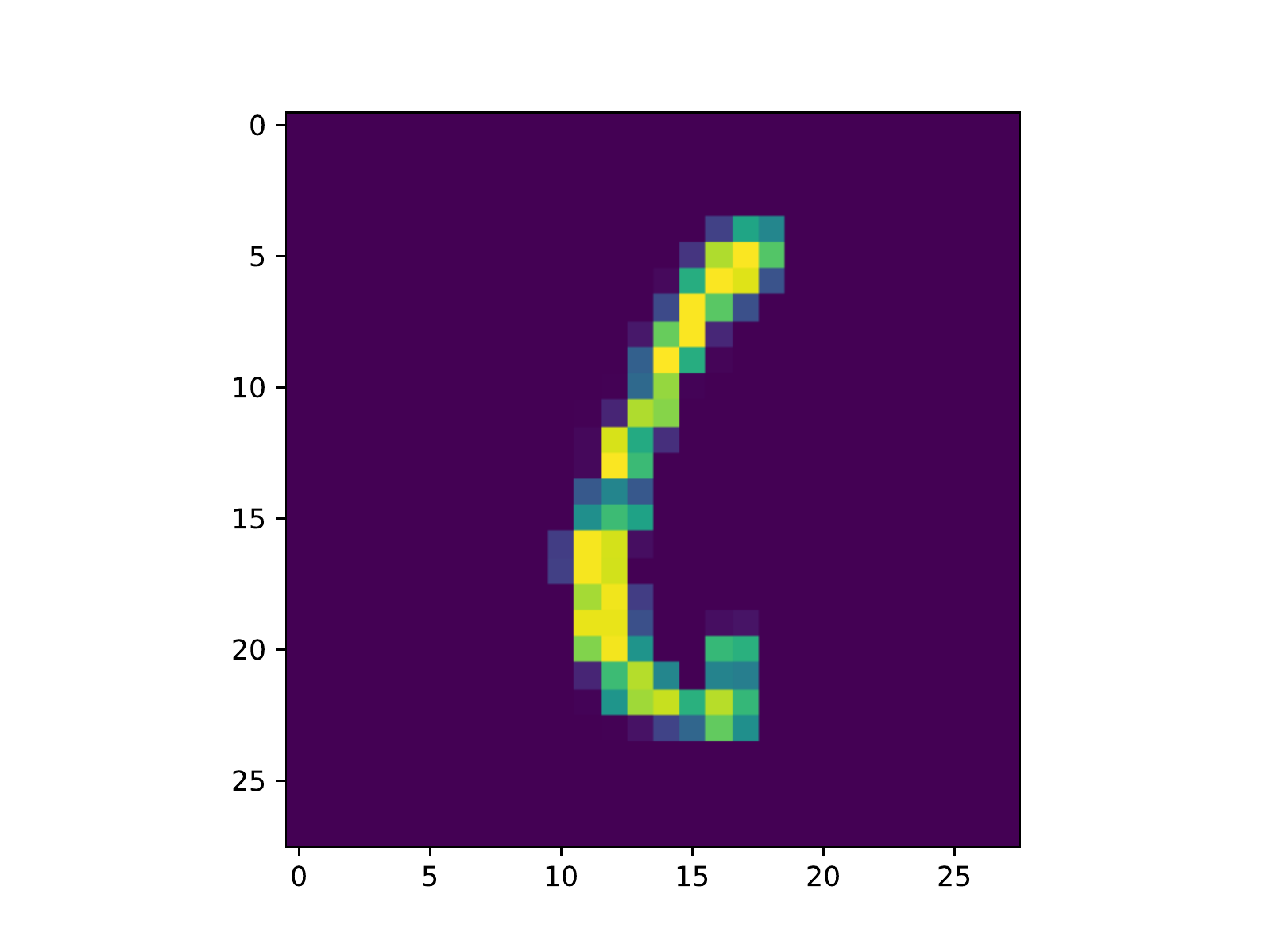}

\caption{
Row A shows interior points of the classes 
(class 0 left and class 1 right);
Row B shows typical boundary points of the classes, while row C
 shows outliers and row D shows the closest point to the outlier. Note that the results are generally consistent with human judgement and provide guidelines for training data enhancement based on the outliers.
}
\label{fig:Ri:MNIST}
\end{figure}
{\bf 4.2. Real Datasets.} 
In this section, we present results on real datasets drawn from 
handwritten number recognition 
MNIST \cite{lecun1998gradient} 
and medical/biological datasets like diabetes and iris data.
\\
{\bf Experiments on MNIST Dataset.}
The state-of-the-art classification method 
for MNIST data uses convolutional neural networks  which extracts useful features.
In this experiment we trained convolution auto-encoder (with ReLU activation)
and represented each image
by the compressed feature representation resulting in 16 features per image.
The results of our classification 
-- performed on this reduced 16D space --
are shown in Figure \ref{fig:Ri:MNIST}. These results demonstrate how our algorithm can be used to classify the training data into interpretable interiors, boundaries and outliers, with the latter two being visibly similar to other neighboring classes leading to possible mis-classification.
\\
{\bf Experiments on Medical Datasets.}
In this experiment we use
\emph{cod-rna} and \emph{diabetes} 
datasets\footnote{All datasets used in this Section are available at 
\url{https://www.csie.ntu.edu.tw/~cjlin/libsvmtools/datasets/}.}
each having 8 features.
We trained the optimal metric on 300 random points and tested on larger sets of points (as indicated in table). Note that our algorithm (labeled MeLL0), provides improvement in performance over Euclidean and various LMNN alternatives without any significant tuning of parameters on our part.
\\
{\bf Experiments on Biological Dataset.}
In Figure \ref{fig:IRIS} 
we show results on Iris Flower Dataset \cite{fisher1936use} which contains 
3 classes of 
  150 training samples,  each represented by 4 features.Note that our method separates the classes reasonably well, while also depicting relatively internally uniform characteristics for each class in terms of the Mahalanobis vectors, as seen by the vertical streaks in the images on the top row in Figure \ref{fig:IRIS}.
In all cases demonstrated above we observed 
  that our method produces competitive and interpretable classification results
  as well as outlier identification. This combination is useful for 
 choosing   comprehensive and economical training datasets.
 \section{Conclusion}
 We have presented novel methods for 
 classification based on optimized metric learning. Our methods are designed to overcome
 limitations of state-of-the-art alternatives 
 such as SVM, DNN by being more interpretable, robust and more general.
 We have shown promising results on synthetic and real datasets.

\begin{table} 
\small 
\centering 
\begin{tabular}{l|r|r|r}
approach & \emph{cod-rna}   & \emph{cod-rna}    & \emph{diabetes} 
\\
\# of test & 4,000 & 8,000 & 769
\\
\hline \hline 
Euclidean distance & 14.47\% & 13.43\% & 29.42\%
\\
MeLL0 & {\bf 6.27\%} & {\bf 6.17\%} &{\bf 23.30\%}
\\
LMNN $K=1$ & 8.82\% & 9.18\% & 29.81\%
\\
LMNN $K=3$ & 8.29\% &8.61\% & 30.59\%
\\
LMNN $K=5$ & 8.19\% & 8.47\%&29.94\%
\\ 
LMNN $K=7$ & 8.32\%& 8.78\%&  30.07\%
\\
LMNN $K=11$ &  8.64\%& 8.88\%& 30.46\%

\end{tabular}
\caption{Comparison of classification error for various distance metrics. Note that MeLL0, which uses outlier removal in both training and testing phases, provides some improvement over 
state-of-the-art alternatives. Further, the outliers, as before, suggest new desirable data  to improve performance.}
\centering 
\label{tbl:cod-rna_diabetes}

\end{table}

\section*{Acknowledgements} 

The
work of MT was partially supported by the U.S. National Science Foundation, under award number
NSF:CCF:1618717, NSF:CMMI:1663256 and NSF:CCF:1740796.

\bibliographystyle{icml2017}
\bibliography{literature}

\clearpage

\appendix
\onecolumn
\part{Appendix}
\section{Class boundaries}

\begin{figure}[h!] 
\centering
\includegraphics[scale=.4]{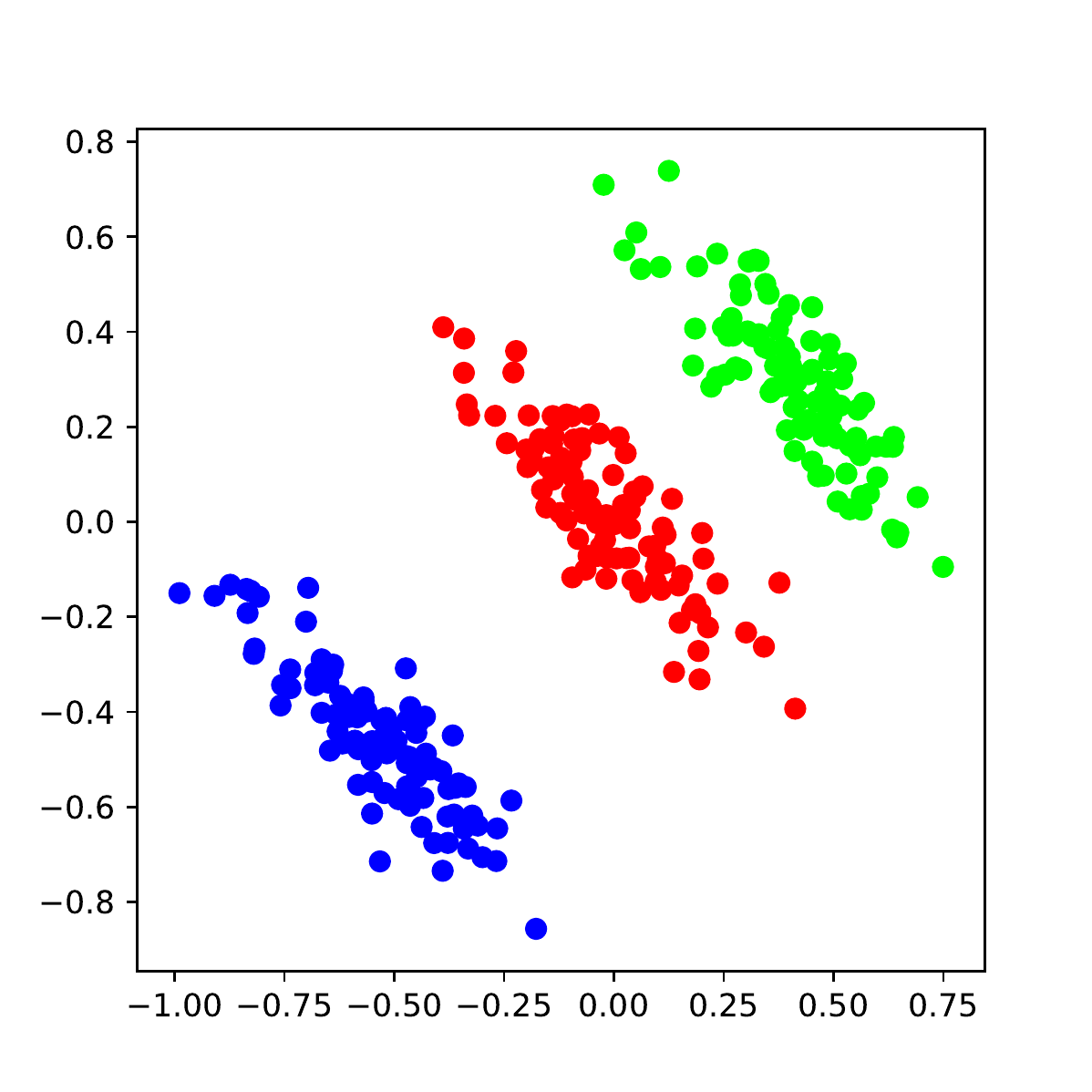}
\includegraphics[scale=.27]{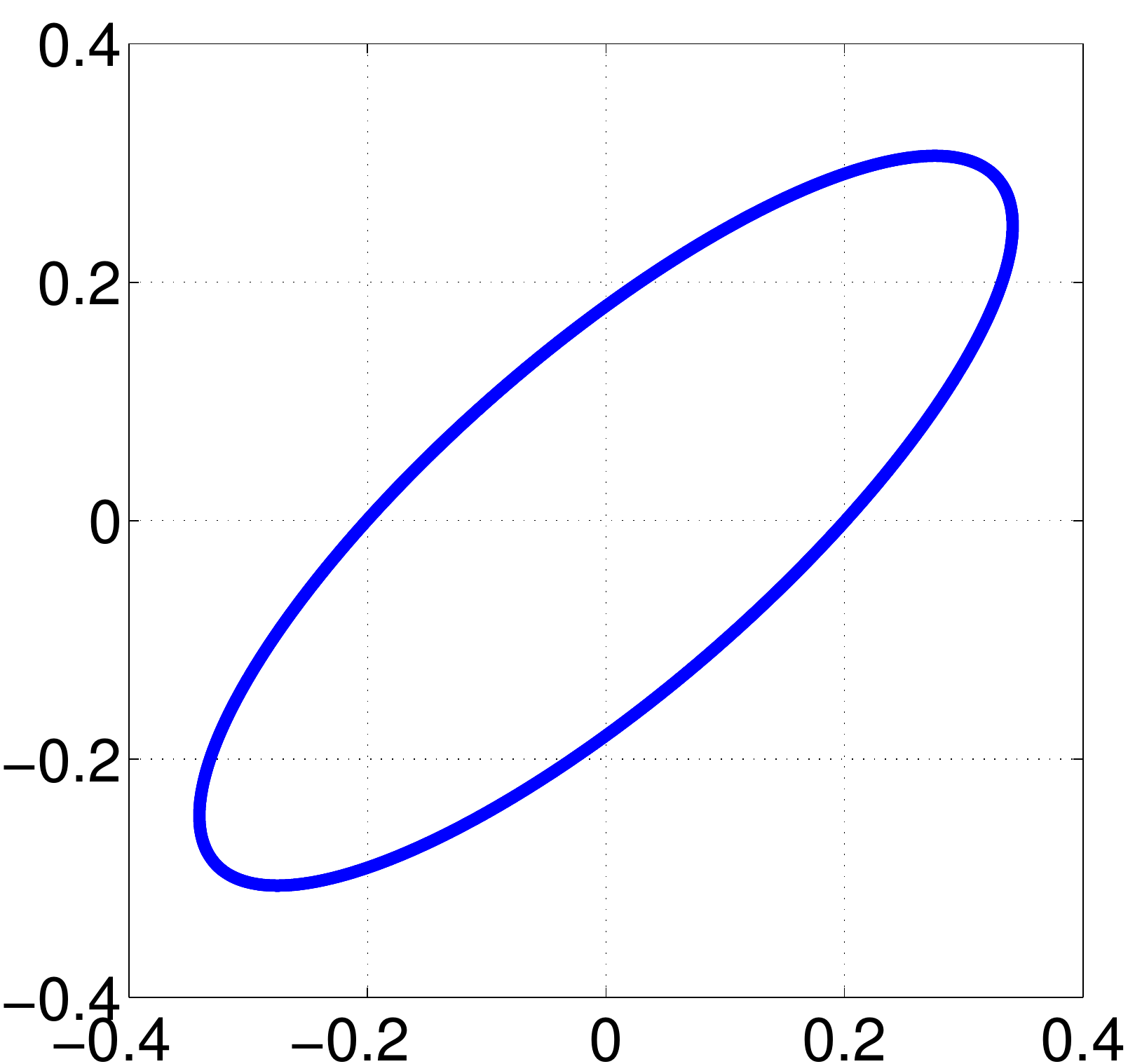}
\includegraphics[scale=.4]
{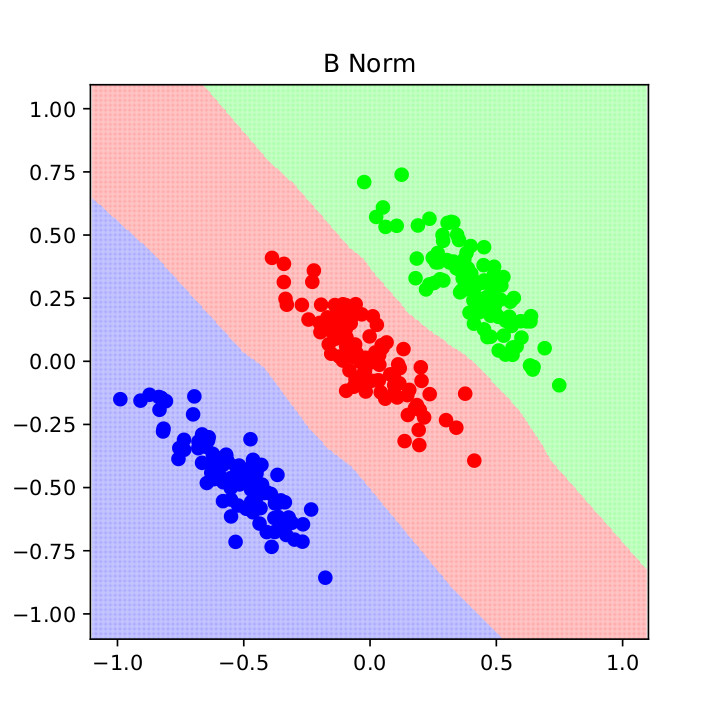}

\includegraphics[scale=.4]
{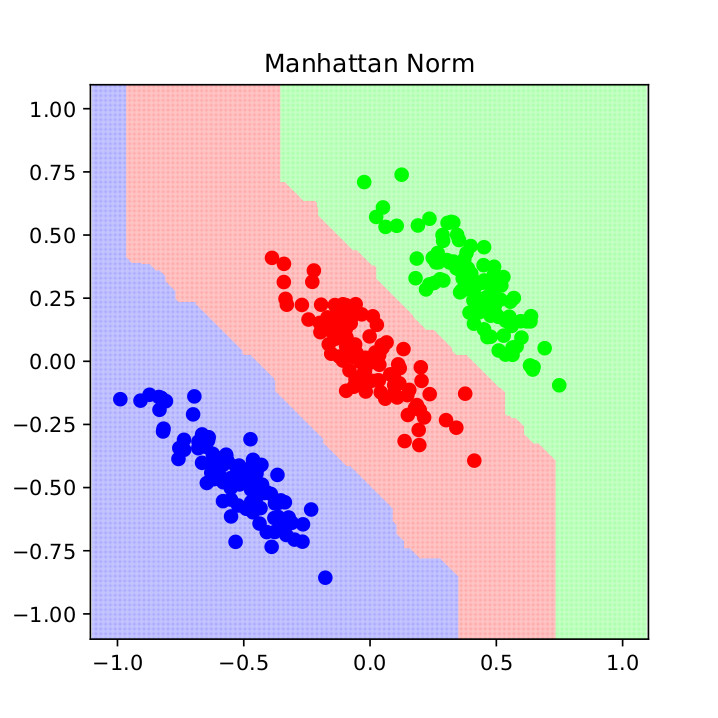}
\includegraphics[scale=.4]
{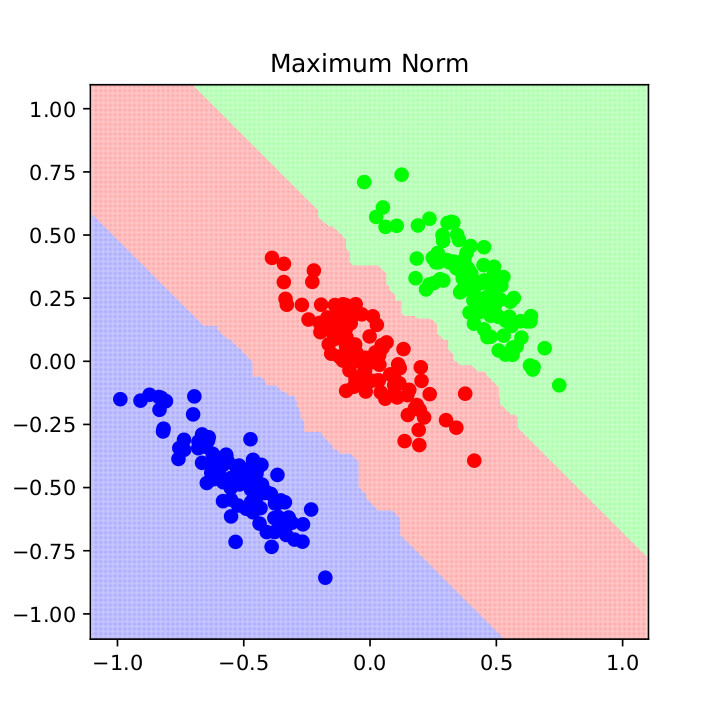}
\includegraphics[scale=.4]
{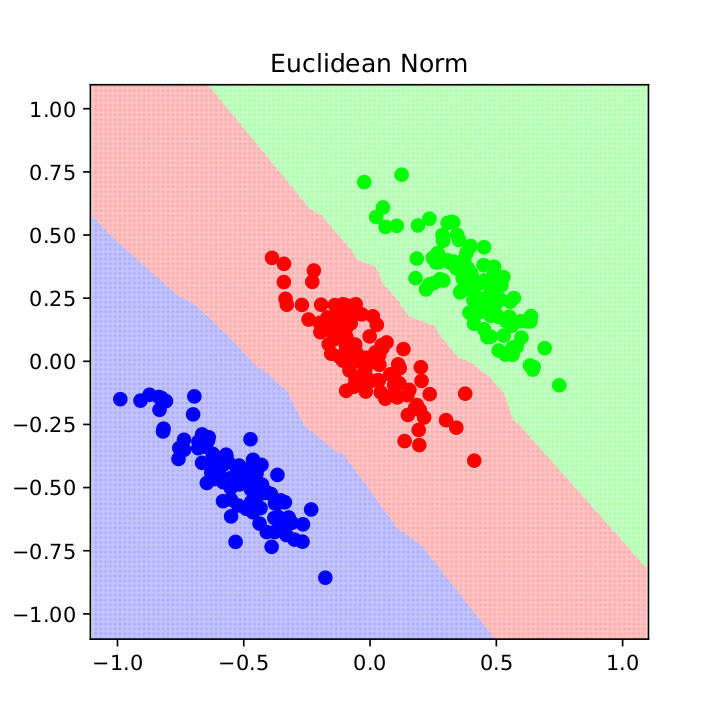}

\caption{
The boundaries learnt by our algorithm are smoother than alternatives. 
Top left is the original data; top right is the result produced by our algorithm; 
bottom 3 are the alternative approaches.
}
\label{fig:boundary}

\end{figure}

Figure~\ref{fig:boundary}
demonstrates the characteristics of 
class boundaries inferred from our optimal metric in comparison to commonly used metrics like Manhattan ($\ell_1$), Euclidean ($\ell_2$) and Maximum ($\ell_\infty$).
Smooth and robust boundaries are often practically important for robust online classification.
 Note that our boundaries are significantly smoother and 
less prone to overfitting, although the Euclidean metric performs similarly. However, in the presence of outliers
and noise, we will later show that the Euclidean metric performs much worse.

\begin{figure}
\centering

\subfigure[shows the original data consisting of sparse outliers, e.g. red point in the blue region.]{\includegraphics[scale=.3]
{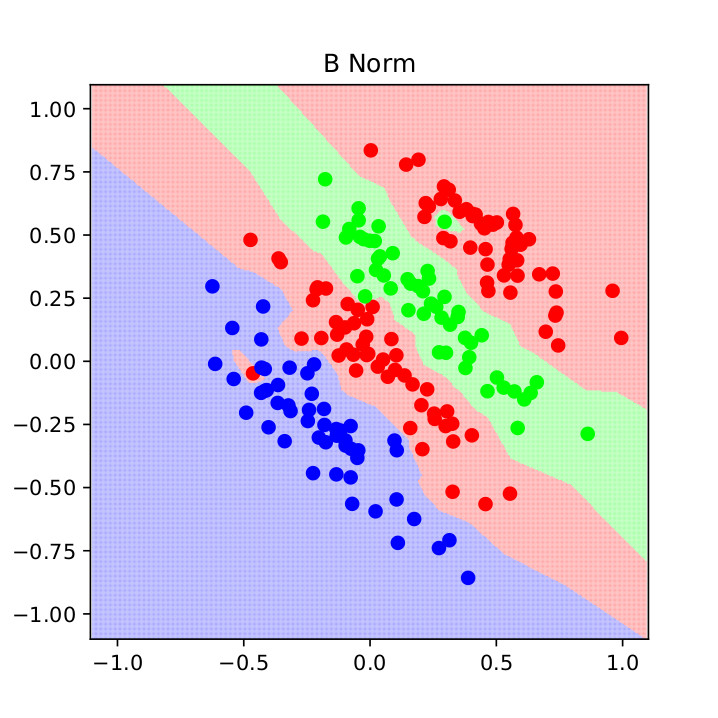}
}
\subfigure[identification of outliers shown as black dots]{ 
\includegraphics[scale=.3]
{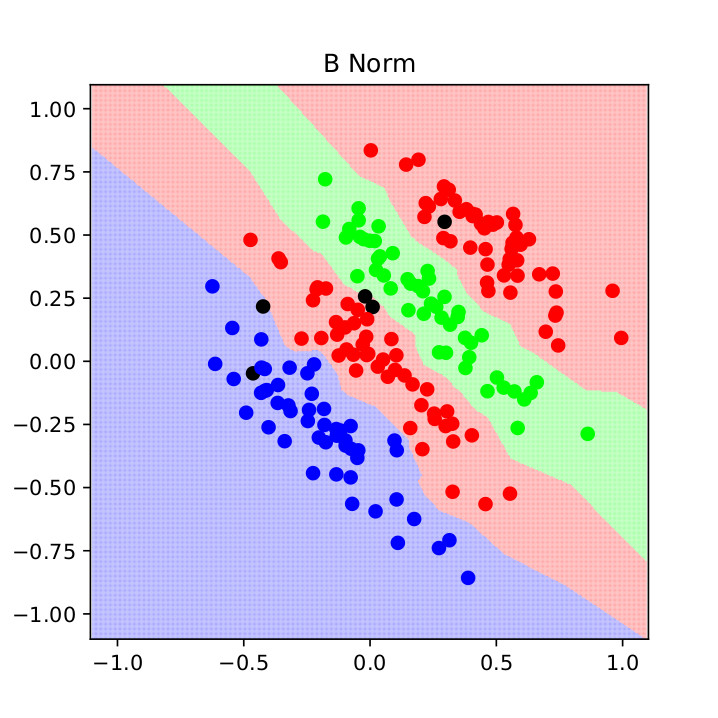}
} 
\subfigure[smooth class regions from refined metric]{ 
\includegraphics[scale=.3]
{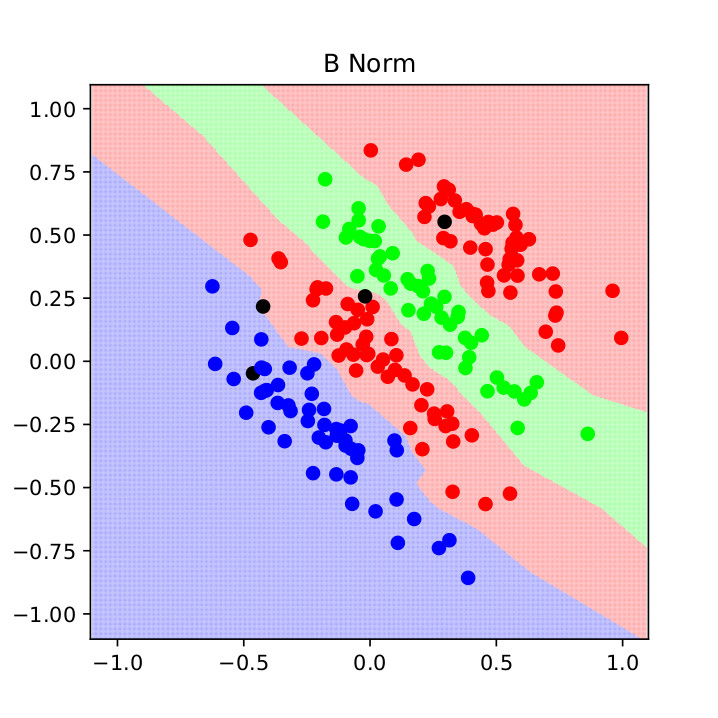}
}

\subfigure[is the optimal metric  for a full data]{ 
\includegraphics[scale=.2]{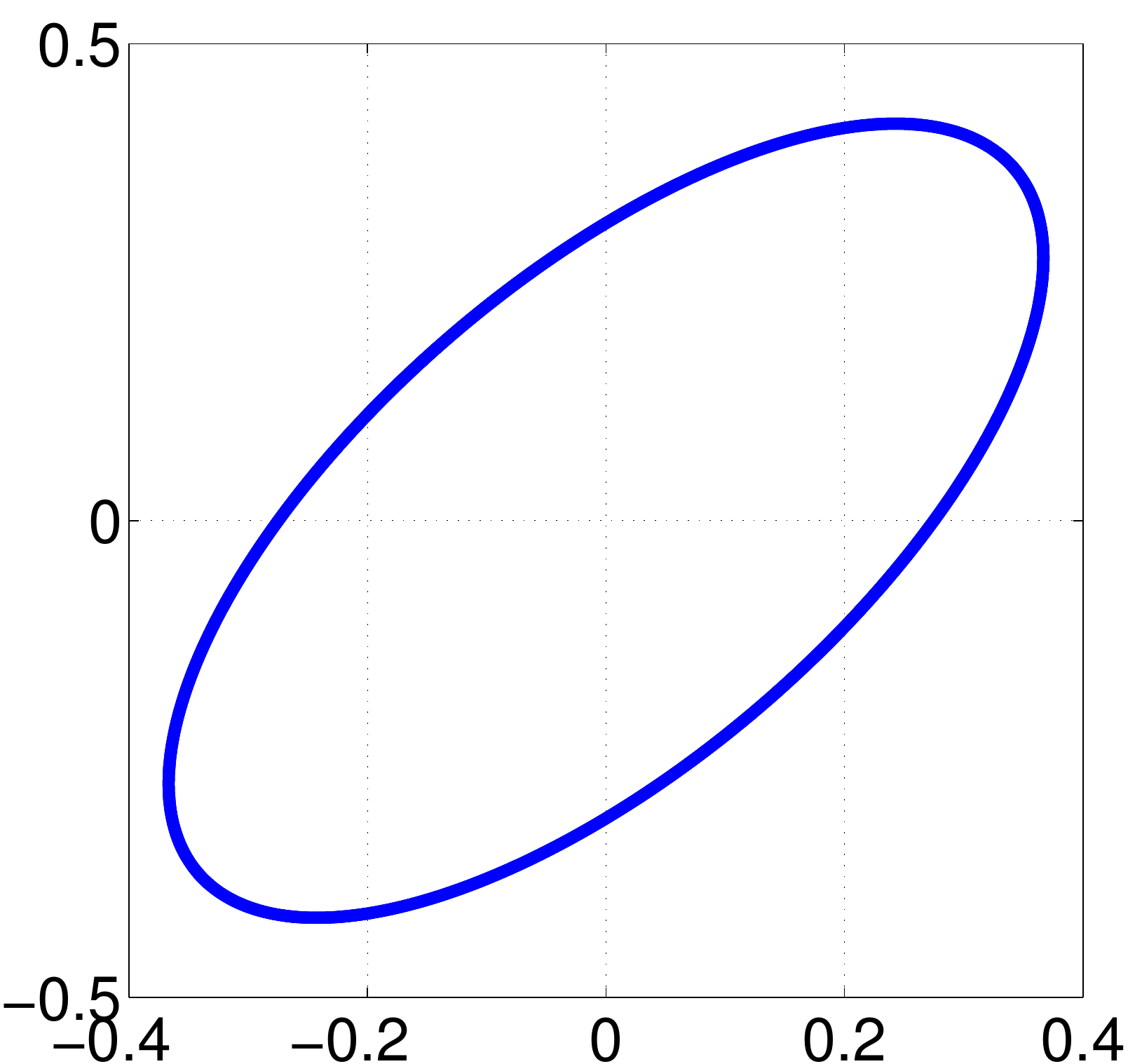}
} 
\subfigure[refined metric with outliers removed]{\includegraphics[scale=.2]{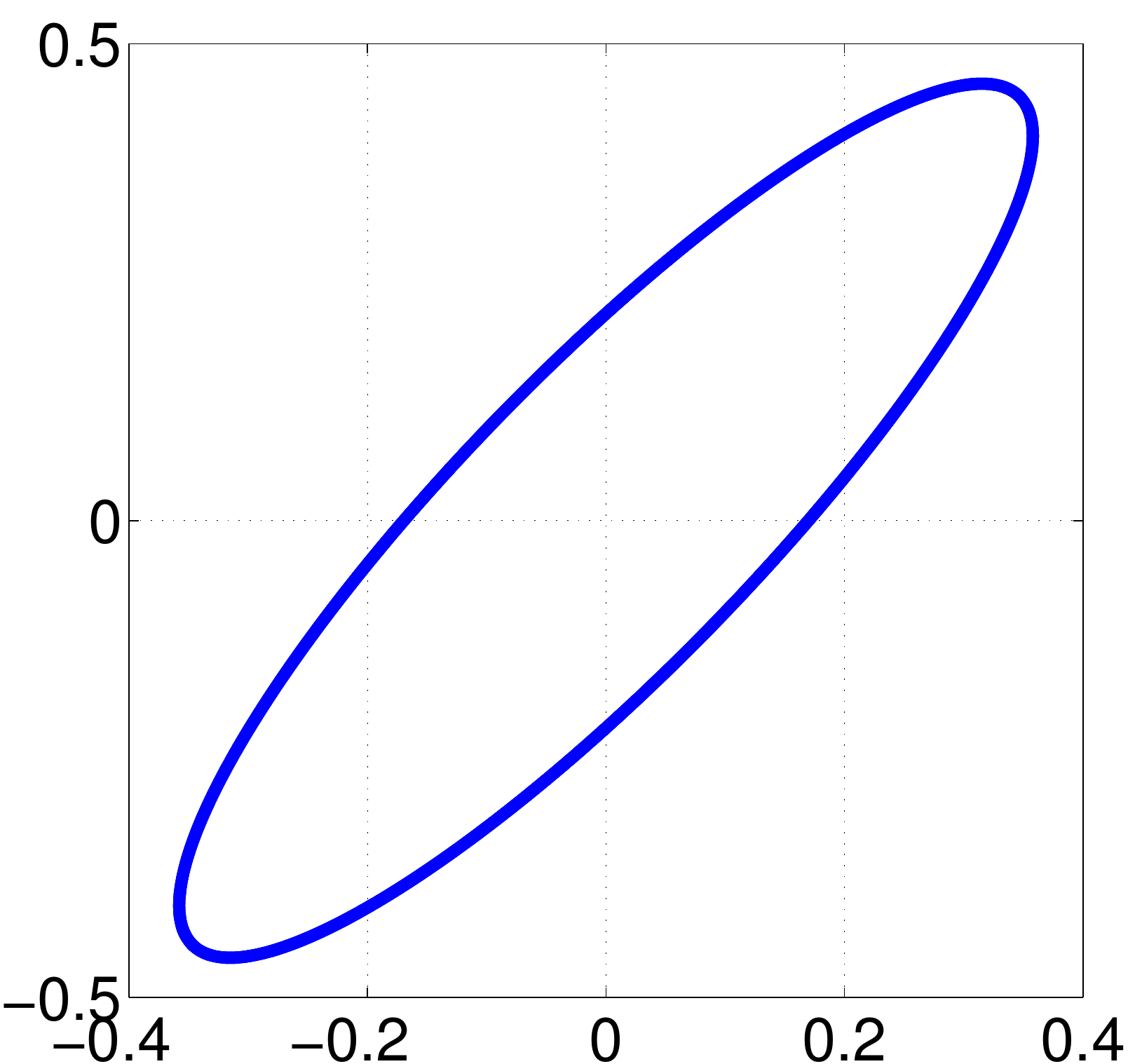}
} 

\caption{Outlier removal.}
\label{fig:outliers}

\end{figure}

\subsection{Handling Outliers}

Handling outliers is a very important task for robust classification.
In this section, we show how we can improve the distance metric by iteratively removing outliers using synthetic example.   
 Figure \ref{fig:outliers}
 demonstrates our outlier removal procedure,
 which consists of 3 major iterations. 
\begin{enumerate} 
\item determination of optimal metric for the whole dataset;
\item identification of outliers using the distance ratio;
\item reoptimization of the metric after removal of outliers.
\end{enumerate}
 Steps 1--3 can be optionally repeated iteratively  until satisfactory metric is obtained.\footnote{This procedure is a heuristic version of our thorough scheme which incorporates
trade-off curves. }

\begin{figure}
\centering 

\includegraphics[scale=.4]
{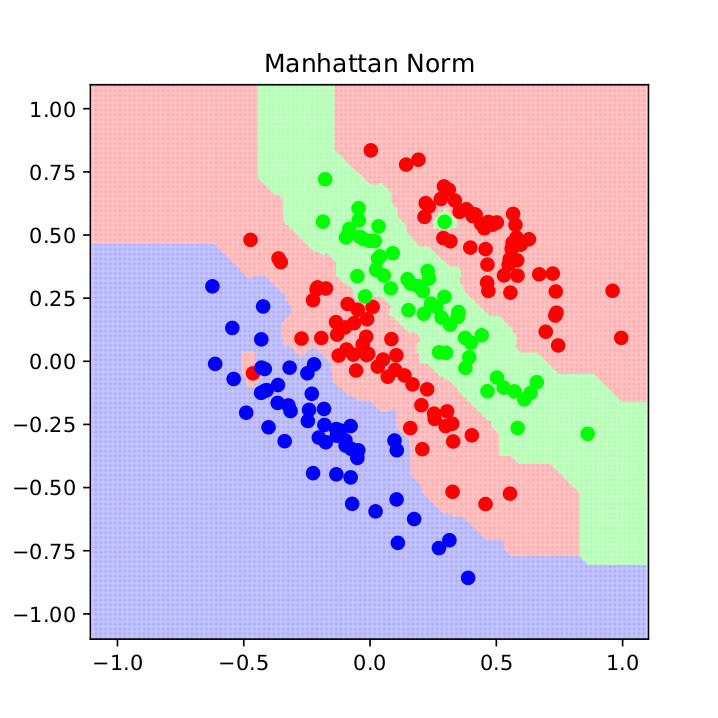}
\includegraphics[scale=.4]
{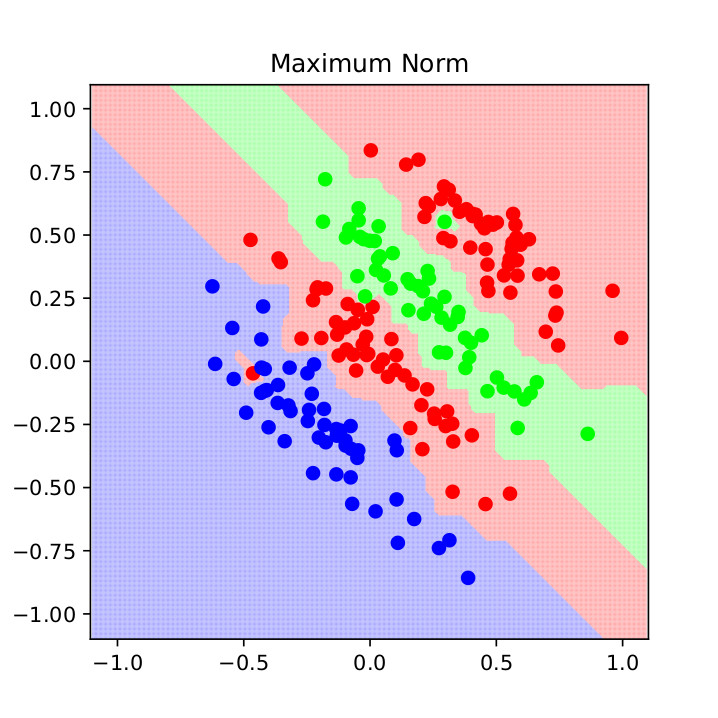}

\includegraphics[scale=.4]
{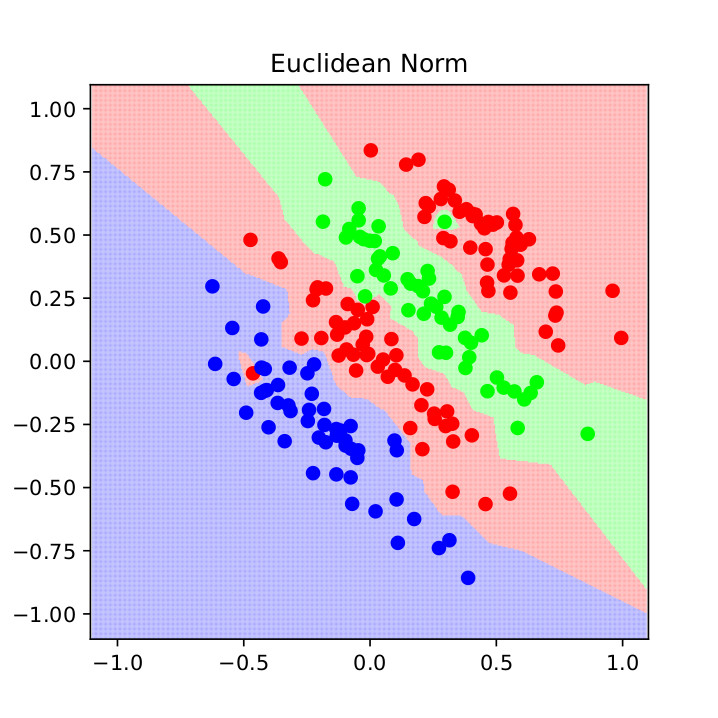}
\includegraphics[scale=.4]
{syntheticWithOutliers_txt_3-eps-converted-to_small.jpg}


\caption{Our result (bottom right) produces smoother class boundaries with outlier removal compared to alternatives.}

\label{fig:outliersAndOther}

\end{figure}

Finally we present a comparison of our optimal metric with outlier removal to the commonly used Euclidean, Manhattan and Maximum metric.
As mentioned earlier, Figure \ref{fig:outliersAndOther} 
shows that our optimal metric produces smoother class boundaries with outliers. In contrast, Euclidean distance gives significantly inferior results. 
 

 \section{Training DNN using only boundary points}

In this experiment, we explore how our approach can be used in training a deep neural network using far fewer points than typically required.  Specifically, after finding a near-optimal metric using our proposed strategy, we identify the ``boundary'' points and ``outliers'' of the data set and then feed these points to a DNN for training.

\begin{figure*}

\includegraphics[width=6cm]
{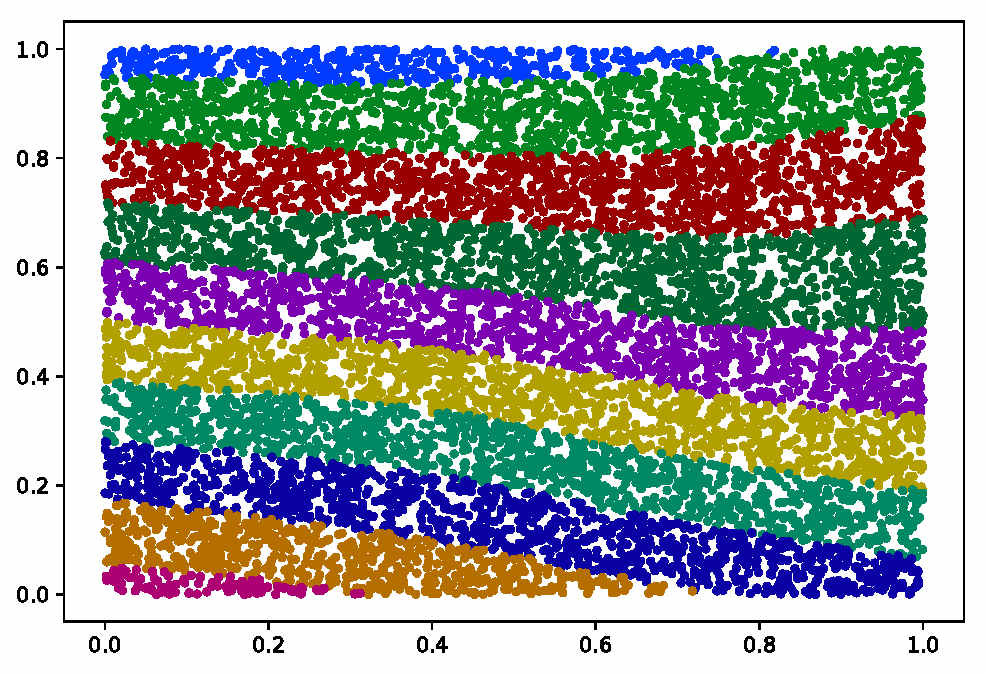}
\includegraphics[width=6cm]{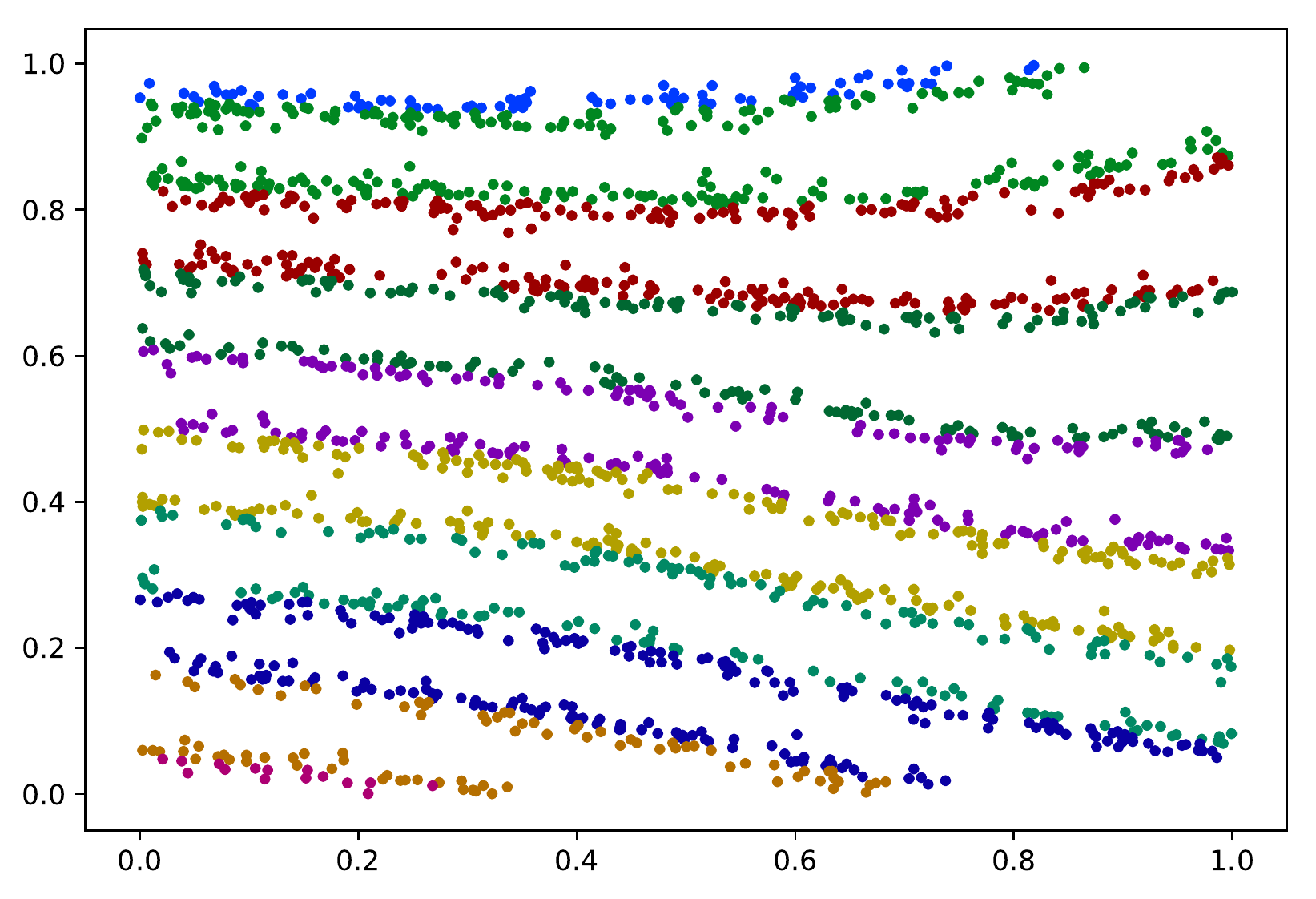}
\includegraphics[width=6cm]
{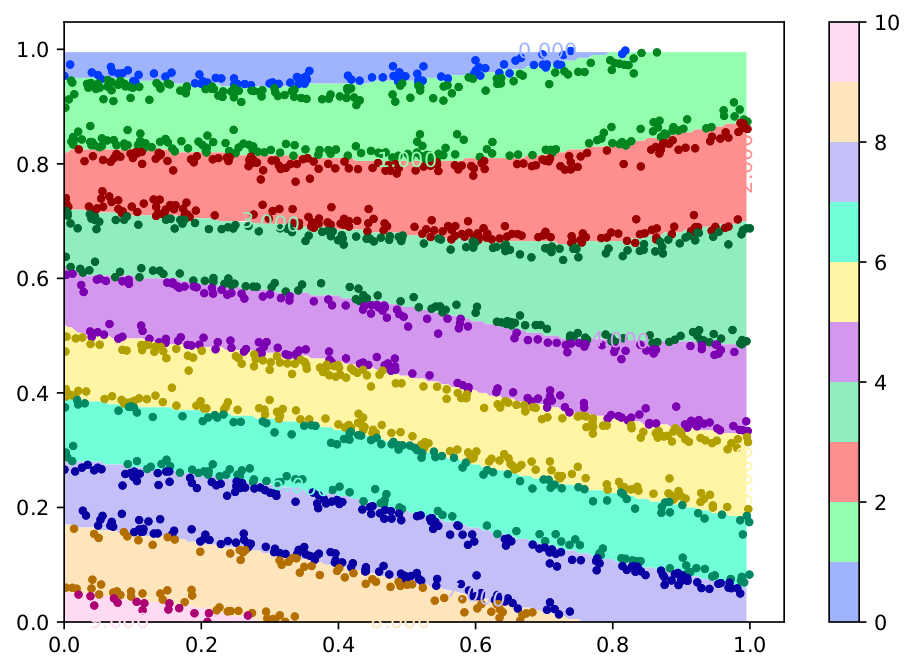}

\caption{Left figure shows a full, densely sampled 10-class, 10000-point dataset in 2D. Middle figure shows the boundary points computed using the $R$-ratio~(\ref{eq:R-ratio}). Right figure shows DNN classification of the entire domain using only the boundary points as training set.}
\label{fig:dnn:boundary_vs_interior}
\end{figure*}

In Figure~\ref{fig:dnn:boundary_vs_interior}
left, we show a simple datasets with 10 classes
containing 10000 points in 2D.
In the middle, we show the boundary points which were then used to train a fully-connected DNN with 4 hidden layers and ReLU activation function. We used soft-max cross-entropy loss function
and trained it using only boundary points.
In the right figure, we show how the DNN predicts classes for different points in the unit square.
The testing error for interior points is 0.7836\%, i.e. less than 1\%. This experiment hence further confirms the value of careful data selection using boundary points based on a suitably chosen metric, by demonstrating substantial computational and data savings even for other more intensive techniques like DNN.

\section{Comparison with LMNN}

\subsection{Qualitative comparison with LMNN}
Given that the LMNN approach of \cite{weinberger2009distance} is arguably the nearest neighbor to our approach,
we now highlight the salient differences between the two approaches. 
Let $\mc{T}_i$ be the pre-defined target co-class neighbors of point $i$ needed as input for LMNN;
$\mc{T} = \{ (i,j) : i \in \mc{N}, j \in \mc{T}_i \}$ be the set of target pairs;
$\mc{U} = \{ (i,j,k) : i \in \mc{N}, j \in \mc{T}_i, k \in \bar{\mc{C}}_i \}$ be the triplets for which a large margin of separation is desired. 
In our notation, the LMNN optimization problem becomes
\begin{subequations} \label{model:lmnn}
\begin{alignat}{2}
\min_{\v{B},\v{s}}~~& (1-\mu) \sum_{(i,j) \in \mc{T}} \vdelta_{ij}^{\top} \v{B}\vdelta_{ij} + \mu \sum_{(i,j,k) \in \mc{U}} s_{ijk}  \\
\st~~& \vdelta_{ij}^{\top} \v{B}\vdelta_{ij} + 1 \leq \vdelta_{ik}^{\top} \v{B}\vdelta_{ik} + s_{ijk},\  \forall (i,j,k) \in \mc{U}, \\
    & s_{ijk} \geq 0, \quad \forall (i,j,k) \in \mc{U}, \\
    & \v{B} \succeq \v{0}. \label{eq:lmnn_B_psd}
\end{alignat}
\end{subequations}
Here $\mu \in [0,1]$ is a scalar weight to balance the trade-off between imposter violation penalties and the choice of distance metric.
Constraint~\eqref{eq:lmnn_B_psd} enforces $\v{B}$ to be positive semidefinite.

There are several notable differences between our approach and the LMNN.
First, \cite{weinberger2009distance} rely on a set $\mc{T}$ of pre-defined target co-class neighbors, which can lead to distorted distance metrics if not chosen judiciously (see Figure~\ref{fig:target-neighbors}). 

Second, we use a 0-1 loss function, as opposed to a hinge loss function, for outliers/imposters.
This means that, in contrast to LMNN, which incurs a small penalty for a small violation of condition ~\eqref{eq:metric_feasibility}, our approach treats even the slightest violation as a grave infringement. 
A common argument for avoiding the 0-1 loss function is that the resulting optimization problem is NP-hard.
While this is true in theory, it by no means implies that such problems are unsolvable or prohibitively expensive for current methods.  On the contrary, mixed-integer optimization solvers have witness tremendous improvements over the last two decades \cite{bixby2012brief} and challenging MILPs with millions of decision variables and constraints are routinely solved today.
Moreover, there are powerful heuristics for solving MILPs that do not guarantee provable optimality, yet are capable of quickly generating high quality solutions. 

Third, LMMN and other approaches require a minimum margin between co-class and non-class point while allowing distances to go to infinity. In contrast, we impose an upper bound of 1 on all distance pairs. It is computationally advantageous when using a 0-1 loss function for outliers to have a known finite bound on all distances.  Specifically, it allows for smaller values of $M_{ik}$, which, in turn, leads to tighter linear relaxations and faster solve times for standard mixed-integer optimization methods. 

Fourth, our distance metric is more general than a Mahalanobis metric. However, when we restrict our method to search only for a Mahalanobis metric as done in the LMNN approach, then we see that LMNN strictly enforces positive semidefiniteness, whereas our current formulation does not.  

\subsection{Numerical comparison with LMNN}

\begin{figure}
\centering 
\includegraphics[scale=1]{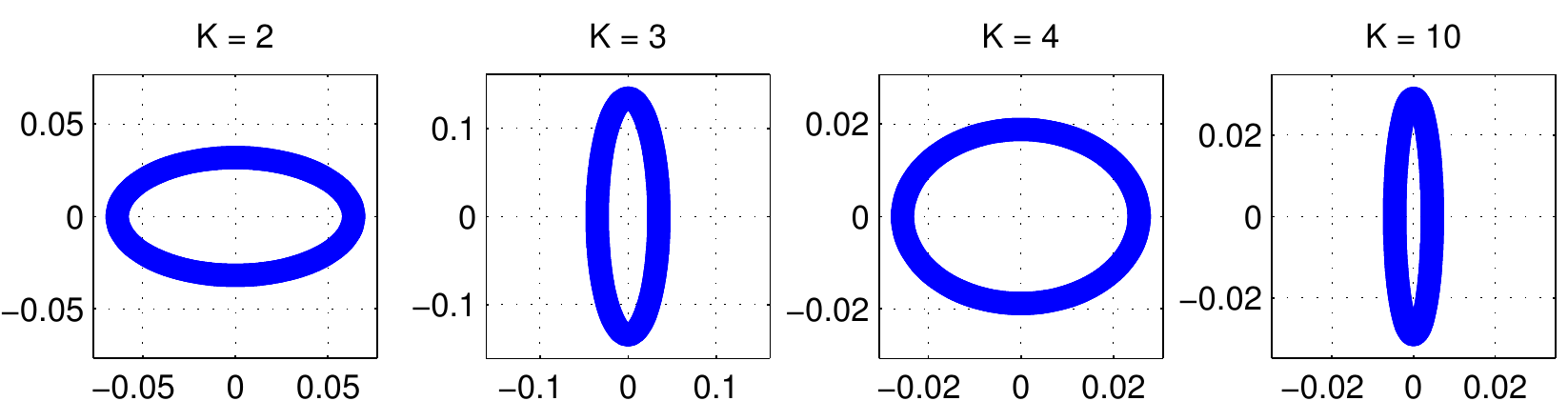}

\caption{  
Suboptimal performance of LMNN
on dataset described in  Figure \ref{fig:target-neighbors}.
The optimal metric for various $K$
from LMNN is inconsistent, thus demonstrating the difficulty of choosing target neighbours effectively. 
}
\label{fig:synthetic}
\end{figure}

As we argued in Section~\ref{sec:lit_review},
state-of-the-art methods that require target neighbors are susceptible to the choice of target neighbors, which can lead to distorted distance metrics.  
Figure~\ref{fig:synthetic} demonstrates precisely this undesirable behavior for LMNN on a synthetic data set resembling that in  Figure \ref{fig:target-neighbors}
with small vertical distance between classes.
The optimal metric for various $K$
from LMNN is inconsistent, thus demonstrating the difficulty of choosing target neighbours effectively -- even in this easy example. 
In contrast, our approach identifies the ideal Mahalanobis  metric $\v{B}^*=
\textstyle{\begin{pmatrix}
0 & 0 \\
0 & 1 \\
\end{pmatrix}}$\footnote{This means that only if we move in $y$-axis, we are measuring some distance, all points in $x$-axis have distance 0.}. 
Further, note that the
prescribed option in LMNN is the use of Euclidean distance to choose target neighbours, which can be highly misleading if the optimal metric is skewed.

\section{Metric learning modeling extensions}

\subsection{Enforcing positive semi-definiteness}

We may require the matrix $\v{B}$ to be positive semi-definite.  Gershgorin's diagonal dominance theorem provides a partial solution to this approach as it restricts $\v{B}$ to live in a restricted space of positive semi-definite matrices.  To accomplish this with our matrix $\v{B}$, let $R_{k}(\v{B})=\sum _{\ell \neq k} \left|b_{k\ell}\right|$ and note that the condition $\lambda_G^{\min}(\v{B}) \geq 0$ implies that $\min_{k=1,\ldots,p} \{b_{kk} - R_k(\v{B})\} \geq 0$, which is equivalent to $b_{kk} - \sum _{\ell \neq k} \left|b_{k\ell}\right| \geq 0$ for all $k=1,\ldots,p$.  After introducing auxiliary non-negative decision variables $b^+_{k\ell}$ to model $|b_{k\ell}|$ for all $k \neq \ell$, the latter can be converted into a set of linear constraint as follows:
\begin{subequations} \label{model:gershgorin_restriction}
\begin{alignat}{4}
& b_{kk} - \sum _{\ell \neq k} b^+_{k\ell} \geq 0 && \qquad \forall k=1,\ldots,p \\
& b^+_{k\ell} \geq b_{k\ell} && \qquad \forall k=1,\ldots,p, \ell=1,\ldots,p~(k \neq \ell) \\
& b^+_{k\ell} \geq -b_{k\ell} && \qquad \forall k=1,\ldots,p, \ell=1,\ldots,p~(k \neq \ell) \\
& b^+_{k\ell} \geq 0 && \qquad \forall k=1,\ldots,p, \ell=1,\ldots,p~(k \neq \ell)~.
\end{alignat}
\end{subequations}
Appending these constraints to the feasible region~\eqref{set:feasible_distance_metrics} allows the user to enforce a restricted version of positive semi-definiteness.

\subsection{Sparsification}

Regularization terms can easily be incorporated into the objective function to promote a sparse distance metric.
Indeed, adding the term $\mu \sum_{k,\ell: k \leq \ell} | b_{k,\ell} |$ to the objective/loss function as is done in lasso can accomplish this task.  Going a step further, one can easily include a cardinality constraint to ensure that no more than $U$ coefficients are included. 
Let $B^{\max}$ be an upper bound on the absolute value of any coefficient $b_{k,\ell}$ and let $q_{k,\ell}$ denote a binary variable that takes value 1 if $b_{k,\ell}$ is non-zero (0 otherwise). Appending the following constraints to the feasible region~\eqref{set:feasible_distance_metrics} allows the user to enforce sparsification:
\begin{equation}
-B^{\max} q_{k,\ell} \leq b_{k,\ell} \leq B^{\max} q_{k,\ell} \quad \forall k,\ell 
\end{equation}
\begin{equation}
\sum_{k,\ell: k \leq \ell} q_{k,\ell} \leq U
\end{equation}

\end{document}